\newif\ifcomments
\commentstrue
\pdfoutput=1

\documentclass[acmsmall]{acmart}

\AtBeginDocument{%
  \providecommand\BibTeX{{%
    \normalfont B\kern-0.5em{\scshape i\kern-0.25em b}\kern-0.8em\TeX}}}
    
\setcopyright{acmcopyright}
\copyrightyear{2022}
\acmYear{2022}
\acmDOI{} 
\acmJournal{CSUR}

\ifcomments
\fi

\usepackage[utf8]{inputenc}
\usepackage{graphicx}
\usepackage{wrapfig}
\usepackage{caption}
\usepackage{caption,subcaption}
\usepackage{listings}
\usepackage{paralist}
\usepackage{xcolor}
\usepackage{booktabs}
\usepackage{color}
\usepackage{soul}
\usepackage{csquotes}
\usepackage{verbatim}
\usepackage{hyperref}
\usepackage{amsthm}
\usepackage[inline]{enumitem}
\usepackage{cleveref}
\usepackage{bm}
\usepackage[english]{babel}

\definecolor{codegreen}{rgb}{0,0.6,0}
\definecolor{codegray}{rgb}{0.5,0.5,0.5}
\definecolor{codepurple}{rgb}{0.58,0,0.82}
\definecolor{backcolour}{rgb}{1.0,1.0,1.0}

\lstdefinestyle{mystyle}{
    backgroundcolor=\color{backcolour},   
    commentstyle=\color{codegreen},
    keywordstyle=\color{magenta},
    numberstyle=\tiny\color{codegray},
    stringstyle=\color{codepurple},
    basicstyle=\ttfamily\footnotesize,
    breakatwhitespace=false,         
    breaklines=true,                 
    captionpos=b,                    
    keepspaces=true,                 
    numbers=left,                    
    numbersep=5pt,                  
    showspaces=false,                
    showstringspaces=false,
    showtabs=false,                  
    tabsize=2,
    frame=tlrb,
}

\newenvironment{aligneq*}%
{
    \begin{equation*}
        \begin{aligned}
            }{
        \end{aligned}
    \end{equation*}
}

{
    \begin{equation}
        \begin{aligned}
            }{
        \end{aligned}
    \end{equation}
}

\begin{document}

\title{Constructing Neural Network-Based Models for Simulating Dynamical Systems}

\author{Christian Møldrup Legaard}
\email{cml@ece.au.dk}
\orcid{0000-0002-1914-9863}
\affiliation{%
  \institution{Aarhus University}
  \city{Aarhus}
  \country{Denmark}
}

\author{Thomas Schranz}
\email{thomas.schranz@tugraz.at}
\author{Gerald Schweiger}
\email{gerald.schweiger@tugraz.at}
\affiliation{%
  \institution{TU Graz}
  \city{Graz}
  \country{Austria}
}
\author{J\'an Drgo\v na}
\email{jan.drgona@pnnl.gov}
\affiliation{%
  \institution{Pacific Northwest National Laboratory}
  \city{Richland}
  \country{USA}
}
\author{Basak Falay}
\email{b.falay@aee.at}
\affiliation{%
  \institution{AEE-Institute for Sustainable Technologies}
  \city{Gleisdorf}
  \country{Austria}
}

\author{Cláudio Gomes}
\email{claudio.gomes@ece.au.dk}

\author{Alexandros Iosifidis}
\email{ai@ece.au.dk}

\author{Mahdi Abkar}
\email{abkar@mpe.au.dk}

\author{Peter Gorm Larsen}
\email{pgl@ece.au.dk}

\affiliation{%
  \institution{Aarhus University}
  \city{Aarhus}
  \country{Denmark}
}

\renewcommand{\shortauthors}{Legaard et al.}


\begin{abstract}
    Dynamical systems see widespread use in natural sciences like physics, biology, chemistry, as well as engineering disciplines such as circuit analysis, computational fluid dynamics, and control.
    For simple systems, the differential equations governing the dynamics can be derived by applying fundamental physical laws. 
    However, for more complex systems, this approach becomes exceedingly difficult.
    Data-driven modeling is an alternative paradigm that seeks to learn an approximation of the dynamics of a system using observations of the true system.
    In recent years, there has been an increased interest in data-driven modeling techniques, in particular neural networks have proven to provide an effective framework for solving a wide range of tasks.
    This paper provides a survey of the different ways to construct models of dynamical systems using neural networks.
    In addition to the basic overview, we review the related literature and outline the most significant challenges from numerical simulations that this modeling paradigm must overcome.
    Based on the reviewed literature and identified challenges, we provide a discussion on promising research areas.
\end{abstract}


\maketitle

\section{Introduction}

Mathematical models are fundamental tools for building an understanding of the physical phenomena observed in nature~\cite{Cellier1991}.
Not only do these models allow us to predict what the future may look like, but they also allow us to develop an understanding of what causes the observed behavior.
In engineering, models are used to improve the system design~\cite{Friedman2006,Schramm2010}, design optimal control policy~\cite{GARCIA1989335,DIEHL2002577,DRGONA2020190}, simulate faults~\cite{Pintard2013,Moradi2019}, forecast future behavior~\cite{Sohlberg2008}, or assess the desired operational performance~\cite{Jiang2014}.

The focus of this survey is on the type of models that allow us to predict how a physical system evolves over time for a given set of conditions.
Dynamical systems theory provides an essential set of tools for formalizing and studying the dynamics of this type of model.
However, when studying complex physical phenomena, it becomes increasingly difficult to derive models by hand that strike an acceptable balance between accuracy and speed.
This has led to the development of fields that are concerned with creating models directly from data such as \emph{system identification}~\cite{Nelles2001, Ljung2006}, \emph{machine learning}~(ML)~\cite{Bishop2006,Murphy2012} and more recently, \emph{deep learning}~(DL)~\cite{Goodfellow2016}.

In recent years, the interest in DL has increased rapidly as evident from the volume of research being published on the topic~\cite{Plebe2019}.
The exact causes behind the success of \emph{neural networks}~(NNs) are hard to pinpoint.
Some claim that practical factors like the availability of large quantities of data, user-friendly software frameworks~\cite{Paszke2019,Abadi2016}, and specialized hardware~\cite{Mittal2019} are the main cause for its success, while others claim that the success of NNs can be attributed to their structure being well suited to solving a wide variety of problems~\cite{Plebe2019}.

The goal of this survey is to provide a practical guide on how to construct models of dynamical systems using NNs as primary building blocks.
We do this by walking the reader through the most important classes of models found in the literature; for many of which we provide an example implementation.
We put special emphasis on the process for training the models, since it differs significantly from traditional applications of DL that do not consider evolution over time.
More specifically, we describe how to split the trajectories used during training, and we introduce optimization criteria suitable for simulation.
After training, it is necessary to validate that the model is a good representation of the true system.
Like other data-driven models we determine the validity empirically by using a separate set of trajectories for validation.
We introduce some of the most important properties and how they can be verified.

\begin{figure}
    \centering
    \includegraphics[width=.75\textwidth]{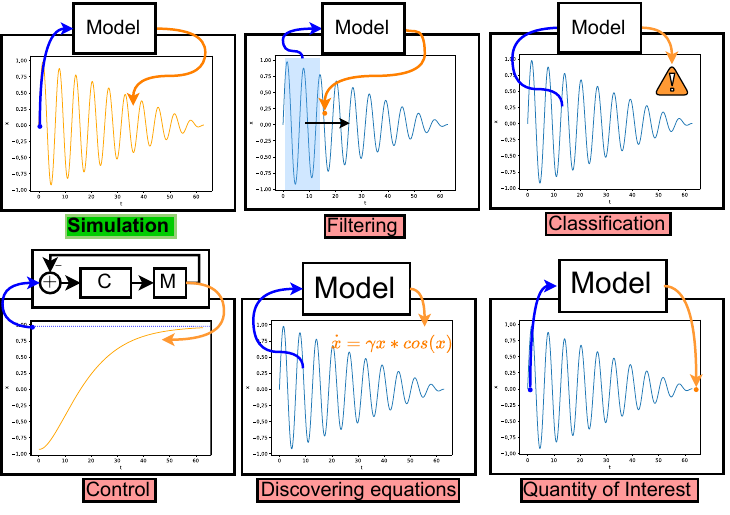}
    \caption{
    \emph{Simulation} and related application areas where ML techniques are commonly applied. 
    The focus of the survey is exclusively on techniques that can generate a simulation based on an initial condition, as shown on the top left. 
    Although interesting on their own, topics other than simulation are not covered by the survey.
    \emph{Filtering} refers to applications where a sliding window over past observations is used to predict the next sample or some other quantity of interest.
    \emph{Classification} refers to applications where a model takes a sequence of observations and produces a categorical label, for instance, indicating that the system is in an abnormal state.
    \emph{Control} refers to applications where a NN-based controller is used to drive the system to a desired state.
    \emph{Discovering Equations} refers to techniques based on ML that aim to discover the underlying equations of the system.
    \emph{Quantity of Interest} refers to applications where a neural network is used to provide a mapping from an initial condition to some quantity of interest, for instance the steady-state of the system.}
    \label{fig:scoping}
\end{figure}

It should be emphasized that the type of model we wish to construct should allow us to obtain a simulation of the system.
Rather than providing a formal definition of simulation we refer to \cref{fig:scoping}, which shows several topics related to simulation that are not covered by this paper.

\textbf{The source code and instructions for running the experiments can be accessed in the following repository\footnote{\url{https://github.com/clegaard/deep_learning_for_dynamical_systems}}}. 

\subsection{Related Surveys}

We provide an overview of existing surveys related to our work. 
Then we compare our work with these surveys and describe the structure of the remainder of the paper. 

\paragraph{Application Domain}

The broader topic of using ML in scientific fields has received widespread attention within several application domains \cite{Rolnick2019,Brunton2020,Butler2018,Ching2018}.
Common for these review papers is that they focus on providing an overview of the prospective use cases of ML within their domains, but put limited emphasis on how to apply the techniques in practice.

\paragraph{Surrogate Modeling}

The field of surrogate modeling, i.e. the theory and techniques used to produce faster models, is intimately related to the field of simulation with NNs.
So it is important that we highlight some surveys in this field.
The work in~\cite{Koziel2020} presents a thorough introduction to data-driven surrogate modeling, which encompasses the use of NNs.
The authors of~\cite{Viana2010} summarize advanced and yet simple statistical tools commonly used in the design automation community: (i) screening and variable reduction in both the input and the output spaces, (ii) simultaneous use of multiple surrogates, (iii) sequential sampling and optimization, and (iv) conservative estimators.
Since optimization is an important use case of surrogate modeling,~\cite{Forrester2009} reviewed advances in surrogate modeling in this field.
Finally, with a focus on applications to water resources and building simulation, we highlight the work in~\cite{Westermann2019,Razavi2012}.

\paragraph{Prior Knowledge} 
\label{sec:prior_knowledge}

One of the major trends to address some challenges arising in NNs based simulation is to encode prior knowledge such as physical constraints into the network itself or during the training process, ensuring the trained network is physically consistent.
The work in~\cite{Karpatne2017} coins this \emph{theory-guided data science} and provides several examples of how knowledge may be incorporated in practice. Closely related to this is the work in~\cite{Rueden2020,Rueden2020a,Rai2020}, which proposes a detailed taxonomy describing the various paths through which knowledge can be incorporated into a NN model.

\paragraph{Comparison with this survey}

Our work complements the above surveys by providing an in-depth review focused specifically on NNs rather than ML as a whole.
The concrete example helps the reader's understanding and highlights the similarities and inherent deficiencies of each approach.

We also outline the inherent challenges of simulation and establish a relationship between numerical simulation challenges and DL-based simulation challenges.
The benefit of our approach is that the reader gets the intuition behind some approaches used to incorporate knowledge into the NNs.
For instance, we relate energy-conserving numerical solvers to Hamiltonian neural networks, whose goal is to encode energy conservation, and we discuss concepts such as numerical stability and solver convergence, which are crucial in long-term prediction using NNs.

\subsection{Survey Structure}

\begin{figure}
    \centering
    \includegraphics[width=\textwidth]{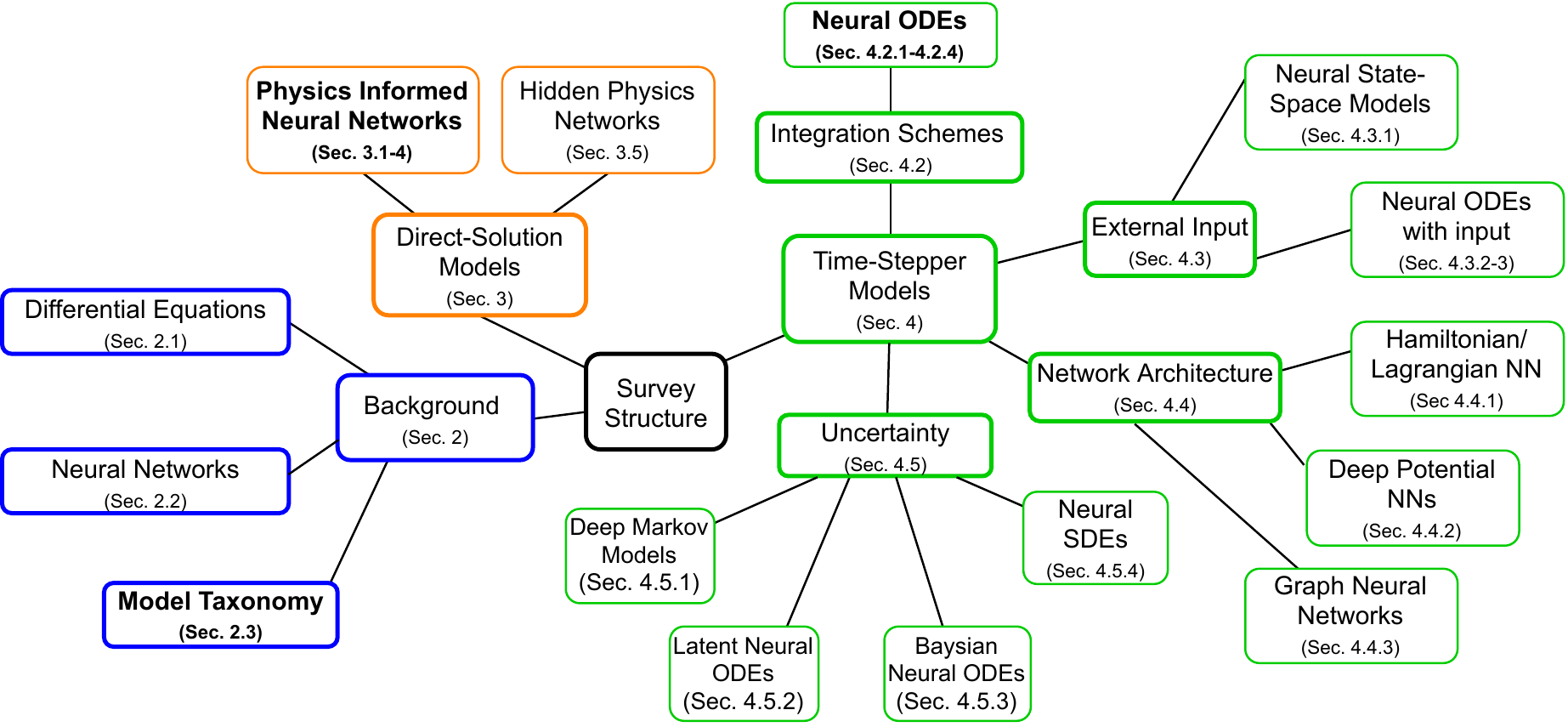}
    \caption{A mind map of the topics and model types covered in the survey.}
    \label{fig:survey_structure}
\end{figure}

The remainder of the paper is structured according to the mind-map shown in \cref{fig:survey_structure}.
First, \cref{sec:background} introduces the central concepts of dynamical systems, numerical solvers, neural networks.
Additionally, the section proposes a taxonomy describing the fundamental differences of how models can be constructed using NNs.
The following two sections are dedicated to describing the two classes of models identified in the taxonomy: \emph{direct-solution models} and \emph{time-stepper models} in \cref{sec:continuous_time_models} and \cref{sec:time_stepper_models}, respectively.
For each of the two categories, we describe:
\begin{itemize}
    \item The structure of the model and the mechanism used to produce simulations of a system.
    \item How the parameters are tuned to match the behavior of the true system.
    \item Key challenges and extensions of the model designed to address them.
\end{itemize}
Following this, \cref{sec:discussion} discusses the advantages and limitations of the two distinct model types and outlines future research directions.
Finally, \cref{sec:conclusion} provides a brief summary of the contributions of the paper and the outlined research directions.

\section{Background}
\label{sec:background}

\emph{Models} are an integral tool in natural sciences and engineering that allow us to deepen our understanding of nature or improve the design of engineered systems.
One way to categorize models is by the \emph{modeling} technique used to derive the model: \emph{First Principles} models derived using fundamental physical laws, and \emph{Data Driven} models created based on experimental data.

First, in \cref{sec:ms_dynamical_systems}, a running example is introduced, where we describe how differential equations can be used to model a simple mechanical system and how a solver is used to obtain a simulation.
Then \cref{sec:modelling_with_nns} introduces the different ways NN-based models of the system can be constructed and trained.
Finally, \cref{sec:background_taxonomy} introduces a taxonomy of the different ways NNs can be used to construct models of dynamical systems.

\subsection{Differential Equations} \label{sec:ms_dynamical_systems}
\begin{figure}
    \centering
    \includegraphics[width=0.3\linewidth]{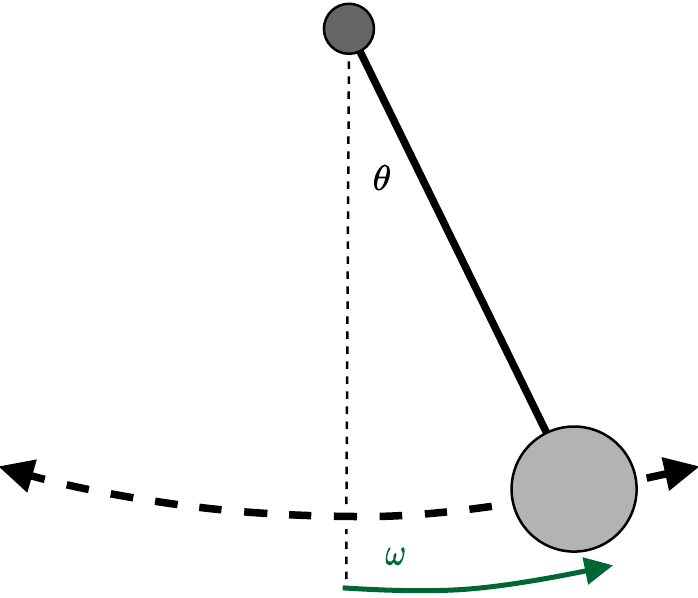}
    \caption{
    The ideal pendulum system used as a case study throughout the paper.
    The pendulum is characterized by an angle,~$\theta$, and an angular velocity,~$\omega$.
    }
    \label{fig:cart_pendulum_diagram}
\end{figure}

An \emph{ideal pendulum}, shown in \cref{fig:cart_pendulum_diagram}, refers to a mathematical model of a pendulum that, unlike its physical counterpart, neglects the influence of factors such as friction in the pivot or bending of the pendulum arm.
The state of this system can be represented by two variables: its angle $\theta$ (expressed in radians), and its angular velocity $\omega$.
These variables correspond to a mathematical description of the system's state and are referred to as \emph{state variables}.
The way that a given point in state-space evolves over time can be described using \emph{differential equations}.
Specifically, for the ideal pendulum, we may use the following \emph{ordinary differential equation}~(ODE):
\begin{equation} \label{eq:ideal_pendulum}
    \frac{\partial^2 \theta}{\partial t^2} + \frac{g}{l} \sin{\theta} = 0,
\end{equation}
where $g$ is the gravitational acceleration, and $l$ is the length of the pendulum arm.
The ideal pendulum \cref{eq:ideal_pendulum} falls into the category of \emph{autonomous} and \emph{time-invariant}-systems since the system is not influenced by external stimulus and the dynamics do not change over time.
While this simplifies the notation and the way in which models can be constructed, it is not the general case.
We discuss the implication of these issues in \cref{sec:external_inputs}.

The equation can be rewritten as two first order differential equations and expressed compactly using vector notation as follows:

\begin{equation} \label{eq:ode}
        f(x) = \begin{bmatrix} 
        \frac{\partial \omega}{\partial t} \\
        \frac{\partial\theta}{\partial t} 
    \end{bmatrix} = 
    \begin{bmatrix} 
        -\frac{g}{l}\sin{\theta} \\
        \omega 
    \end{bmatrix}.
\end{equation}
\noindent where $x$ is a vector of the system's state variables.
In the context of this paper, we refer to $f(x)$ as the \emph{derivative function} or as the derivative of the system.

While the differential equations describe how each state variable will evolve over the next time instance, they do not provide any way of determining the solution $x(t)$ on their own.
Obtaining the solution of an ODE $f(x)$ given some \emph{initial conditions} $x_0$ is referred to as an \emph{initial value problem}~(IVP) and can be formalized as:

\begin{align} \label{eq:ivp}
    \frac{\partial}{\partial t}x(t) &= f(x(t)), \\
    x(t_0) &= x_0
\end{align}
\noindent where $x(\cdot)$ is called the \emph{solution}, $x : \mathbb{R} \rightarrow \mathbb{R}^n$ and $n \in \mathbb{N}$ is the dimension of the system's state space.

\begin{figure}
    \centering
    \begin{subfigure}[t]{0.51\textwidth}
        \centering
        \includegraphics[width=\textwidth]{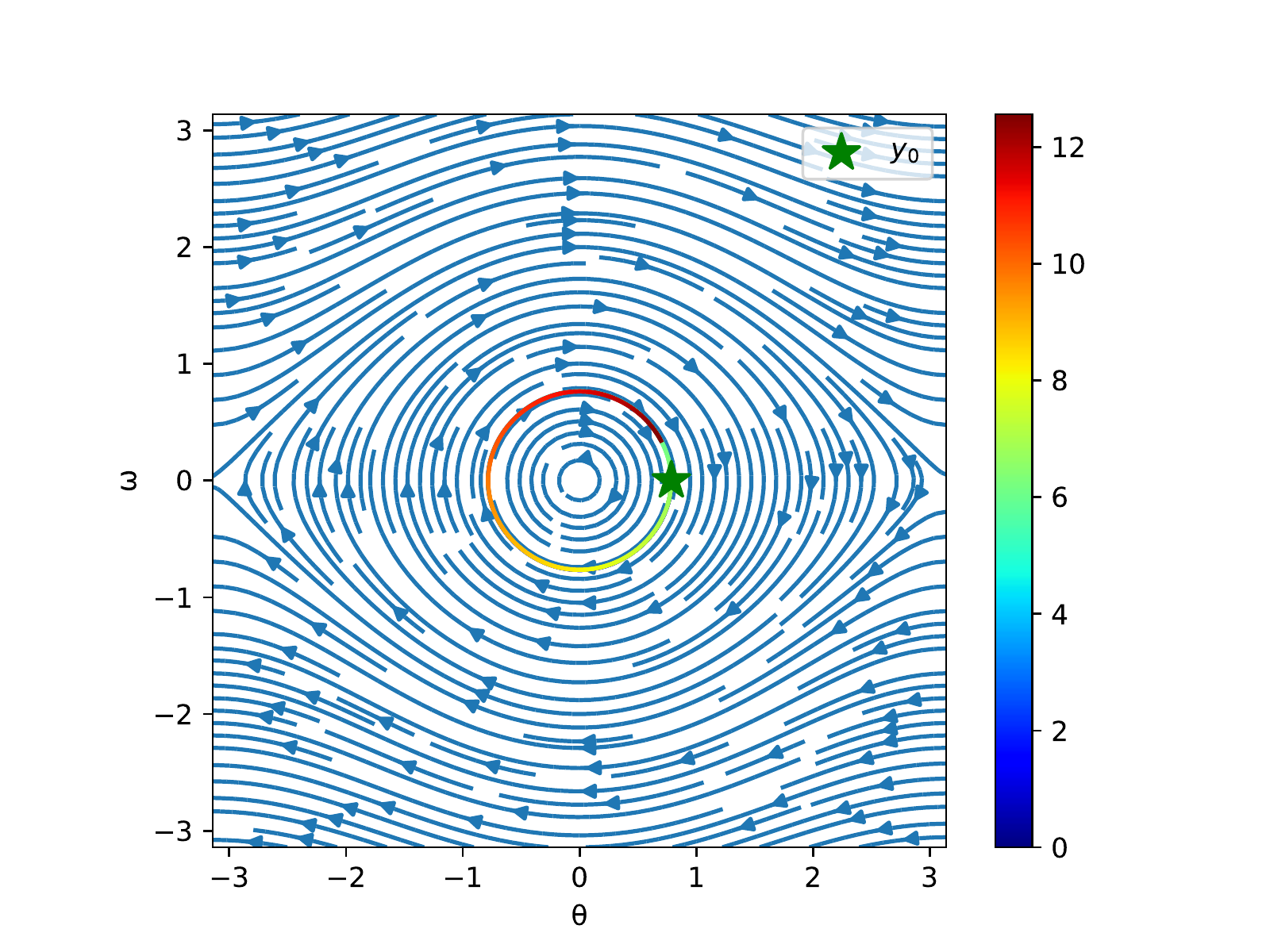}
        \caption{Phase portrait of the ideal pendulum with a single trajectory drawn onto the phase space. The color denotes time.}
        \label{fig:pendulum_phase_portrait}
    \end{subfigure}
    \hfill
    \begin{subfigure}[t]{0.47\textwidth}
        \centering
        \includegraphics[width=\textwidth]{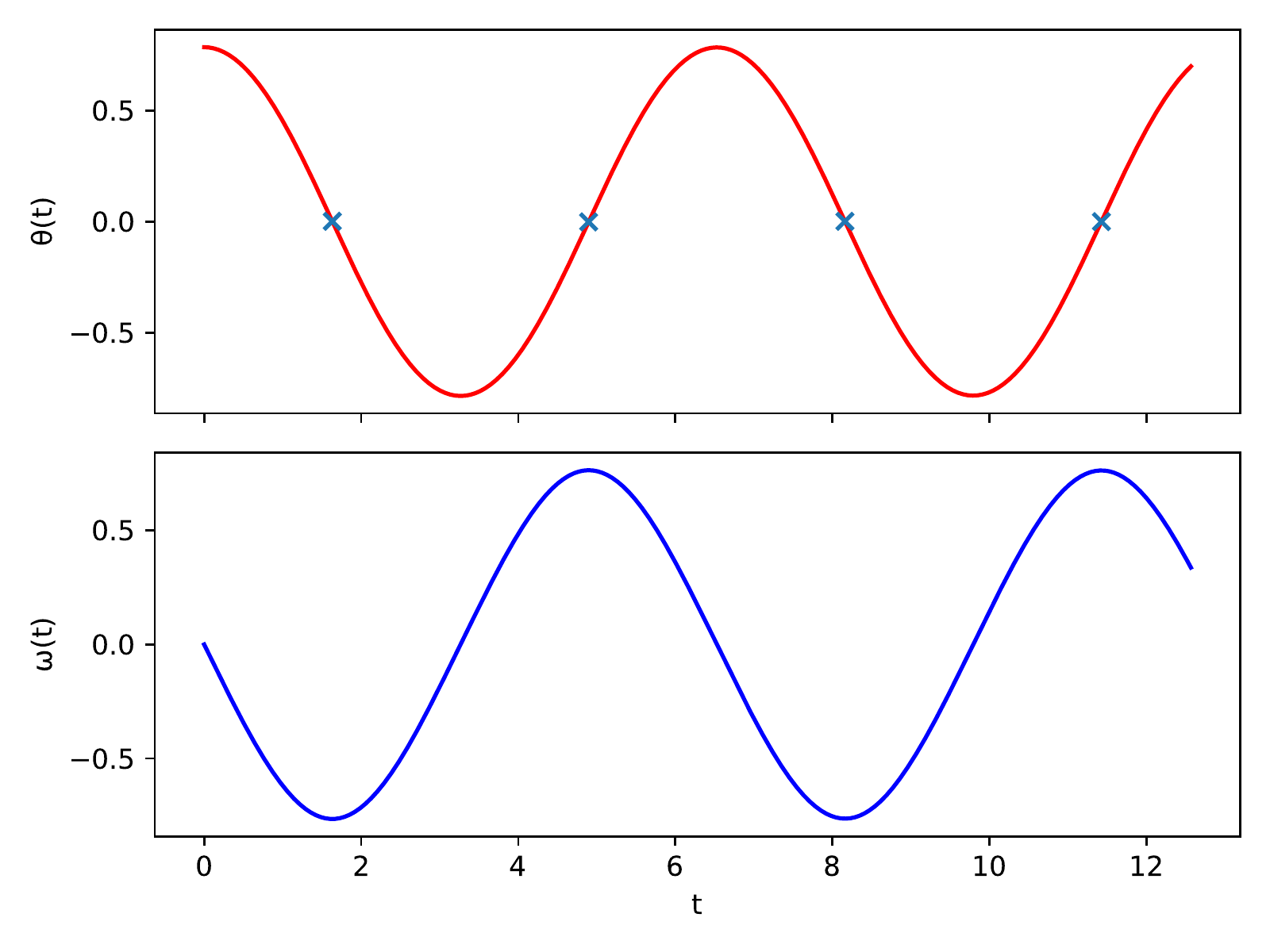}
        \caption{Solution of \cref{eq:ivp} for the initial condition marked with a star in  \cref{fig:pendulum_phase_portrait}.}
        \label{fig:pendulum_simulation}
    \end{subfigure}
    \caption{Diagram of pendulum system and example of the trajectory generated when solving the equation using a numerical solver.}
\end{figure}

The result of solving the IVP corresponding to the pendulum can be seen in \cref{fig:pendulum_simulation} which shows how the two state variables $\theta$ and $\omega$ evolve from their initial state.
An alternative view of this can be seen in the \emph{phase portrait} in \cref{fig:pendulum_phase_portrait}. 

In many cases it is impossible to find an exact analytical solution to the IVP, and instead numerical methods are used to approximate the solution.
Numerical solvers are algorithms that approximate a continuous IVP, as the one in \cref{eq:ode}, into a discrete time dynamical system.
These systems are often modeled with difference equations:
\begin{equation}\label{eq:difference_eq}
    x_{k+1} = F(x_k),
\end{equation}

\noindent where $x_k$ represents the state vector at the $k$-th time point, $x_{k+1}$ represents the next state vector, and $F : \mathbb{R}^n \rightarrow \mathbb{R}^n$ models the system behavior.
Just as with ODEs, the initial state can be represented by a constraint on $x_0$, and the solution to \cref{eq:difference_eq} with an initial value defined by such constraint is a function $x_k$ defined for all $k \geq 0$.
In \cref{eq:difference_eq}, time is implicitly defined as a discrete set.

We start by introducing the simplest and most intuitive numerical solver, because it highlights the main challenges well.
There are many numerical solvers, each presenting unique trade-offs.
The reader is referred to \cite{Cellier2006} for an introduction to this topic, to \cite{Wanner1991,Hairer1996} for more detailed expositions on numerical solution of ODEs and differential-algebraic system of equations (DAEs), to \cite{LeVeque2007} for the numerical solution to partial differential equations (PDEs), to \cite{Marsden2001} for an overview of more advanced numerical schemes, and to \cite{Kofman2001} for an introduction to quantized state solvers.

Given an IVP -- \cref{eq:ivp}
-- and a simulation step size $h>0$, the Forward Euler (FE) method computes a sequence in time of points $\tilde{x}_k$, where $\tilde{x}_k$ is the approximation of the solution to the IVP at time $h k$: $\tilde{x}_k \approx x_k = x(h k) $.
It starts from the given initial value $\tilde{x}_0 = x(0)$ and then computes iteratively:
\begin{equation}\label{eq:fw_euler_solver}
    \tilde{x}_{k+1} = \tilde{x}_k + h f(t_k,\tilde{x}_k), 
\end{equation}
where $f : \mathbb{R} \times \mathbb{R}^n \rightarrow \mathbb{R}^n$ is the ODE right-hand side in \cref{eq:ode} and $t_k = h k$.

A graphical representation of the solutions IVP starting from different initial conditions can be seen in \cref{fig:pendulum_phase_portrait}.
For a specific point, the solver evaluates the derivative (depicted as curved arrows in the plot) and takes a small step in this direction.
Applying this process iteratively results in the full trajectory, which for the pendulum corresponds to the circle in the phase space.
The circle in the phase space implies that the solution is repeating itself, i.e. corresponds to an oscillation in time as seen in \cref{fig:pendulum_simulation}.

The ideal pendulum is an example of a well-studied dynamical system for which the dynamics can be described using simple ODEs that can be solved using standard solvers.
Unfortunately, the simplicity of the idealized model comes at the cost of neglecting several factors which are present in a real pendulum.
For example, the arm of the real pendulum may bend and energy may be lost in the pivot due to friction.
The idealized model can be extended to account for these factors by incorporating models of friction and bending.
However, this is time-consuming, leads to a model that is harder to interpret, and it does not guarantee that all factors are accounted for.


\subsection{Neural Networks} \label{sec:modelling_with_nns}

Today, the term \emph{neural network} has come to encompass a whole family of models, which collectively have proven to be effective building blocks for solving a wide range of problems.
In this paper, we focus on a single class of networks, the \emph{fully-connected}~(FC) NNs, due to their simplicity and the fact that they will be used to construct the models introduced in later sections.
We refer the reader to~\cite{Goodfellow2016} for a general introduction to the field of DL.

\begin{figure}[ht]
\begin{subfigure}{0.5\columnwidth} 
    \centering
    \includegraphics[width=\textwidth]{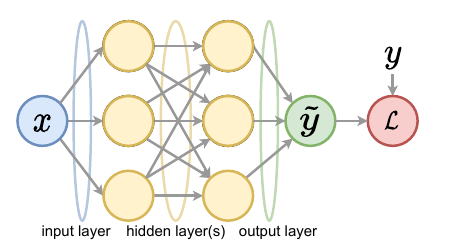}
    \caption{ }
    \label{fig:fc_regression}
\end{subfigure}
\begin{subfigure}{0.45\columnwidth} 
\centering
\begin{subfigure}{\textwidth} 
\begin{lstlisting}[mathescape=true]
$\tilde{y}$ = network(x)
\end{lstlisting}
\caption{ Inference.}
\end{subfigure}
\begin{subfigure}{\textwidth}
\begin{lstlisting}[mathescape=true]
$\tilde{y}$ = network(x)
loss = $\mathcal{L}(\tilde{y},y)$
optimizer.step(loss)
\end{lstlisting}
\caption{ Training. This step is typically repeated many times for different inputs and desired output values.}
\end{subfigure}
\end{subfigure}

\caption{ A Fully-connected neural network being used to perform regression from an input $x$ to $y$, where $\tilde{y}$ represents the approximation provided by the NN. 
Each layer of the network is characterized by a set of weights that are tuned during training to produce the desired output for a given input.
During training, the loss function $\mathcal{L} : \mathbb{R}^n \rightarrow \mathbb{R}$ is used to measure the divergence between the output produced by the network, $\tilde{y}$, and the desired output $y$. }
\end{figure}

Like other data-driven models, NNs are generic structures which prior to training have no behavior specific to the problem they are being applied to.
For this reason, it is essential to consider not only how the network produces its outputs, but also how the network's parameters are tuned to solve the problem.
For instance, we may consider using a FC NN to perform regression from a scalar input, $x$, to a scalar output, $y$, as shown in \cref{fig:fc_regression}. 
In the context of the survey, we will refer to the process of producing predictions as \emph{inference} and the process of tuning the network's weights to produce the desired results as \emph{training}.
There can be quite drastic differences in the complexity of the two phases, the training phase typically being the most complex and computationally intensive.
During training, a loss function, $\mathcal{L} : \mathbb{R}^n \rightarrow \mathbb{R}$ defines a mapping from the predicted trajectory, to a scalar quantity that is a measure of how close the prediction is to the true trajectory.
Unless stated otherwise, we define this to be the \emph{mean squared error}~(MSE) between the predicted and true trajectory:
\begin{equation} \label{eq:mse}
    \mathcal{L}(\tilde{y};y) = \frac{1}{nm}\sum_{i=1}^{m}\sum_{j=1}^{n} (\tilde{y}_{ij} - y_{ij})^2
\end{equation}
\noindent where $m$ is the length of the trajectory, $n$ is the dimension of the system's state-space, and $y_{ij}$ denotes the value of the $j$-th state at the $i$-th point of time of the trajectory.

\subsection{Model Taxonomy} \label{sec:background_taxonomy}

A challenge of studying any fast evolving research field, such as DL, is that the terminology used to describe important concepts and ideas may not always have converged.
This is especially true in the intersection between DL, numerical simulation and physics, due to the influx of ideas and terminology from the different fields.
In the literature, there is also a tendency to focus on the success of a particular technique on a specific application, with little emphasis on explaining the inner workings and limitations of the technique.
A consequence of this is that important contributions to the field become lost due to the papers being hard to digest.

\begin{figure}[ht]
    \centering
    \begin{subfigure}[t]{.50\columnwidth}
        \centering
        \includegraphics[width=\columnwidth]{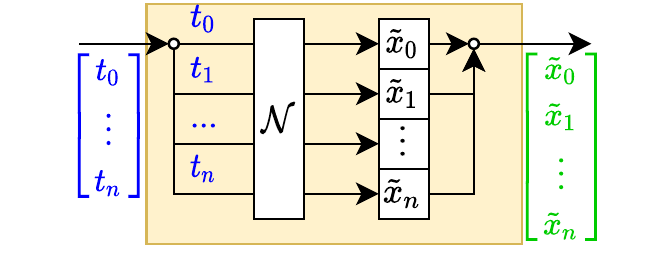}
        \caption{ Direct-solution model. A NN is used to parameterize a mapping from a time instance to the solution corresponding to that time instance.}
        \label{fig:overview_continuous_time_model}
    \end{subfigure}
    \hfill
    \begin{subfigure}[t]{.45\columnwidth}
        \centering
        \includegraphics[width=\columnwidth]{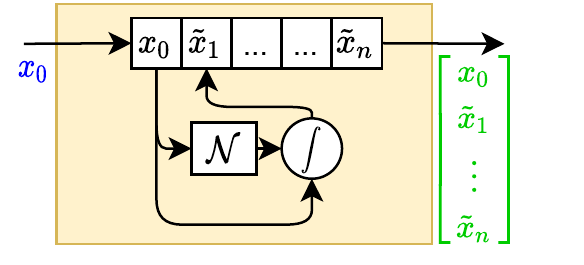}
        \caption{ Time-stepper model. The network,~$\mathcal{N}$, provides the derivative of the system at various points in state-space, which is then integrated by a numerical solver, here depicted as $\int$.}
        \label{fig:overview_time_stepper_model}
    \end{subfigure}
    \caption{Overview of two distinct model types. 
    Direct-solution models are trained to produce a simulation without performing numerical integration explicitly. 
    Conversely, time-stepper models use the same techniques known from numerical simulation to produce a simulation of the system.}
    \label{fig:model_type_overview}
\end{figure}

In an attempt to alleviate this, we propose a simple taxonomy describing how models can be constructed consisting of two categories: \emph{direct-solution models} and \emph{time-stepper models}, as shown in~\cref{fig:model_type_overview}.
Direct-solution models, described in~\cref{sec:continuous_time_models}, do not employ integration; but rather produce an estimate of the state at a particular time by feeding in the time as an input to the network. 
Time-stepper models, found in~\cref{sec:time_stepper_models}, can be characterized by using a similar approach to numerical solvers, where the current state is used to calculate the state at some time into the future.
The difference between the time-stepper and continuous models has significant implications for how the model deals with varying initial conditions and inputs.
Per design, the time-stepper models handle different initial conditions and inputs, whereas direct-solution models have to be re-trained.
In other words, the time-stepper models learn the dynamics while the direct-solution models learn a solution to an IVP for a given initial state and set of inputs.

\section{Direct-Solution Models}
\label{sec:continuous_time_models}

\begin{table}
    \centering
    \caption{Comparison of direct-solution models.}
    \begin{tabular}{|c|c|c|c|c|}
    \hline
        Name & $In_{NN}$ & $Out_{NN}$ & $Out_{AD}$ & Uses Equations  \\
    \hline
         Naive direct-solution & $t$ & $\theta,\omega$ & &  \\ \hline
         Autodiff direct-solution & $t$ & $\theta$ & $\omega$ & \\ \hline 
         Physics-Informed Neural Network & $t$  & ${\theta}$ & $\omega, \partial \omega$  & \checkmark  \\ \hline 
         Hidden Physics Neural Network & $t$ & $\theta,l$ & $\omega, \partial \omega$ & \checkmark  \\ \hline 
    \end{tabular}
    
    \label{tab:ct_comparison}
\end{table}

One approach for obtaining the trace of a system is to construct a model that maps a time, $t_k$, to the solution at that time, $x_k$.
We refer to this type of model as a \emph{direct-solution} model.

To construct the model, a NN is trained to provide an exact solution for a set of \emph{collocation} points which are sampled from the true system.
Another way to view this is that the NN acts as a trainable interpolation engine, which allows the solution to be evaluated at arbitrary points in time, not only those of the collocation points.
An important limitation of this approach is that a trained model is fixed for a specific set of initial conditions.
To evaluate the solution for different initial conditions, a new model would have to be trained on new data.

In the literature, this type of model is often applied to learn the dynamics of systems governed by PDEs and less frequently systems governed by ODEs.
Several factors are likely to influence this pattern of use.
First, PDEs are generally harder and more computationally expensive to solve than ODEs, which provides a stronger motivation for applying NNs as a means to obtain a solution.
Secondly, many practical uses of ODEs require that they can be evaluated for different initial conditions with ease, which is not the case for direct-solution models.

While the motivation for applying direct-solution networks may be strongest for PDEs, they can also be applied to model ODEs.
The main difference is that a network to model an ODE takes only time as input, whereas the network used to model a PDE would take both time and spatial coordinates.

A key challenge in training direct-solution NNs is the amount of data required to reach an acceptable accuracy and level of generalization.
A naive approach that does not leverage prior knowledge, like the one described in \cref{sec:cs_vanilla}, is likely to fit the collocation points very well but fails to reproduce the underlying trend.
A recent trend popularized by \emph{physics-informed neural networks}~(PINNs)~\cite{Raissi2019} is to apply clever use of automatic differentiation and equations encoding prior knowledge to improve the generalization of the model.

The remaining part describes how the different types of direct-solution models, shown in \cref{tab:ct_comparison}, can be applied to model the ideal pendulum system for a specific initial condition.
First, the architecture of the NNs used for the experiments is introduced in \cref{sec:cs_methodology}.
Next, the simplest approach is introduced in \cref{sec:cs_vanilla}, before progressively building up to a model type that incorporates features from all prior models in \cref{sec:cs_hidden}.

\subsection{Methodology} \label{sec:cs_methodology}
The examples of direct-solution models shown in this section use a fully connected NN with 3 hidden layers consisting of 32 neurons each.
The output of each hidden layer is followed by a softplus activation function.

Each model is trained on a trajectory corresponding to the simulation for a single initial condition, which is sampled to obtain a set of collocation points as shown in \cref{fig:prediction_continuous_model_vanilla}.
The goal of the training is to obtain a model that can predict the solution at any point in time, not only those coinciding with the collocation points.

\subsection{Vanilla Direct-Solution}\label{sec:cs_vanilla}

Direct-solution models produce an estimate $\tilde{x}_k$ of the system state $x_k$ at time $t_k$. 
The models learn a continuous function of time that can be evaluated at any arbitrary point in time by introducing $t_k$ into the network:
\begin{equation} \label{eq:continuous_model}
    \tilde{x}_k = N(t_k).
\end{equation}
\noindent where $N : \mathbb{R} \rightarrow \mathbb{R}^n$ represents the mapping from a point in time to the solution at the given time.

\begin{figure}
	\centering
	\begin{subfigure}[t]{0.45\textwidth}
		 \centering
            \includegraphics[width=\textwidth]{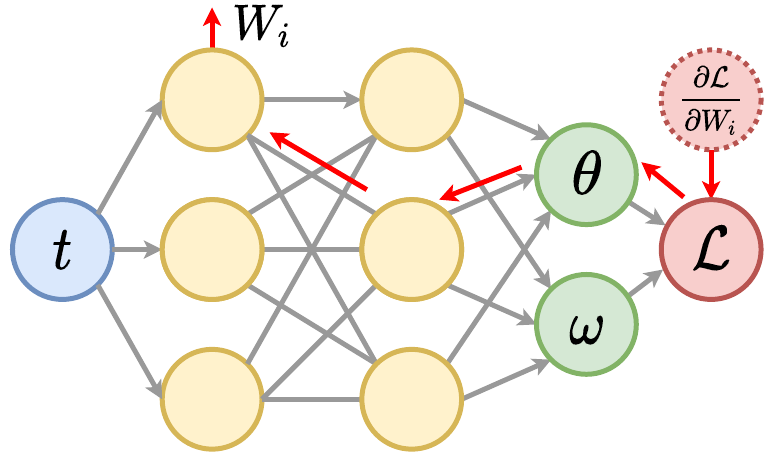}
            \caption{Network structure. }
            \label{fig:continuous_model}
	\end{subfigure}
	\hfill
	\begin{subfigure}[t]{0.45\textwidth}
        \centering
        \includegraphics[width=\textwidth]{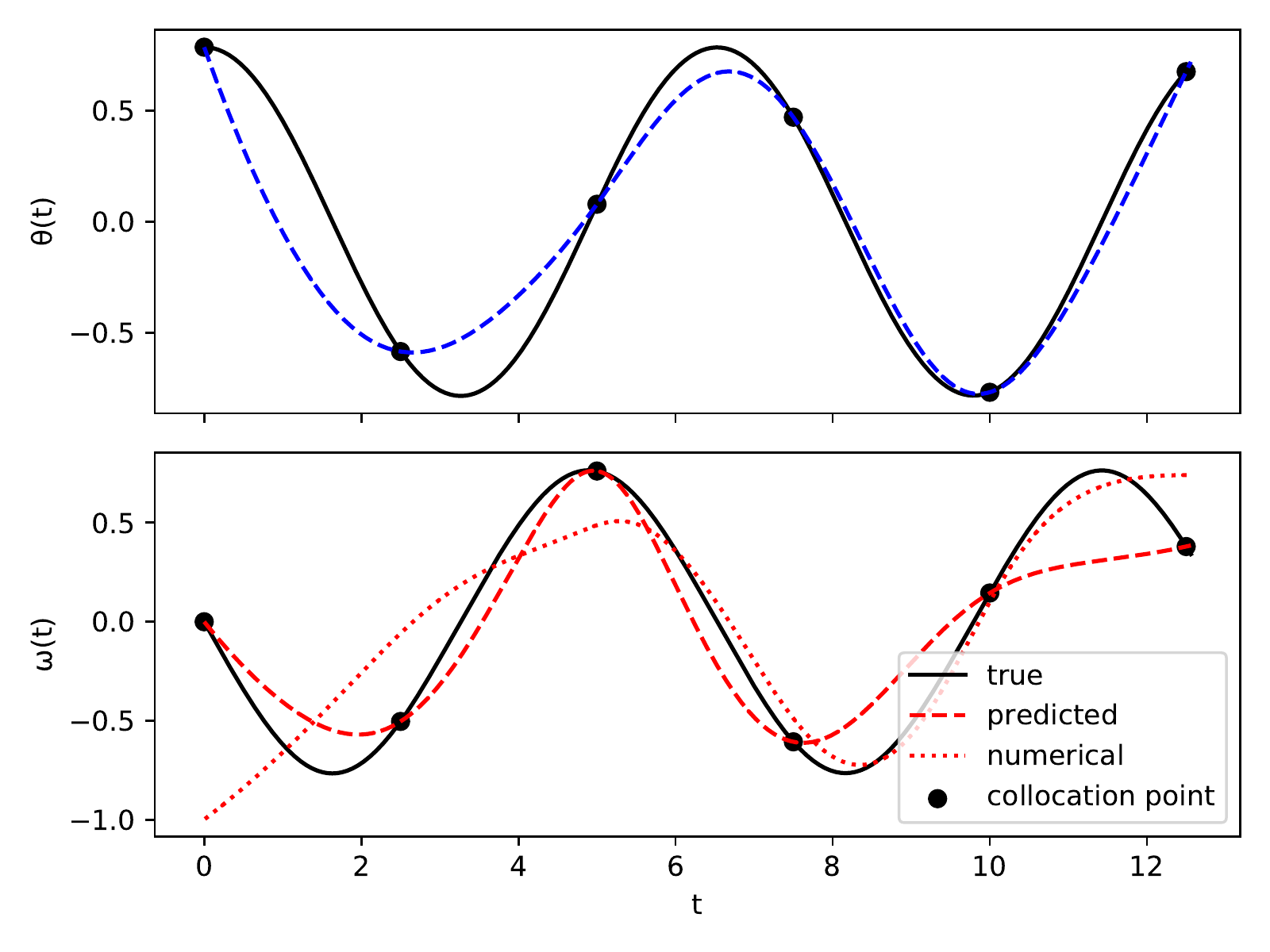}
        \caption{ Predictions. }
        \label{fig:prediction_continuous_model_vanilla}
    \end{subfigure}

\begin{subfigure}[t]{.45\textwidth}
\begin{lstlisting}[
    mathescape=true
]
$\tilde{\theta}$,$\tilde{\omega}$ = network(t)
loss = MSE(($\tilde{\theta}$, $\tilde{\omega}$), ($\theta$, $\omega$))
optimizer.step(loss)
\end{lstlisting}
\caption{Training.}
\label{lst:continuous_vanilla_training}
\end{subfigure}
\hfill
\begin{subfigure}[t]{.45\textwidth}
\begin{lstlisting}[
    mathescape=true
]
$\tilde{\theta}$,$\tilde{\omega}$ = network(t)
\end{lstlisting}
\caption{Inference.}
\label{lst:continuous_vanilla_inference}
\end{subfigure}

\caption{Vanilla direct-solution model. 
Predictions of the two state variables. 
Black dots indicate the collocation points, i.e. the points in which the loss function is minimized.
The network fits all collocation points well, but fails to generalize to the interval between points.
Additionally, the predicted $\omega$ is very different to the approximation obtained using numerical differentiation of $\theta$.}
\end{figure}

To model the pendulum a feed-forward network with a single input $t$ and two outputs $\theta$ and $\omega$ could be used to construct the model, as depicted in \cref{fig:continuous_model}.
To obtain the solution for multiple time instances the network can simply be evaluated multiple times.
There are no dependencies between the estimates of multiple states, allowing one to evaluate all of these in parallel.

The network is trained by minimizing the difference between the predicted and the true trajectory in the collocation points shown in \cref{fig:prediction_continuous_model_vanilla} using a distance metric such as MSE defined by \cref{eq:mse}.

It is important to emphasize that the models learn a sequence of system states characterized by a specific set of initial conditions, i.e. the initial conditions are encoded into the trainable parameters of the network during training and cannot be modified during inference.

Direct-solution models are sensitive to the quality of training data. 
NNs are used to find mappings between sparse sets of input data and the output. 
Even a simple example in the data-sampling strategy can influence their generalization performance. 
Consider the trajectory in~\cref{fig:prediction_continuous_model_vanilla}; 
while the NN trained on the collocation points can produce a prediction that matches the points perfectly, while its generalization performance is poor, i.e. between the collocation points the predicted trajectory does not match the true development.

It is worth noting that there are many ways that this can go wrong, i.e., given a sufficiently sparse sampling, it is not just one specific choice of training points that makes it impossible for the network to learn the true mapping.
The obvious way to mitigate the issue is to obtain more data by sampling at a higher rate. 
However, there are cases where data acquisition is expensive, impractical or where it is simply impossible to change the sampling frequency.


Consider a system where one state variable is the derivative of the other, a setting which is quite common in systems that can be described by differential equations. 
A naive direct-solution model cannot guarantee that the relationship between the predicted state variables respects this property.
Fig.~\ref{fig:prediction_continuous_model_vanilla} provides a graphical representation of the issue.
While the model predicts both system state variables correctly in the collocation points, it can clearly be seen that the estimate for $\omega$ is neither the derivative of $\tilde{\theta}$ nor does it come close to the true trajectory.

\subsection{Automatic Differentiation in Direct-Solution}
\label{sec:continuous_autodiff}

\begin{figure}[ht]
\centering
\begin{subfigure}[t]{0.45\textwidth}
	\centering
	\includegraphics[width=\textwidth]{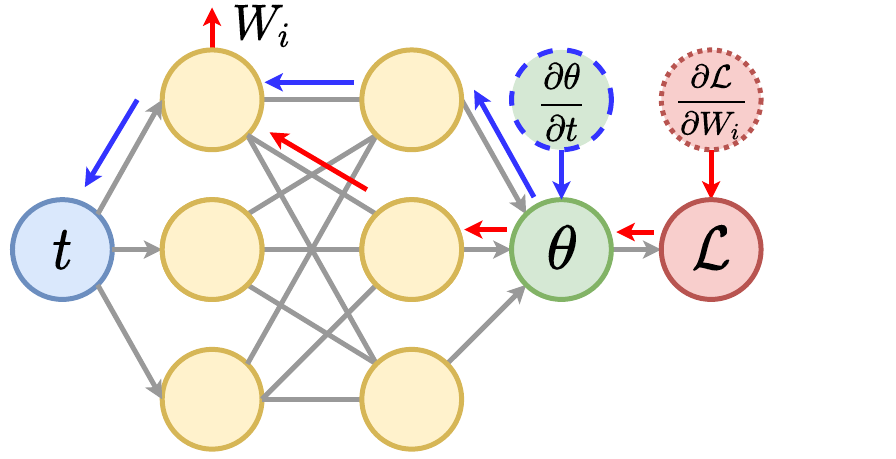}
	\caption{ Network structure. }
	\label{fig:backprop_autograd_net}
\end{subfigure}
\begin{subfigure}[t]{0.45\textwidth}
    \centering
    \includegraphics[width=\textwidth]{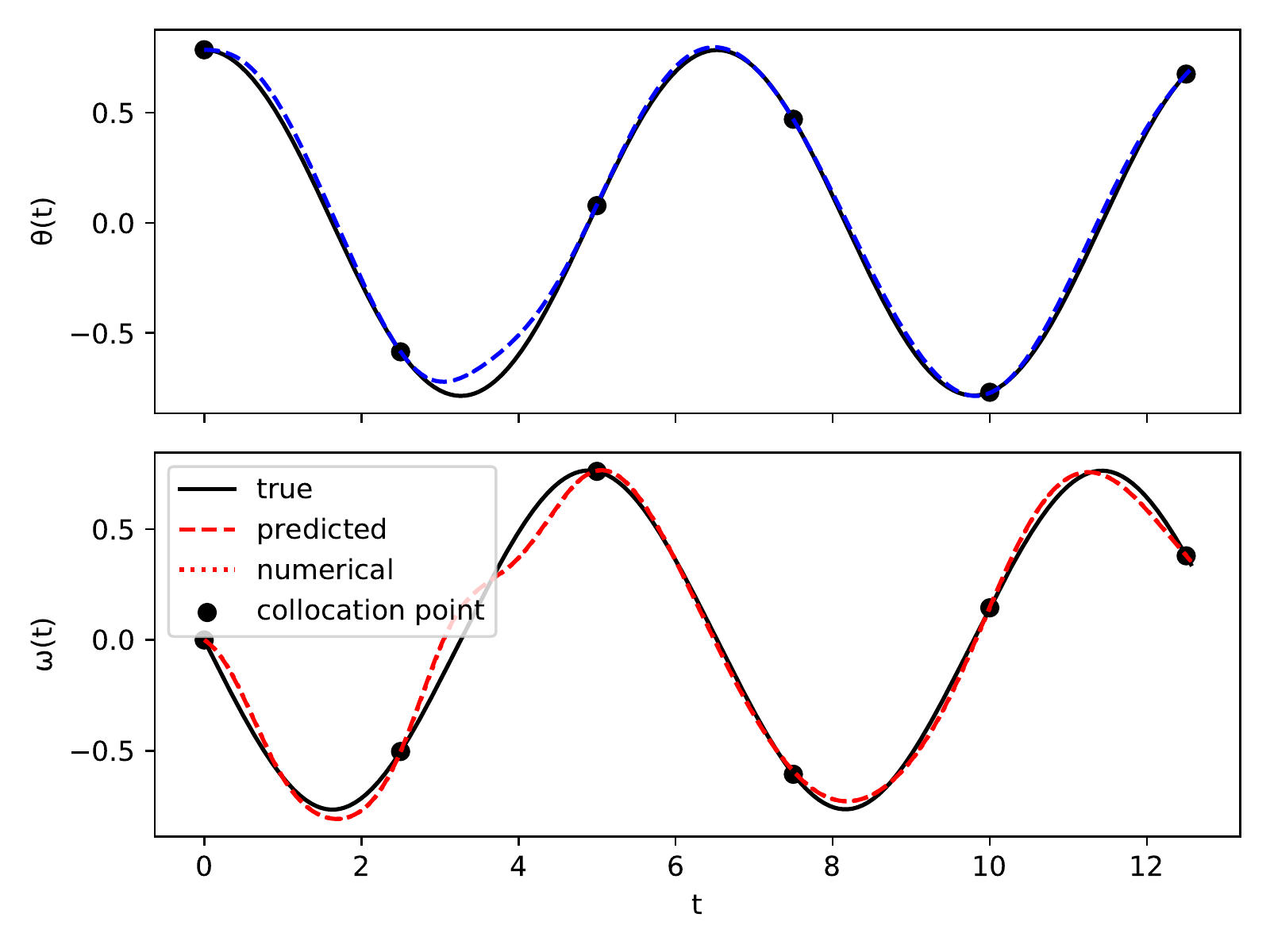}
    \caption{Predictions}
    \label{fig:prediction_continuous_model_autodiff}
\end{subfigure}

\begin{subfigure}[t]{.45\textwidth}
\begin{lstlisting}[
    mathescape=true
]
$\tilde{\theta}$ = network(t)
$\tilde{\omega}$ = gradient($\tilde{\theta}$, t)
loss = MSE(($\tilde{\theta}$, $\tilde{\omega}$), ($\theta$, $\omega$))
optimizer.step(loss)
\end{lstlisting}
\caption{Training.}
\label{lst:continuous_autodiff_training}
\end{subfigure}\hfill
\begin{subfigure}[t]{.45\textwidth}
\begin{lstlisting}[
    mathescape=true
]
$\tilde{\theta}$ = network(t)
$\tilde{\omega}$ = gradient($\tilde{\theta}$, t)
\end{lstlisting}
\caption{Inference.}
\label{lst:continuous_autodiff_inference}
\end{subfigure}
\caption{
Using automatic differentiation in direct-solution model.
The angular velocity is obtained by differentiating the angle with respect to time using automatic differentiation.
This approach ensures that an output, representing the derivative of another output, acts like a true derivative.
As a result, the network generalizes significantly better across both state variables.
}
\end{figure}

One way to leverage known relations is to calculate derivatives of state variables using automatic differentiation instead of having the network predict them as explicit outputs.
In the case of the pendulum, this means using the network to predict $\tilde{\theta}$ only and then obtaining $\tilde{\omega}$ by calculating the first-order derivative of $\tilde{\theta}$ with respect to time, as described in \cref{lst:continuous_autodiff_training} and~\ref{lst:continuous_autodiff_inference}.
Fig.~\ref{fig:prediction_continuous_model_autodiff} shows how much closer the predicted trajectories are to the true ones, when using this approach.

A drawback of obtaining $\omega$ using \emph{automatic differentiation}~(AD) is an increased computation cost and memory consumption depending on which mode of automatic differentiation is used.
Using reverse mode AD (backpropagation) as depicted in \cref{fig:backprop_autograd_net} requires another pass of the computation graph, as indicated by the arrow going from output $\theta$ to input $t$.
For training this is not problematic since the computations carried out during backpropagation are necessary to update the weights of the network as well.
However, using backpropagation during inference is not ideal because it introduces unnecessary memory and computation cost.
An alternative is to use forward AD which allows the derivatives to be computed during the forward pass, eliminating the need for a separate backwards pass.
Unfortunately, not all DL frameworks provide functions for evaluating the derivatives using forward mode AD~\cite{Baydin2018}[table 5].
A likely explanation for this is that the typical workload consists of evaluating the derivative of the loss with respect to the network's weights; a task where reverse-mode AD~(backpropagation) is more efficient.

\subsection{Physics-Informed Neural Networks}

For some modeling scenarios, equations describing the dynamics of the system are known, and using them to train the model is another way of addressing the data-sampling issue.
In what is known as \emph{physics-informed neural networks}~\cite{Raissi2019}, knowledge about the physical laws governing the system is used to impose structure on the NN model.
This can be accomplished through extending the loss function with an \emph{equation loss} term that ensures the solution obeys the dynamics described by the governing equations

\begin{equation}\label{eq:loss_pinn}
    L = L_{eq} + L_{c},
\end{equation}

\noindent where $L_{eq}$ penalizes inconsistencies with the governing equations, and $L_c$ penalizes differences between the predicted and true values (we refer to the set of true values as collocation points).
While this technique was originally proposed for solving PDEs, it can also be applied to solve ODEs.
For instance, to model the ideal pendulum using a PINN, we could integrate the expression of $d\omega$ from \cref{eq:ideal_pendulum} to formulate the loss as

\begin{equation*}
    L = \sum_{k=0}^{N-1} |\tilde{x}_k-x_k| + |\frac{\partial \tilde \omega}{\partial t} - \frac{g}{l}\sin{\tilde \theta_k}|.
\end{equation*}

\begin{figure}
\centering
\begin{subfigure}[t]{0.48\textwidth}
		\centering
		\includegraphics[width=\textwidth]{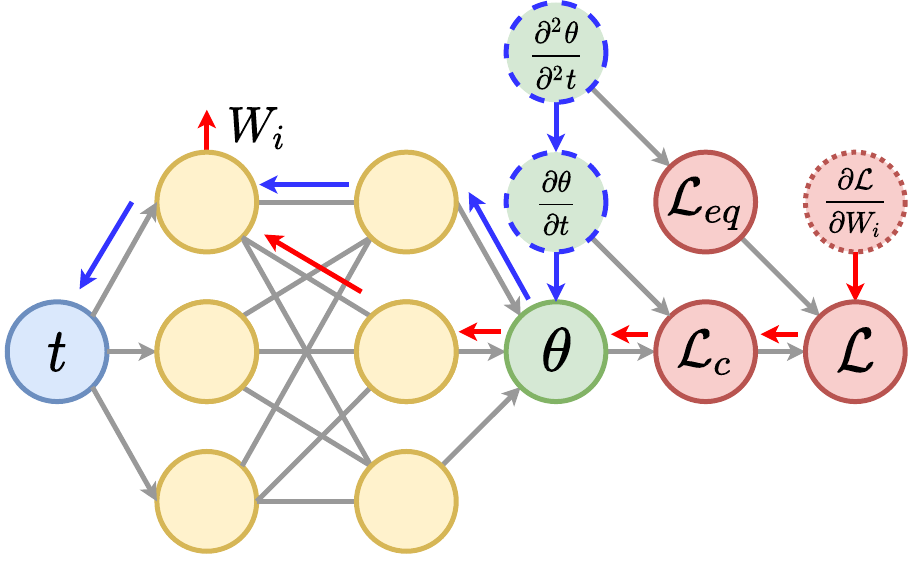}
		\caption{ Network structure. }
		\label{fig:pinn_backprop}
\end{subfigure}\hfill
\begin{subfigure}[t]{0.50\textwidth}
    \centering
    \includegraphics[width=\textwidth]{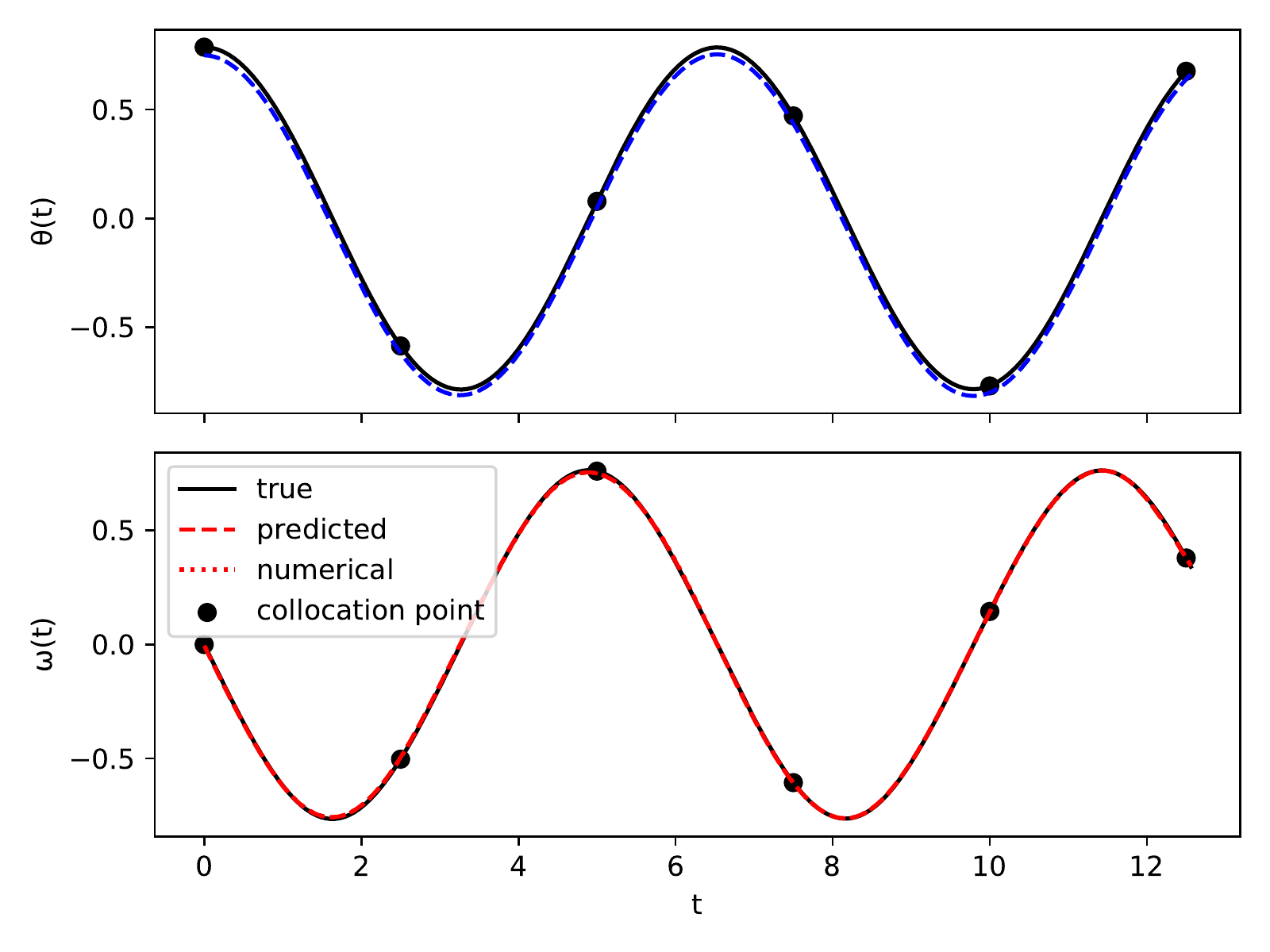}
    \caption{ Predictions.}
    \label{fig:prediction_continuous_model_pinn}
\end{subfigure}
\begin{subfigure}[t]{.45\textwidth}
\begin{lstlisting}[mathescape=true]
$\tilde{\theta}$ = network(t)
$\tilde{\omega}$ = gradient($\tilde{\theta}$, t)
loss_collocation=MSE(($\tilde{\theta}$, $\tilde{\omega}$), ($\theta$, $\omega$))
$\partial\tilde{\omega}$ = gradient($\tilde{\omega}$, t)
$\partial\tilde{\omega}_{eq}$ = -(g/l) * sin($\tilde{\theta}$)
loss_eq = MSE($\partial\tilde{\omega}$, $\partial\tilde{\omega}_{eq}$)
loss = loss_collocation + loss_eq
optimizer.step(loss)
\end{lstlisting}
\caption{Training.}
\label{lst:continuous_model_pinn_training}
\end{subfigure}\hfill
\begin{subfigure}[t]{.45\textwidth}
\begin{lstlisting}[mathescape=true]
$\tilde{\theta}$ = network(t)
$\tilde{\omega}$ = gradient($\tilde{\theta}$, t)
\end{lstlisting}
\caption{Inference.}
\label{lst:continuous_model_pinn_inference}
\end{subfigure}
\caption{ 
Physics-informed neural network. 
The network is trained to minimize the error in the collocation points and to penalize deviations from the equations governing the system.
}
\end{figure}

Again, we can use automatic differentiation to obtain $\frac{\partial \tilde \omega}{\partial t}$ by differentiating $\tilde \theta$ twice, depicted in the computation graph shown in \cref{fig:pinn_backprop}.
As shown in \cref{lst:continuous_model_pinn_training}, this requires only a few lines of code when using AD.

A motivation for incorporating the equation loss term is to constrain the search space of the optimizer to parameters that yield physically consistent solutions.
It should be noted that both the loss term penalizing the prediction error and the equation error are necessary to constrain the predictions of the network.
On its own, the equation error guarantees that the predicted state satisfies the ODE, but not necessarily that it is the solution at the particular time.
Introducing the prediction error ensures that the predictions are not only valid, but are also the correct solutions for the particular points used to calculate the prediction error.
Additionally, it should be noted that the collocation and equation loss terms may be evaluated for a different set of times.
For instance, the equation based loss term may be evaluated for an arbitrary number of time instances, since the term does rely on accessing the true solution for particular time instances.

In addition to proposing the introduction of the equation loss, PINNs also apply the idea of using backpropagation to calculate the derivatives of the state variables, rather than adding them as outputs to the network, as depicted in \cref{fig:pinn_backprop}.
Being able to obtain the n-th order derivatives is very useful for PINNs as they often appear in differential equations which the equation loss is based on. 
For the ideal pendulum, this technique can be used to obtain $\frac{\partial^2\theta}{\partial t^2}(t)$ from a single output of the network $\theta$, which can then be plugged into \cref{eq:ode} to check that the prediction is consistent.
A benefit of using backpropagation compared to adding state variables as outputs of the network is that this structurally ensures that the derivatives are in fact partial-derivatives of the state variables.

Training PINNs using gradient descent requires careful tuning of the learning rate. 
Specifically, it has been observed that the boundary conditions and the physics regularization terms may converge at different rates.
In some cases this manifests itself as a large misfit specifically at the boundary points.
The authors of \cite{Wang2020a, Wang2020} propose a strategy for weighing the different terms of the loss function to ensure consistent minimization across all terms. 

\subsection{Hidden Physics Networks}
\label{sec:cs_hidden}

\emph{Hidden physics neural networks}~(HNNs)~\cite{Raissi2020} can be seen as an extension of PINNs that use governing equations to extract features of the data that are not present in the original training data.
We refer to the unobserved variable of interest as a \emph{hidden variable}.
This technique is useful in cases where the hidden variable is difficult to measure compared to the known variables or simply impossible to measure since no sensor exists that can reliably measure it.

\begin{figure}[ht]
\centering
\begin{subfigure}[t]{0.45\textwidth}
	\centering
	\includegraphics[width=\textwidth]{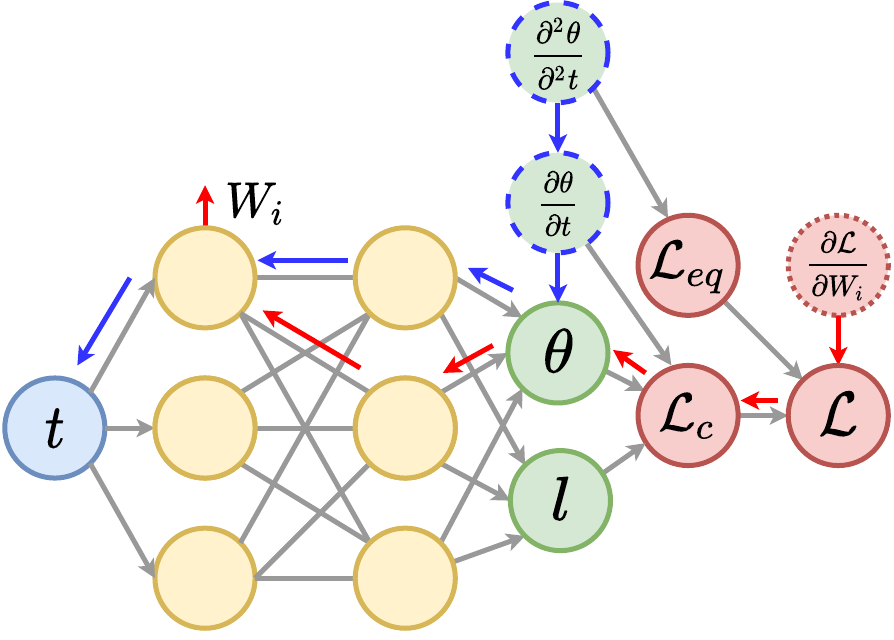}
	\caption{Network structure.}
	\label{fig:hnn}
\end{subfigure}\hfill
\begin{subfigure}[t]{0.44\textwidth}
    \centering
    \includegraphics[width=\textwidth]{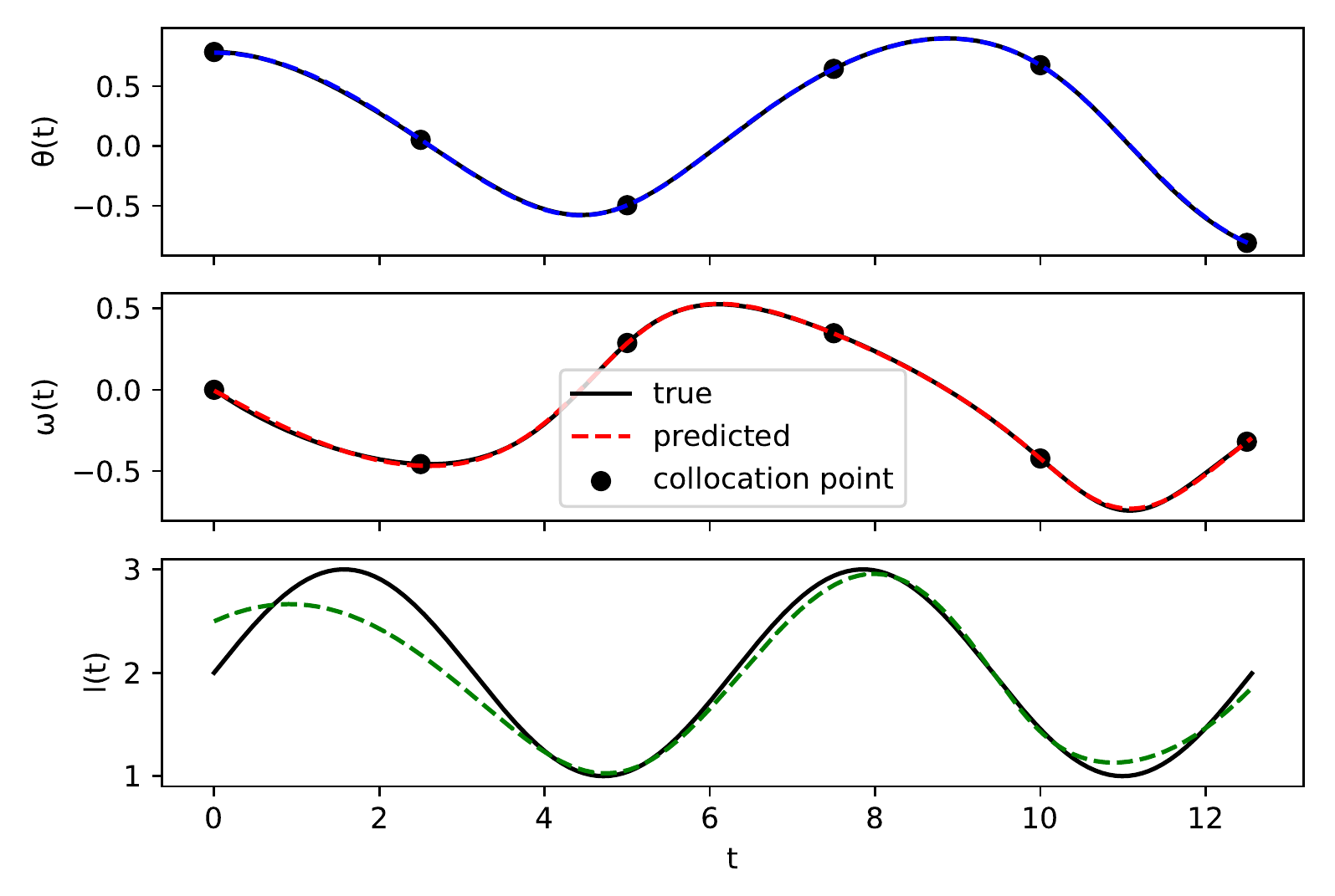}
    \caption{ Predictions.}
    \label{fig:prediction_continuous_model_hidden}
\end{subfigure}

\begin{subfigure}[t]{.45\textwidth}
\begin{lstlisting}[mathescape=true]
$\smash{\tilde{\theta}}$, $\smash{\tilde{l}}$ = network(t)
$\smash{\tilde{\omega}}$ = gradient($\smash{\tilde{\theta}}$, t)
loss_collocation=MSE(($\tilde{\theta}$, $\tilde{\omega}$), ($\theta$, $\omega$))
$\partial\tilde{\omega}$ = gradient($\tilde{\omega}$, t)
$\partial\tilde{\omega}_{eq}$ = -(g / $\tilde{l}$) * sin($\tilde{\theta}$)
loss_eq = MSE($\partial\tilde{\omega}$, $\partial\tilde{\omega}_{eq}$)
loss = loss_collocation + loss_eq
optimizer.step(loss)
\end{lstlisting}
\caption{Training.}
\label{lst:continuous_hidden_training}
\end{subfigure}\hfill
\begin{subfigure}[t]{.45\textwidth}
\begin{lstlisting}[mathescape=true]
$\smash{\tilde{\theta}}$, $\smash{\tilde{l}}$ = network(t)
$\tilde{\omega}$ = gradient($\smash{\tilde{\theta}}$, t)
\end{lstlisting}
\caption{Inference.}
\label{lst:continuous_hidden_inference}
\end{subfigure}
\caption{
Hidden-physics network. 
This network can be viewed as an extension of PINNs, which allows the network to predict physical quantities that are not available directly in the training data.
The example shows the network used to predict the length of the pendulum arm, $l$, which is set to vary in time for the sake of demonstration.
}
\end{figure}

For the sake of demonstration, we may suppose that the length of the pendulum arm is unknown and that it varies with time, as shown in \cref{fig:prediction_continuous_model_hidden}.
For the training this is problematic since $l$ is required to calculate the equation loss.
A solution to this is to add an output $\tilde{l}$ to the network that serves as an approximation of the true length $l$, as depicted in \cref{fig:hnn}.
The estimated value $\tilde{l}$ can then be plugged into the equation based loss term as shown in \cref{lst:continuous_hidden_training}.
It should be emphasized that $\tilde{l}$ is not part of the collocation loss term, since the true value $l$ is not known.
It is only as a result of the equation loss that the network is constrained to produce estimates of $l$ satisfies the system's dynamics.

The authors of~\cite{Raissi2020} use this technique to extract pressure and velocity fields based on measured dye concentrations.
In this particular case, the dye concentration can be measured by a camera, since the opacity of the fluid is proportional to the dye concentration.
They show that this technique also works well even in cases where the dye concentration is sampled at only a few points in time and in space.
Like PINNs, HNNs are easily applied to PDEs, but at the cost of the initial conditions being encoded in the network during training.

The difference between PINNs and HNNs is very subtle; both utilize similar network architectures and use loss functions that penalize any incorrect prediction violations of governing equations.
A distinguishing factor is that, in HNNs, the hidden variable is inferred based on physical laws that relate the hidden variable to the observed variables.
Since the hidden variables are not part of the training data, they can only be enforced through equations.

\section{Time-Stepper Models} 
\label{sec:time_stepper_models}

Consider the approach used to model an ideal pendulum, described in \cref{sec:background}.
First, a set of differential equations, \cref{eq:ode}, were used to model the derivative function of the system. 
Next, using the function, a numerical solver was used to obtain a simulation of the system for a particular initial condition.
The challenge of this approach is that identifying the derivative function analytically is difficult for complex systems.

\begin{figure}
\centering
\includegraphics[width=.75\columnwidth]{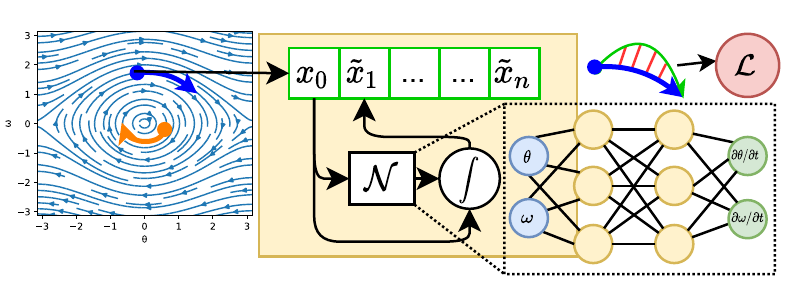}
\caption{
    Time-stepper model.
    Starting from a given initial condition $x_0$, the next state of the system $\tilde{x}_{k+1}$, is obtained by feeding the current state $\tilde{x}_k$ into the derivative network $\mathcal{N}$, producing a derivative that is integrated using an integration scheme $\int$.
    The loss $\mathcal{L}$ is evaluated by comparing the predicted with the training trajectory.
    The process can be repeated for multiple trajectories to improve the generalization of the derivative network.
}
\label{fig:ts_training_data}
\end{figure}

An alternative approach is to train a NN to approximate the derivative function of the system, allowing the network to be used in place of the hand derived function, as depicted in \cref{fig:ts_training_data}.
We refer to this type of model as a \emph{time-stepper model}, since it produces a simulation by taking multiple steps in time, like a numerical solver.
An advantage of this is that it allows well studied numerical solvers to be integrated into a model with relative ease.

The main differences between two given models can be attributed to (i) how the derivatives are produced by the network and (ii) what sort of integration scheme is applied.
For instance, the difference between the \emph{direct} (\cref{sec:ts_direct}) and \emph{Euler} time-stepper models (\cref{sec:ts:resnet}) is that the former does not employ any integration scheme, whereas the latter is similar to the Forward Euler (recall \cref{eq:fw_euler_solver}), leading to a significant difference in predictive ability.
Other networks, such as the \emph{Lagrangian} time-stepper, \cref{sec:ts_lagrangian}, distinguish themselves by the way the NN produces the derivatives.
Specifically, this approach does not obtain $\partial \theta$ and $\partial \omega$ as outputs from a network, but instead uses AD in an approach similar to \cref{sec:continuous_autodiff}. 
Similar to how an ODE can be solved with different numerical solvers, the Lagrangian time-stepper could be modified to use a different integration scheme than FE.

Given the independent relationship between the choice of NN and the numerical solver used, the models introduced in the sections should not be viewed as an exhaustive list of combinations.
Rather, the aim is to describe and compare the models commonly encountered in the literature.

\subsection{Methodology}

A natural question is how to train a time-stepper model.
Compared to the training of a direct-solution model, the training process of a time-stepper model must take several considerations into account.
First, a time-stepper must be able to produce accurate simulations for different initial conditions.
Second, the future predictions of a time-stepper depends on past predictions, which may lead to accumulation of error over multiple steps.

The first factor also influences the kind of data used to train a time-stepper model.
For example, several short trajectories, as shown in \cref{fig:ts_training_data}, may be used to train the network.
Equivalently, a few long trajectories may be captured and used for training.
In both cases, special care should be taken that the training data is representative of the data that can be encountered in the intended application.

The goal of training a time-stepper is to find a model that minimizes discrepancy between the predicted and the true state, for every point used for training.
A simple approach for doing so is to minimize the \emph{single-step error}
\begin{equation}\label{eq:train_step}
     L = \sum_{k=0}^{N-1} |\tilde{x}_k - x_k|,
\end{equation}
\noindent where $\tilde{x}_{k+1} = \tilde{x}_k + h_k*N(\tilde{x}_k)$ (here shown for FE solver) and $\tilde{x}_0=x_0$.
Minimizing the single-step error is just one of the potential ways to train a time stepper.

For the examples of time-steppers described in this section, we use the single-step criterion during training.
Each model is trained on 100 trajectories, each consisting of two samples; the initial state and the state one step into the future.
The initial states are sampled in the interval $\theta: (-1,1)$ and $\omega: (-1,1)$ using Latin hyper-cube sampling, see \cref{fig:ts_training_data}.
Each model uses a fully connected network consisting of 8 hidden layers with 32 neurons each.
Each layer of the network applies a softplus activation function.
The number of inputs and outputs are determined by the number of states characterizing the system, which is 2 for the ideal pendulum.
Exceptions to this are networks such as the Lagrangian network described in \cref{sec:ts_lagrangian}, for which the derivatives are obtained using automatic differentiation rather than as outputs of a network.

To validate the performance of each model, 100 new initial conditions are sampled in a grid.
For each initial condition in the validation set, the system is simulated for $4 \pi$ seconds using the original ODE and compared with the corresponding prediction made by the trained model.
For simplicity, we show only the trajectory corresponding to a single initial condition, like the one on \cref{fig:ts_direct_results}.

\subsection{Integration Schemes}

An important characteristic of a time-stepper model is how the derivatives are evaluated and integrated to obtain a simulation of the system.
Again, it should be emphasized that the choice of the numerical solver is independent of the architecture of the NN used to approximate the derivative function.
In other words, for a given choice of NN architecture, the performance of the trained model may depend on the choice of solver.

The choice of numerical solver not only determines how the model produces a simulation of the system, it also influences how the model must be trained.
Specifically, when minimizing any criterion that is a function of the integrated state, the choice of solver determines how the state is produced.

In the following subsection, we demonstrate how various numerical solvers can be used and evaluate their impacts on the performance of the models.

\subsubsection{Direct Time-Stepper}
\label{sec:ts_direct}

\begin{figure}[h]
    \begin{subfigure}[t]{0.45\textwidth}
        \centering
        \includegraphics[width=.9\textwidth]{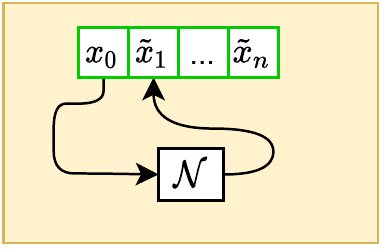}
        \caption{ Network structure. }
        \label{fig:time_stepper_direct}
    \end{subfigure}
    \hfill
    \begin{subfigure}[t]{0.45\textwidth}
        \centering
        \includegraphics[width=\textwidth]{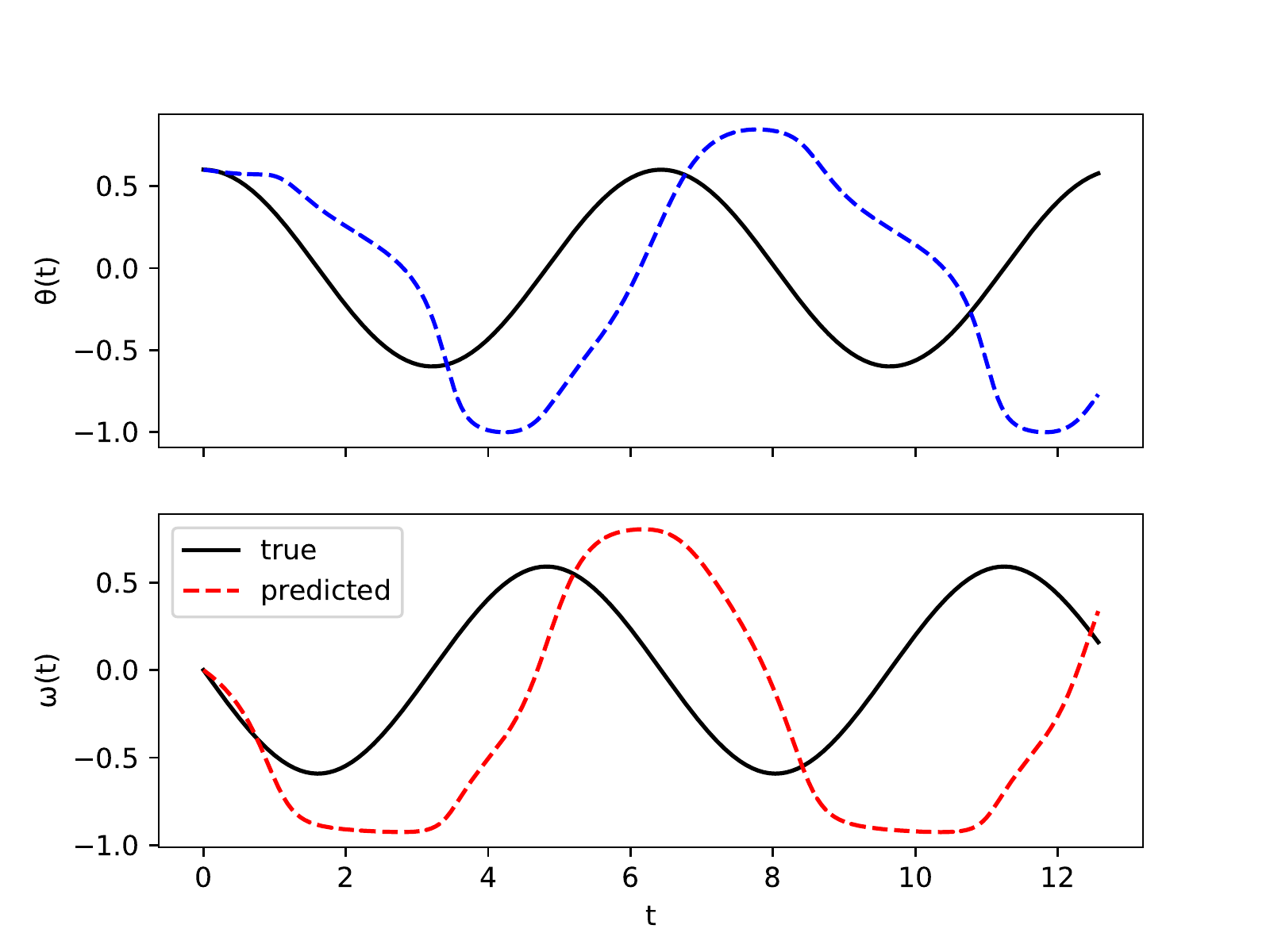}
        \caption{ Predictions. }
        \label{fig:ts_direct_results}
    \end{subfigure}
    \begin{subfigure}[t]{.45\textwidth}
\begin{lstlisting}[mathescape=true]
$\tilde x$ = network(x[0:end-1])
loss = MSE(x[1:end],$\tilde x$)
optimizer.step(loss)
\end{lstlisting}
\caption{Training.}
\label{lst:ts_direct_training}
\end{subfigure} \hfill
\begin{subfigure}[t]{.45\textwidth}
\begin{lstlisting}[mathescape=true]
$\tilde x[0]$ = $x_0$
for n in 0...N-1
  $\tilde{x}$[n+1] = network($\tilde x$[n])
\end{lstlisting}
\caption{Inference.}
\label{lst:ts_direct_inference}
\end{subfigure}
\caption{
Direct time-stepper. 
The output of the network is used as the prediction for the next step without any form of numerical integration.
An issue of this type of model is that it fails to generalize beyond the exact points in state-space that it has been trained for.
Over several steps, the error compounds, which leads to an inaccurate simulation.
}
\label{fig:ts_direct}
\end{figure}

The simplest approach of obtaining the next state is to use the prediction produced by the network directly, as summarized in \cref{fig:time_stepper_direct}:
\begin{equation*}
\label{eq:time_stepper}
	\tilde{x}_{k+1} = N(\tilde{x}_k),
\end{equation*}
where $N$ represents a generic neural network with arbitrary architecture and $\tilde{x}_0 = x_0$.

The network is trained to produce an estimate of the next state, $\tilde{x}_{k+1}$, from the current state, $x_k$.
During training, this operation can be vectorized such that every state at every timestamp, omitting the last, is mapped one step into the future using a single invocation of the network, as shown in \cref{lst:ts_direct_training}.
The reason for leaving out the last sample in when invoking the NN is that this would produce a prediction, $x_{N+1}$, for which there does not exist a sample in the training set.

At inference time, only the initial state $x_0$ is known.
The full trace of the system is obtained by repeatedly introducing the current state into the network, as depicted in \cref{lst:ts_direct_inference}.
Note that the inference phase cannot be parallelized in time, since predictions for time $k+1$ depend on predictions for time $k$.
However, it is possible to simulate the system for multiple initial states in parallel, as they are independent of each other.

The simulation for a single initial condition can be seen in \cref{fig:ts_direct_results}.
While, the simulation is accurate for the first few steps, it quickly diverges from the true dynamics.

\subsubsection{Residual Time-Stepper}
\label{sec:ts:resnet}

\begin{figure}
    \begin{subfigure}[t]{0.45\textwidth}
        \centering
        \includegraphics[width=.85\textwidth]{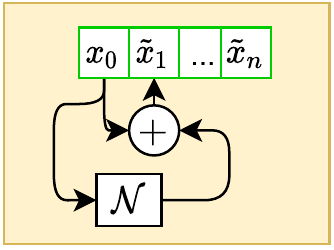}
        \caption{ Network structure. }
        \label{fig:time_stepper_resnet}
    \end{subfigure}
    \hfill
    \begin{subfigure}[t]{0.45\textwidth}
        \centering
        \includegraphics[width=\textwidth]{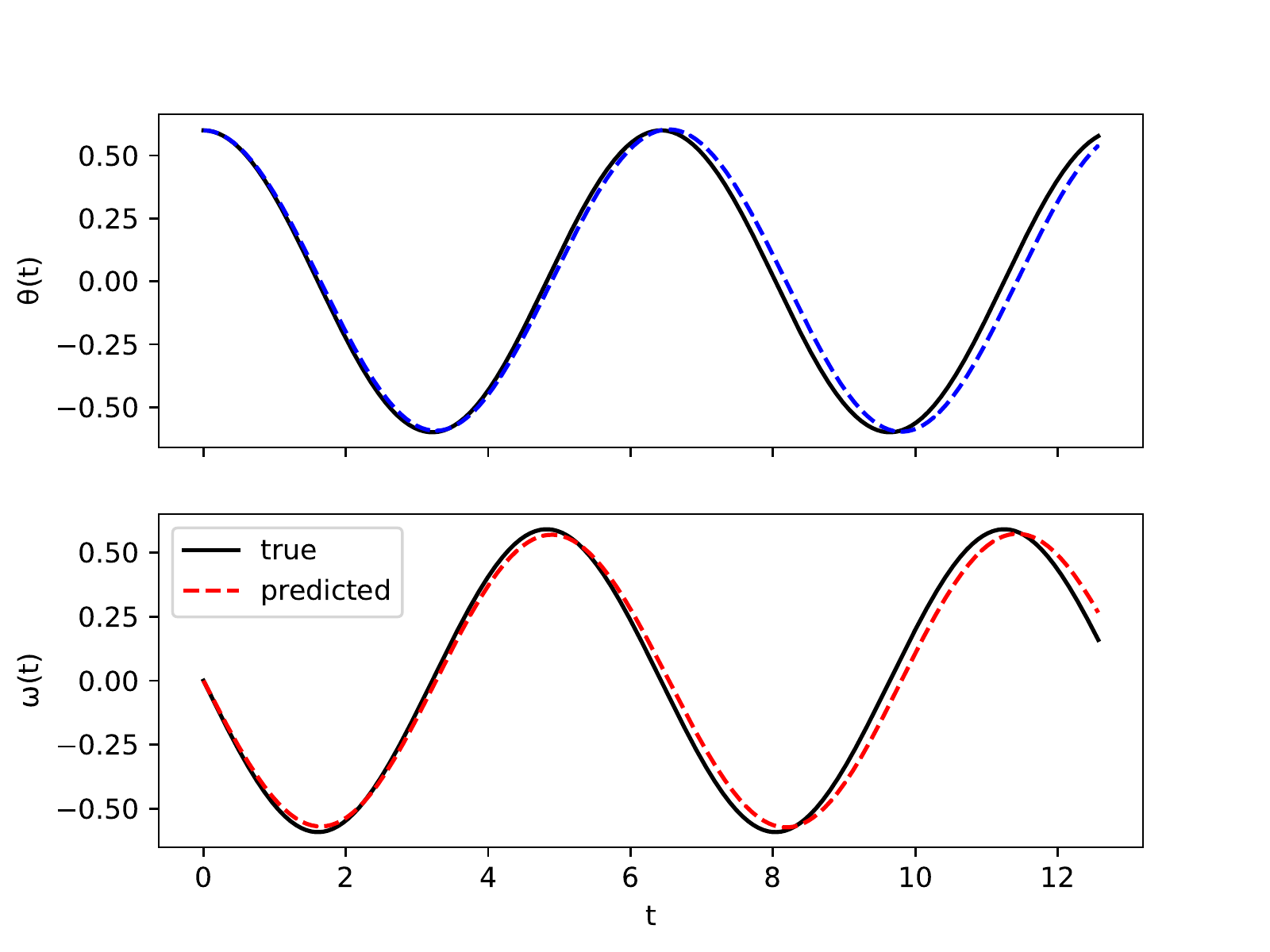}
        \caption{ Predictions. }
        \label{fig:true_vs_predicted_resnet}
    \end{subfigure}
\begin{subfigure}[t]{.45\textwidth}
\begin{lstlisting}[mathescape=true]
$\Delta x$ = network(x[0:end-1])
$\tilde x$ = x[1:end] + $\Delta x$
loss = MSE(x[0:end-1],$\tilde x$)
optimizer.step(loss)
\end{lstlisting}
\caption{Training.}
\label{lst:ts_resnet_training}
\end{subfigure} \hfill
\begin{subfigure}[t]{.45\textwidth}
\begin{lstlisting}[mathescape=true]
$\tilde x[0]$ = $x_0$
for n in 0...N-1
  $\Delta x$ = network($\tilde x$[n])
  $\tilde{x}$[n+1] = $\tilde{x}$[n] + $\Delta x$
\end{lstlisting}
\caption{Inference.}
\label{lst:ts_resnet_inference}
\end{subfigure}
\caption{
Residual time-stepper. 
The output of the network is added to the current state to form a prediction of the next state.
Compared to the direct time-stepper, this method produces simulations that are much closer to the true system.
}
\end{figure}

A network can be trained to predict a derivative like quantity which can then be added to the current state to yield the next as shown in \cref{fig:time_stepper_resnet}:
\begin{equation*} \label{eq:time_stepper_resnet}
    \tilde{x}_{k+1} = \tilde{x}_k + N(\tilde{x}_k).
\end{equation*}
DL practitioners may recognize this as a residual block that forms the basis for \emph{residual networks}~(ResNets)\cite{He2015} which are used with great success in applications spanning from image classification to natural language processing.
Readers familiar with numerical simulation will likely notice that the previous equation closely resembles the accumulated term in the forward Euler integrator (recall \cref{eq:fw_euler_solver}), but without the term that accounts for the step size.
If the data is sampled at equidistant time steps, the network scales the derivative to adapt the step size.

The central motivation for using a residual network is that it may be easier to train a network to predict how the system will change, rather than a direct mapping between the current and next state.

\subsubsection{Euler Time-Stepper}

\begin{figure}[h]
    \begin{subfigure}[t]{0.45\textwidth}
        \centering
        \includegraphics[width=.75\textwidth]{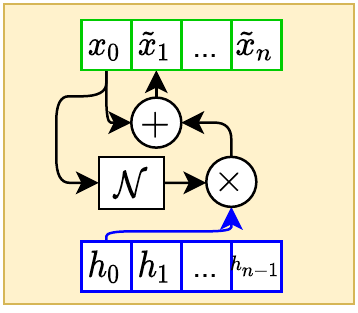}
        \caption{ Network structure. }
        \label{fig:time_stepper_euler}
    \end{subfigure}
    \hfill
    \begin{subfigure}[t]{0.45\textwidth}
        \centering
        \includegraphics[width=\textwidth]{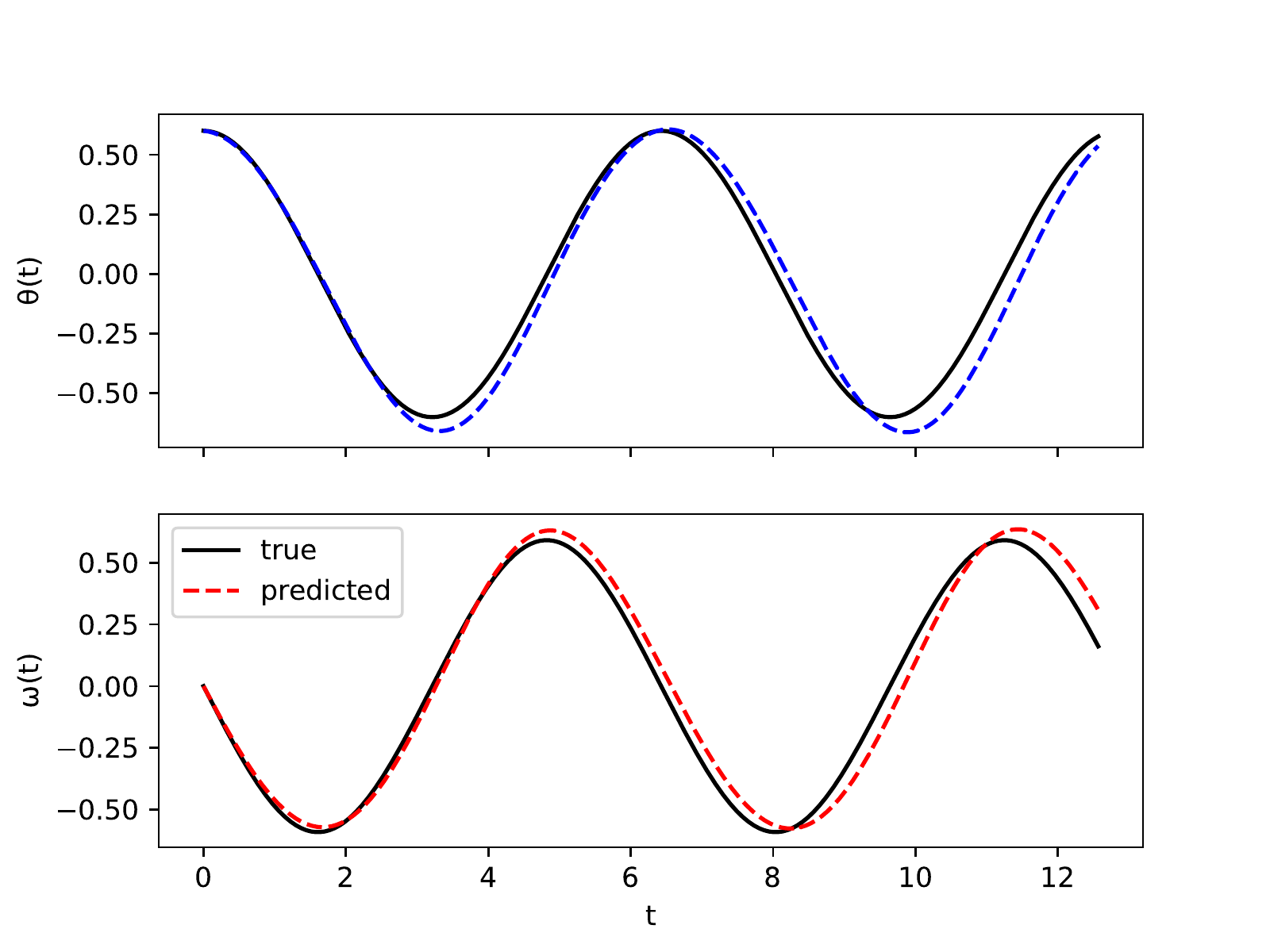}
        \caption{ Predictions. }
        \label{fig:true_vs_predicted_euler}
    \end{subfigure}
    
\begin{subfigure}[t]{.45\textwidth}
\begin{lstlisting}[mathescape=true]
$dx$ = network(x[0:end-1])
$\tilde x$ = x[0:end-1] + h*$dx$
loss = MSE(x[1:end],$\tilde x$)
optimizer.step(loss)
\end{lstlisting}
\caption{Training.}
\label{lst:ts_euler_training}
\end{subfigure} \hfill
\begin{subfigure}[t]{.45\textwidth}
\begin{lstlisting}[mathescape=true]
$\tilde x[0]$ = $x_0$
for n in 0...N-1
  $dx$ = network($\tilde x$[n])
  $\tilde{x}$[n+1] = $\tilde x$[n] + h[n]*$dx$
\end{lstlisting}
\caption{Inference.}
\label{lst:ts_euler_inference}
\end{subfigure}
\caption{
    Euler time-stepper. 
    The output of the network is multiplied by the step-size and is added to the current state to form a prediction for the next state.
    In this case accounting for the step-size leads to minimal improvements, if any, compared to the residual time-stepper.
    This is likely due to the fact that the step-size used during training is the same as the one used to plot the trajectory in~\cref{fig:true_vs_predicted_euler}. 
}
\end{figure}

Alternatively, the step-size can be encoded in the model by scaling the contribution of the derivative by the step size $h_k$ as shown in \cref{fig:time_stepper_euler}:
\begin{equation} \label{eq:time_stepper_euler}
    \tilde{x}_{k+1} = \tilde{x}_k + h_k*N(\tilde{x}_k).
\end{equation}
This resemblance has been noted several times \cite{Qin2019} and has resulted in work that interprets residual networks as ODEs allowing classical stability analysis to be used~\cite{Chang2018,Ruthotto2018,Ruthotto2020}.

The \emph{forward Euler}~(FE) integrator shown in~\cref{eq:time_stepper_euler} is simple to implement.
However, it accumulates a higher error than more advanced methods, such as the Midpoint, for a given step size.
This issue has motivated the integration of more sophisticated numerical solvers in time-stepper models.
For example, \emph{linear multistep}~(LMS) methods are used in~\cite{Raissi2018a}.
LMS uses several past states and their derivatives to predict the next state, resulting in a smaller error compared to FE.
Like FE, LMS only requires a single function evaluation per step, making it a very efficient method.
But if the system is not continuous, this method needs to be re-initialized after a discontinuity occurs~\cite{Gear1984b}.



\subsubsection{Neural Ordinary Differential Equations}

\emph{Neural ordinary differential equations}~(NODEs)~\cite{Chen2019} is a method used to construct models by combining a numerical solver with a NN that approximates the derivative of the system.
Unlike the previously introduced models, the term NODEs is not used to refer to models using a specific integration scheme, but rather to the idea of treating an ML problem as a dynamical system that can be solved using a numerical solver.

\begin{figure}
\centering
\includegraphics[width=.6\textwidth]{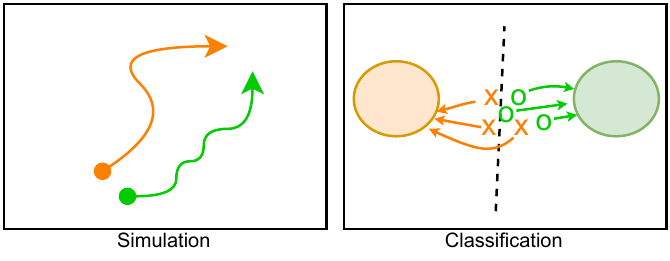}
\caption{
    Different applications of NODEs. 
    NODEs can be used to simulate a dynamical system with the goal of obtaining a trajectory corresponding to an initial condition.
    In this case, the goal is to train the network to produce a derivative that provides a good estimate of the true state at every step of the trajectory.
    Another use is for classification by treating each input sample as a point in state-space, which evolves according to the derivative produced by the network.
    In this case, the goal is to train the network to learn dynamics that leads to samples belonging to each class ending in distinct clusters that are easily separable.
}    
\label{fig:nodes_classification_vs_simulation}
\end{figure}

Some confusion may arise from the fact that NODEs are frequently used for image classification throughout the literature, which may seem completely unrelated to numerical simulations.
The underlying idea is that an image can be represented as a point in state-space which moves on a trajectory defined by an ODE, as shown in \cref{fig:nodes_classification_vs_simulation}.
The goal of this is to find an ODE that results in images of the same class converging to a cluster that is easily separable from that of unrelated classes.
For single inference, e.g. in image classification, intermediate predictions have no inherent meaning, i.e. they typically do not correspond to any measurable quantity of the system; we are only interested in the final estimate $\hat{x}_n$.
Due to the lack of training samples corresponding to intermediate steps, it is impossible to minimize the single step error.

\begin{figure}[h]
    \begin{subfigure}[t]{0.45\textwidth}
        \centering
        \includegraphics[width=\textwidth]{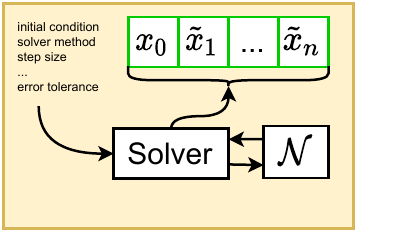}
        \caption{Network Structure.}
        \label{fig:model_node}
    \end{subfigure}
    \begin{subfigure}[t]{0.45\textwidth}
        \centering
        \includegraphics[width=\textwidth]{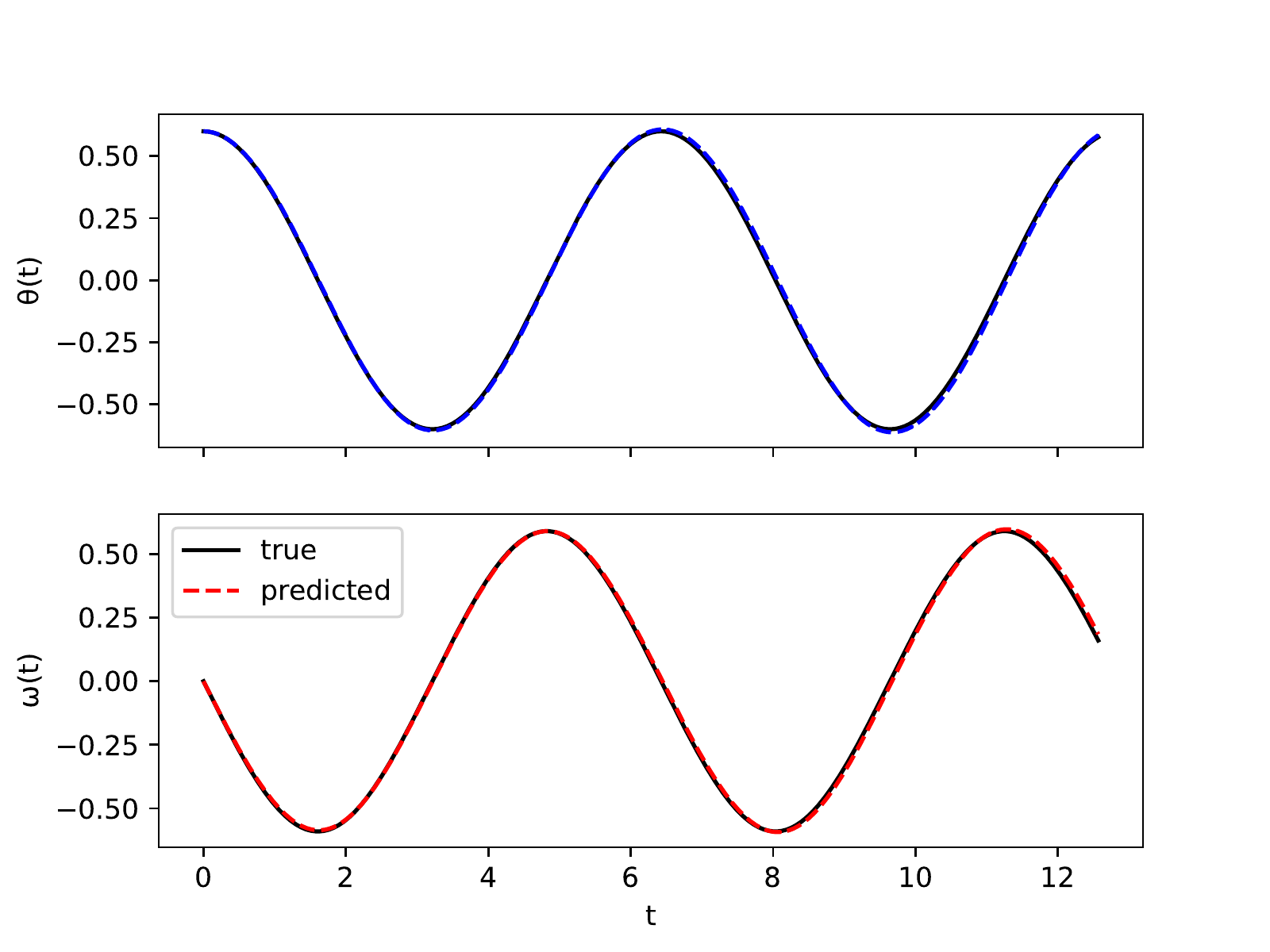}
        \caption{Predictions.}
        \label{fig:true_vs_predicted_node}
    \end{subfigure}
    
\begin{subfigure}[t]{.45\textwidth}
\begin{lstlisting}[mathescape=true]
$\tilde x$ = odeint(net,$x_0$,$t_{start}$,$t_{start}+h$,"rk4")
loss = MSE(x,$\tilde x$)
optimizer.step(loss)
\end{lstlisting}
\caption{Training.}
\end{subfigure} \hfill
\begin{subfigure}[t]{.45\textwidth}
\begin{lstlisting}[mathescape=true]
$\tilde x$ = odeint(net,$x_0$,$t_{start}$,$t_{end}$,"rk4")
\end{lstlisting}
\caption{Inference.}
\label{lst:ts_node_inference}
\end{subfigure}
\caption{
    Neural ordinary differential equations. 
    Neural ODEs generally refer to models that are constructed to use a numerical solver to integrate the derivatives through time.
    Unlike the previously introduced integration schemes which mapped to concrete architectures, neural ODEs refer to the idea of using well established numerical solvers inside a model.
    Part of neural ODEs popularity is due to the fact that it mimics the programming APIs of traditional numerical solvers, which makes it easy to switch between different types of solvers.
}
\end{figure}

The authors of~\cite{Chen2019} motivate the use of an adaptive-step size solver by its ability to adjust the step-size to match a desired balance between numerical error and performance.
An alternative way to view NODEs is as a \emph{continuous-depth model} where the number of layers is a result of the step-size chosen by the solver.

From this perspective, stability of NODEs is closely related to the stability of integration schemes of classical ODEs.
To address the convergence issues during training, some authors propose NODEs with stability guarantees by exploiting Lyapunov stability theory~\cite{massaroli2020stable} and spectral projections~\cite{quaglino2020snode}.
Another standing issue of NODEs is their large computational overhead during training compared to classical NNs.
Authors in~\cite{finlay2020train} demonstrated that stability regularization may improve convergence and reduce the training times of NODEs.
\cite{poli2020graph} proposes graph NODEs resulting in training speedups, as well as improved performance due to incorporation of prior knowledge.

To improve the performance, others have introduced various inductive biases such as Hamiltonian NODE architecture~\cite{SymODEnet2019}, or penalizing higher order derivatives of the NODEs in the loss function~\cite{kelly2020learning}.
To account for the noise and uncertainties, some authors proposed stochastic NODEs~\cite{liu2019neural,JumpStochNODEs2019,guler2019robust,li2020scalable} as generalizations of deterministic NODEs.

A fundamental issue of interpreting trained NODEs as a proper ODEs is that they may have trajectory crossings, and their performance can be sensitive to the size-size used during inference~\cite{Ott2020}.
Contrary to this, the solutions of ODEs with unique solutions would never have intersecting trajectories, as this would imply that, for a given state (the point of intersection), the system could evolve in two different ways.
Some authors have noted that there seems to be a critical step-size for which the trained network starts behaving like a proper ODE~\cite{Ott2020}. 
That is, if trained with the particular step-size, the network will perform equally well or better if used with a smaller step-size during inference.
Another approach is to use regularization to constrain the parameters of the network to ensure that solutions are unique.
For ResNets this can be achieved by ensuring that the Lipschitz constant of the network to be less than $1$ for any point in the state-space, which guarantees that a unique solution~\cite{Behrmann2019}.

To deal with external inputs in NODEs, the authors of~\cite{dupont2019augmented,norcliffe2020second} propose lifting the state space via additional augmented variables. 
A more general way of explicitly modeling the input dynamics via additional NNs is proposed by~\cite{massaroli2021dissecting}.

\subsection{External Input}

\begin{figure}
    \centering
    \begin{subfigure}[t]{0.35\textwidth}
        \centering
        \includegraphics[width=\columnwidth]{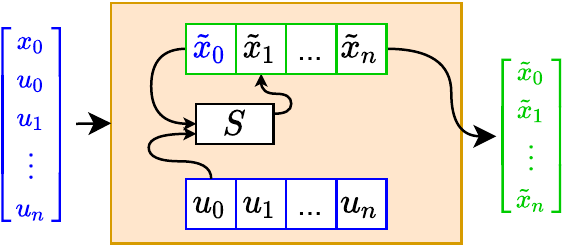}
        \caption{ State and input stacked and fed into the same network. }
        \label{fig:time_stepper_inputs_augment}
    \end{subfigure}
    \hspace{1cm}
    \begin{subfigure}[t]{0.35\textwidth}
        \centering
        \includegraphics[width=\columnwidth]{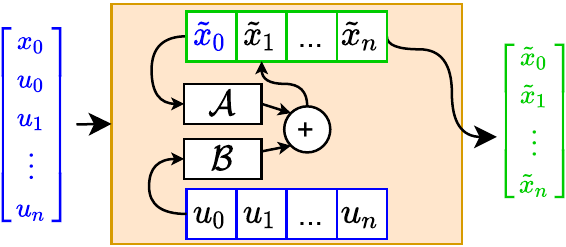}
        \caption{ State and input fed into separate networks. }
        \label{fig:time_stepper_inputs_separate}
    \end{subfigure}
    \caption{ Incorporation of inputs in time stepping model. }
    \label{fig:time_stepper_inputs}
\end{figure}

So far, we have only considered how to apply time-stepper models to systems where the derivative function is determined exclusively by the system's state.
In practice, many systems encountered are influenced by an external stimulus that is independent of the dynamics, such as external forces acting on the system or actuation signals of a controller.
To avoid confusion, we refer to these external influences as \emph{external input} to distinguish it from the general concept of a NN's inputs.

The structure of a time-stepper model lends itself well to introducing external inputs at every evaluation of the derivative function.
As a result, it is possible to integrate external inputs in time-stepper models in many ways.

\subsubsection{Neural State-Space Models}
\label{sec:external_inputs}

Inputs can be added to the time-stepper models in a couple of ways.
One way is to concatenate the inputs with the states, as illustrated in \cref{fig:time_stepper_inputs_augment}:
\begin{equation}
\label{eq:SSM}
    x_{k+1} = \mathcal{N}([x_k, u_k]),
\end{equation}
where $x_k$ and $u_k$ represent states and inputs at time $t_k$, respectively. 
The evolution of the future state $ x_{k+1}$ is fully determined by the derivative network $\mathcal{N}$.
A possible rationale for lumping system states and inputs are parameter-varying systems, where the inputs influence the system differently depending on the current state.
This approach does not impose any structure on how the state and input information are aggregated in the network, since the layers of the network make no distinction between the two.

Alternatively, two separate networks $\mathcal{A}$ and $\mathcal{B}$ can be used to model contributions of the autonomous and forced parts of the dynamics, respectively, as seen in \cref{fig:time_stepper_inputs_separate}.
This information can then be aggregated by taking the sum of the two terms:
\begin{equation}
\label{eq:SSM_block}
    x_{k+1} = \mathcal{A}(x_k) + \mathcal{B}(u_k).
\end{equation}
This approach is suitable for systems where the influence of the inputs is known to be independent of the state of the system, since it structurally enforces models that are independent.

In system identification and control theory, both variants~\eqref{eq:SSM} and~\eqref{eq:SSM_block} are referred to as \emph{state-space models}~(SSM)~\cite{KROLL2014496,Schoukens2019,SCHOUKENS2017446,KERSCHEN2006505}.
More recently, researchers~\cite{krishnan2016structured,LatentDynamics2018,NIPS2018_8004,SUK2016292} proposed to model non-linear SSMs by using NNs, which we refer to a \emph{neural state-space models}~(NSSM).

Some works proposed to combine neural approximations with classical approaches with linear state transition dynamics $\mathcal{A}$, resulting in Hammerstein~\cite{OgunmoluGJG16}, and Hammerstein-Wiener architectures~\cite{HW_RNN2008}, or
using linear operators representing transfer function as layers in deep NNs~\cite{forgione2020dynonet}.
While, others leverage encoder-decoder neural architectures to handle partially observable dynamics~\citep{gedon2020deep,MastiCDC2018}.
Authors in~\cite{skomski2021constrained,skomski2020physicsinformed,drgona2020physicsconstrained} applied principles of gray-box modeling by imposing 
physics-informed constraints on learned neural SSM.
The authors of \cite{ogunmolu_nonlinear_2016} analyzed the effect of different neural architectures on the system identification performance of non-linear systems, and concluded that, compared to classical non-linear regressive models, deep neural networks scale better and are easier to train.

\subsubsection{Neural ODEs with External Input}

The challenge of introducing external input to NODEs is that the numerical solver may try to evaluate the derivative function at time instances that align with the sampled values of the external input.
For instance, an adaptive step-size solver may choose its own internal step-size based on how rapidly the derivative function changes in the neighborhood of the current state.
The issue can be solved using interpolation to obtain values of external inputs for time instances that do not coincide with the sampling.

External input can also be used to represent static parameter values that remain constant through a simulation.
In the context of the ideal pendulum system, we could imagine that the length of the pendulum could be made a parameter of the model, allowing the model to simulate the system under different conditions.
The authors of~\cite{Lee2021} calls this approach \emph{parameterized} NODEs, and use this mechanism to train models that can solve PDEs for different parameter values.

Another approach is  \emph{neural controlled differential equations}~(NCDEs)~\cite{Kidger2020}.
The term "controlled" should not be confused with the field of \emph{control theory}, but rather the mathematical concept of controlled differential equations from the field of rough analysis.
The core idea of NCDEs is to treat the progression of time and the external inputs as a signal that \emph{drives} the evolution of the system's state over time.
The way that a specific system responds to this signal is approximated using a NN. 
A benefit of this approach is that it generalizes how a system's autonomous and forced dynamics are modelled.
Specifically, it allows NCDEs to be applied to systems where NODEs would be applied, as well as systems where the output is purely driven by the external input to the system.

\subsection{Network Architecture}
Part of the success of NNs can be attributed to the ease of integrating specialized architectures into a model.
In this section, we introduce a few examples of how to integrate domain specific NNs into a time-stepper model.

First, \cref{sec:ts_lagrangian} describes how energy conserving dynamics can be enforced by encoding the problem using Hamiltonian or Lagrangian mechanics.
Next, \cref{sec:ts_potential} demonstrates another way of enforcing energy conservation, which is often encountered in molecular dynamics.
Finally, \cref{sec:ts_graph} describes how graph neural can be integrated in a time-stepper to solve problems which can be naturally be represented as graphs.

\subsubsection{Hamiltonian and Lagrangian Networks} 
\label{sec:ts_lagrangian}

\begin{wrapfigure}{R}{0.45\textwidth}
\centering
\includegraphics[width=.40\textwidth]{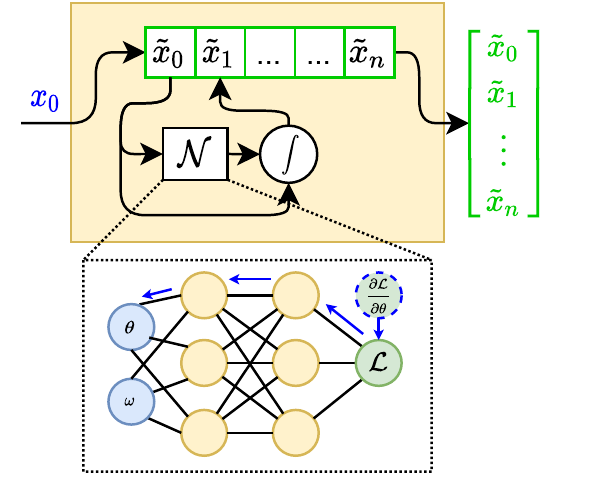}
\caption{ 
    Lagrangian time-stepper. 
    The Lagrangian, $\mathcal{L}$~(not to be confused with the loss function), is differentiated using AD to obtain the derivative of the state.
}
\label{fig:time_stepper_lagrangian}
\end{wrapfigure}

Recall that the movement in some physical systems happens as a result of energy transfers within the system, as opposed to systems where energy is transferred to/from the system.
The former is called an energy conservative system.
For instance, if the pendulum introduced in \cref{fig:cart_pendulum_diagram} had no friction and no external forces acting on it, it would oscillate forever, with its kinetic and potential energy oscillating without a change in its total energy.
In physics, a special class of closely related functions has been developed for describing a total energy of a system called Hamiltonian and Lagrangian functions.
Both Hamiltonian $ \mathcal{H}$ and Lagrangian $ \mathcal{L}$  are defined as a sum of total kinetic $T$ and potential energy $V$ of the system. 
We start with the Hamiltonian defined as
\begin{equation}
    \label{eq:hamiltonian}
       \mathcal{H}(x) = T(x) - V(x),
\end{equation}
\noindent where $x = [q, p]$ represents the concatenated state vector of generalized coordinates $q$ and generalized momenta $p$.
By taking the gradients of the energy function~   \eqref{eq:hamiltonian}, we can derive a corresponding 
differential $ \dot{x} =f(x)$  equation as
\begin{equation}
    \label{eq:energy_timestep}
        \dot{x} = S \nabla \mathcal{H}(x),
\end{equation}
where $S$ is a symplectic matrix. 
Please note that the difference between  $\mathcal{H}$  and $ \mathcal{L}$ is their corresponding coordinate system: for the Lagrangian, instead of $x = [q, p]$, we consider $x = [q, \dot{q}]$, where  $\dot{p} = M(q) \dot{q}$, with $M(q)$ being a generalized mass matrix. 

Despite their mathematical elegance, deriving analytical Hamiltonian and Lagrangian functions for complex dynamical systems is a grueling task. 
In recent years, the research community turned its attention to deriving these types of scalar valued energy functions by means of data-driven methods~\cite{Lutter2019,Greydanus2019,SymODEnet2019}.
Specifically, the goal is to train a neural network to approximate the Hamiltonian/Lagrangian of the system, as shown in \cref{fig:time_stepper_lagrangian}.
A key aspect of this approach is that the derivatives of the states are not outputs of the network, but are instead obtained by differentiating the output of the network~$\mathcal{L}$, with respect to the state variables~$[\theta,\omega]$ and plugging the results into~\cref{eq:hamiltonian}.
The main advantage of Hamiltonian~\cite{Greydanus2019,Toth2020} NNs and the closely related Lagrangian~\cite{Cranmer2020,Lutter2019} NNs, is that they naturally incorporate the preservation of energy into the network structure itself.
Research into simulation of energy preserving systems has yielded a special class of solvers, called symplectic solvers.
The authors of~\cite{SympNets2020} propose a new specialized network architecture, referred to as \emph{symplectic networks}, to ensure that the dynamics of the model are energy conserving.
Similarly, the authors of~\cite{Finzi2020} propose extensions for including explicit constraints via Lagrange multipliers for improved training efficiency and accuracy.

\subsubsection{Deep Potential Energy Networks}
\label{sec:ts_potential}

A similar concept to that of \emph{Hamiltonian} and
\emph{Lagrangian} neural networks involves learning neural surrogates for potential energy functions $V(x)$ of a dynamical system, where the primary difference with \emph{Hamiltonians} and
\emph{Lagrangians} is that the kinetic terms are encoded explicitly in the time stepper by considering classical Newtonian laws of motion:
\begin{subequations}
    \label{eq:potential_timestep}
    \begin{align}
        x_{k+1} & = x_{k} + v_{k}, \\
        v_{k+1} & = v_{k} -  \frac{\nabla \mathcal{V}(x)}{m},
    \end{align}
\end{subequations}
where $x_{k}$, and $v_{k}$ are  positional and velocity vectors of the system. The gradients of the potential function are equal to the interaction forces $F = - \nabla \mathcal{V}(x)$, while $m$ being a vector of ``masses''. 

This approach is extensively used, mainly in the domain of molecular dynamics (MD) simulations~\cite{Behler2015,Jiang2016,Wang2019,Unke2018,PhysNet2019,PhysRevLett143001}.
In modern data-driven MD, the learned neural potentials $V(x)$ replace expensive 
quantum chemistry calculations based, e.g., on density functional theory (DFT).
The  advantage of this approach for large-scale systems, compared to directly learning high-dimensional maps of the time steppers, is that the
learning of the scalar valued potential function $V(x): \mathbf{R}^n \to \mathbf{R}$  represents a much simpler regression problem.
Furthermore, this approach allows prior information to be encoded in the architecture of the 
deep potential functions $V(x)$, such as considering only local interactions between atoms~\cite{schutt2017schnet}, and encoding spatial symmetries~\cite{TorchANI2020,zhang2018endtoend}.
As a result, these methods are allowing researchers in MD to achieve unprecedented scalability, allowing simulation of up to 100M atoms on supercomputers~\cite{jia2020pushing}.
In contrast, training a single naive time stepper for such a model would require learning a 300M dimensional mapping.  

\subsubsection{Graph Time-Steppers}
\label{sec:ts_graph}

\begin{wrapfigure}{R}{0.45\textwidth}
\centering
\includegraphics[width=.40\textwidth]{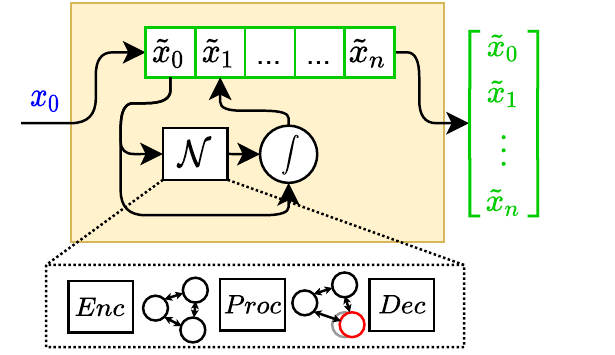}
\caption{
    Simplified view of a graph time-stepper. 
    During each step of the simulation, the current state is encoded as a graph~(Enc) which is then used to compute the change in state variable between the current and next time step~(Proc).
    Finally, the change in state is decoded to the original state-space to update the state of the system~(Dec).
}
\label{fig:time_stepper_graph}
\end{wrapfigure}

Many complex real-world systems from social networks, molecules, to power grid systems can be represented as graph structure describing the interactions between individual subsystems.
Recent research in \emph{graph neural networks}~(GNNs) embraces this idea by embedding or learning the underlying graph structure from data. 
There exists a large body of work on GNNs, but covering this is outside the scope of this survey. 
We refer the interested reader to overview papers~\cite{Bronstein2021,Battaglia2021,ZHOU202057,GNN_survey2019,Zhang2015,Scarselli2009}. 
For the purposes of this section, we focus solely on GNN-based time stepper models applied to modeling of dynamical systems~\cite{kipf2018neural,Yunzhu2018}.

The core idea of using GNNs inside time-steppers is to use a GNN-based pipeline to estimate the derivatives of the system, as shown in \cref{fig:time_stepper_graph}.
Generally, the pipeline can be split into 3 steps; first the current state of the system is encoded as a graph, next the graph is processed to produce an update of the system's state, finally the update is decoded and used to update the state of the system.

One of the early works includes interaction networks~\cite{BattagliaPLRK16} or \emph{neural physics engine} (NPE)~\cite{ChangUTT16} demonstrating the ability to learn the dynamics in various physical domains in smaller scale dimensions, such as n-body problems, rigid-body collision, and non-rigid dynamics.
Since then, the use of GNNs rapidly expanded, finding its use in neural ODE time steppers~\cite{Gonzalez2019} including control inputs~\cite{GNNs_control2018,propagation2018}, dynamic graphs~\cite{DynamicGraphNN2020}, or considering feature encoders enabling learning dynamics directly from the visual signals~\cite{NIPS2017_8cbd005a}.
Modern GNNs are trained using message passing (MP) algorithms introduced in the context of quantum chemistry application~\cite{GilmerSRVD17}.
In GNNs, each node has associated latent variables representing values of physical quantities such as positions, charges, or velocities, then in the MP step, the aggregated values of the latent states are passed through the edges to update the values of the neighboring nodes. 
This abstraction efficiently encodes local structure-preserving interactions that commonly occur in the natural world.
While early implementations of GNN-based time steppers suffered from larger computational complexity, more recent works~\cite{Gonzalez2020} have demonstrated their scalability to ever larger dynamical systems with thousands of state variables over long prediction horizons.
Due to their expressiveness and generic nature, GNNs could in principle be applied in all the time-stepper variants summarized in this manuscript, some of which would represent novel architectures up to date.

\subsection{Uncertainty} 
\label{sec:ts_prob}
So far, we have considered only the cases of modeling systems where noise-free trajectories were available for training.
In reality, it is likely that the data captured from the system does not represent the true state of the system, $x$, but rather a noisy version of the original signal perturbed by measurement noise.
Another source of uncertainty is that the dynamics of the system itself may exhibit some degree of randomness.
One cause of this would be unidentified external forces acting on the system.
For instance, the dynamics by a physical pendulum may be influenced by vibrations from its environment.
The following subsections introduce several models that explicitly incorporate uncertainty in their predictions.

\subsubsection{Deep Markov Models}
\label{sec:ts_dtmc}

\begin{figure}[t]
\includegraphics[width=.5\textwidth]{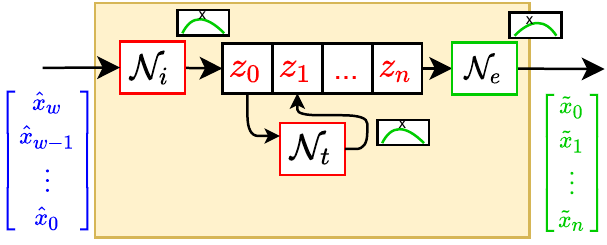}
\caption{
    Deep Markov model with inference network. 
    The value of $z_0$ is estimated by an inference network $N_i$ based on several samples of the observed variable.
    The transmission function, approximated by the network $N_t$, maps the current value of $z$ to a distribution over $z$ one step ahead in time.
    The emission function, approximated by $N_e$, maps each predicted latent variable to a distribution of the corresponding $x$ value in the original observed space.
    Note that the output of each network is the parameters of a distribution, which is then sampled to obtain a value that can be fed into the next stage of the model.
}
\label{fig:time_stepper_markov}
\end{figure}

A \emph{deep Markov model}~(DMM)~\cite{Krishnan2016,mustafa2019comparative,awiszus2018markov,liu2019powering,fraccaro2016sequential} is a probabilistic model that combines the formalism of Markov chains with the use of NN for approximating unknown probability density functions.
A Markov chain is a latent variable model, which assumes that the values we observe from the system are determined by an underlying latent variable, which can not be observed directly.
This idea is very similar to an SSM, the difference being that a Markov chain assumes that the mapping from the latent to the observed variable is probabilistic and that evolution of the latent variable is not fully deterministic.

The relationship between the observed and latent variables of a DMM, can be specified as:
\begin{subequations}
    \label{eq:DMM}
\begin{align}\label{eq:transition}
    {z}_{k+1} & \sim \mathcal{Z}(N_{t}({z}_k)) & \text{(Transition)}\\
    {x}_{k} & \sim \mathcal{X} (N_{e}({z}_k)) & \text{(Emission)}
\end{align}
\end{subequations}
\noindent where ${z}_{k}$ represents the latent state vector, and ${{x}}_{k}$ is the output vector.
Here, $\mathcal{Z}$ and $\mathcal{X}$ represent probability distributions, commonly Gaussian distributions, modeled by maps $N_{T}(\mathbf{z}_k)$ or $N_{e}(\mathbf{z}_{k})$, respectively.

A natural question to ask is how the observed and latent variables are represented, given that they are probability density functions and not numerical values.
A solution to pick distributions that can be represented in terms of a few characteristic parameters.
For instance, a Gaussian can be represented by its mean and covariance.
The process of performing inference using a DMM is shown in \cref{fig:time_stepper_markov}.

An obstacle to training DMMs using supervised learning is the fact that the training data only contains targets for the observed variables $x$, not the latent variables $z$.
A popular approach for training DMMs is using \emph{variational inference}~(VI).
It should be noted that VI is a general method for fitting the parameters of statistical models to data.
In this special case, we happen to be applying it in a case where there is a dependence between samples in time.
For a concrete example of a training algorithm based on VI that is suitable for training DMM, we refer to~\cite{Krishnan2016}.

While probability distributions in classical DMMs are assumed to be Gaussian, recent extensions proposed the use of more expressive but also more computationally expensive deep normalizing flows~\cite{rezende2016variational,DMM_NF_2021}.
Another variant of DMM includes additional graph structure for possible encoding of useful inductive biases~\cite{qu2019gmnn}.
DMMs are typically being trained using the stochastic counterpart of the backpropagation algorithm~\cite{rezende2014stochastic}, that is part of popular open-source libraries such as Pytorch-based Pyro~\cite{bingham2019pyro} or TensorFlow Probability~\cite{Dillon2017}.
Applications in dynamical systems modeling span from climate forecasting~\citep{pmlr-v80-che18a}, molecular dynamics~\citep{wu2019deep}, or generic time series modeling with uncertainty quantification~\citep{montanez2015inertial}.

\subsubsection{Latent Neural ODEs} \label{sec:ts_latent}

\begin{figure}
\centering
\includegraphics{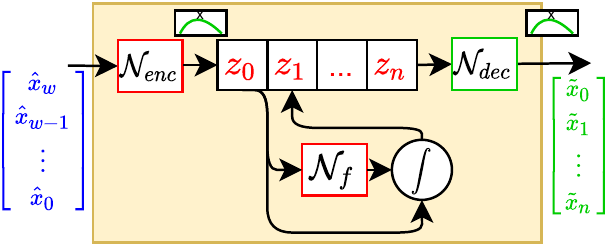}
\caption{
    Latent neural ODEs. 
    An encoder network is used to obtain a latent representation of the system's initial state, $z_0$, by aggregate information from several observations of the systems $[\hat{x}_w,\hat{x}_{w-1},...,\hat{x}_0]$.
    The system is simulated for multiple steps to obtain $[z_0,z_1,...,z_n]$.
    Finally, the latent variables are mapped back to the original state-space by a decoder network.
}
\label{fig:time_stepper_latent_ode}
\end{figure}

\emph{Latent neural ordinary differential equations}~(latent NODEs)~\cite{Chen2019} is an extension of NODEs which introduces an encoder and decoder NN to the model as shown in \cref{fig:time_stepper_latent_ode}.
The core of the idea is that information from multiple observations can be aggregated by the encoder network $N_{enc}$ to obtain a latent state $z_0$, which characterizes the specific trajectory.
A convenient choice of encoder network for time series is a RNN because it can handle a variable number of observations.
The system can then be simulated using the same approach as NODEs to produce a solution in the latent space.
Finally, a decoder network maps each point of the latent solution to the observable space to obtain the final solution.

Separating the measurement, $\mathbf{x}_k$, from the latent system dynamics, $\mathbf{z}_k$, allows us to exploit the modeling flexibility of wider NNs capable of generating more complex latent trajectories.
However, by doing so it creates an inference problem of estimating unknown initial conditions of the hidden states for both deterministic~\cite{Lenz2015DeepMPCLD,skomski2021constrained} and stochastic time-steppers~\cite{krishnan2015deep,Lenz2015DeepMPCLD,Krishnan2017StructuredIN,NIPS2015_b618c321}.

A difference between a latent NODEs and DMMs is that the former treats the state variable as a continuous-time variable and the latter treats it as discrete-time.
Additionally, latent NODEs assumes that the dynamics are deterministic.

\subsubsection{Bayesian Neural Ordinary Differential Equations}

\begin{figure}[h]
\centering
\includegraphics[width=.50\columnwidth]{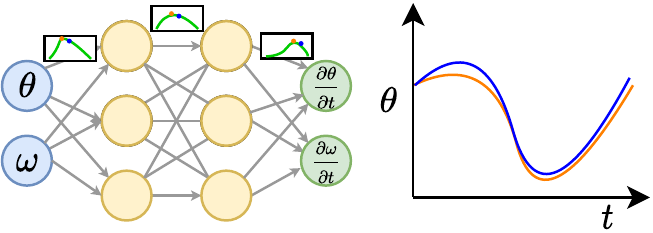}
\caption{
    Bayesian Neural Ordinary Differential Equations. 
    The parameters of the network are characterized by a probability distribution. 
    The parameter distributions are sampled multiple times and used to simulate the system, producing multiple trajectories as shown to the right. 
    To get a single prediction, the predictions can be averaged. 
}
\label{fig:ts_bayesian_nodes}
\end{figure}

\emph{Bayesian neural ordinary differential equations}~(BNODEs)~\cite{Dandekar2021} combine the concept of a NODE with the stochastic nature of \emph{Bayesian neural networks}~(BNN)~\cite{Jospin2020}.
In the context of a BNN, the term \emph{Bayesian} refers to the fact that the parameters of the network are characterized by a probability density function rather of an exact value.
For instance, the weights of the networks may be assumed to be approximately distributed according to a multivariate Gaussian.

A possible motivation for applying this formalism is that the uncertainty of the model's predictions can be quantified, which would otherwise not be possible.
To obtain an estimate of the uncertainty, the model can be simulated several times using different realizations of the model's parameters, resulting in several trajectories as shown in~\cref{fig:ts_bayesian_nodes}.
The ensemble of trajectories can then be used to infer confidence bounds and to obtain the mean value of the trajectories.

A drawback of using BNNs and extensions like BNODEs is that they use specialized training algorithms that generally do not scale well to large network architectures.
An alternative approach is to introduce sources of stochasticity during the training and inference, for instance by using dropout.
A categorization of ways to introduce stochasticity that do not require specialized training algorithms is provided in~\cite[Sec~8]{Jospin2020}.

\subsubsection{Neural Stochastic Differential Equations} \label{sec:ts_nsde}

\begin{figure}[ht]
\centering
\includegraphics[width=0.5\textwidth]{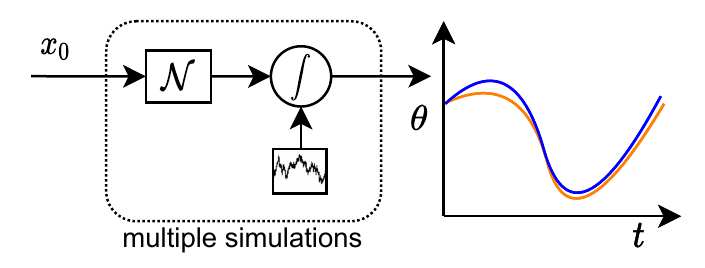}
\caption{
    Neural stochastic differential equations. 
    The network $\mathcal{N}$ is used to approximate the deterministic drift term of the SDE and the diffusion term is a Wiener process. 
    Multiple trajectories are produced by solving the SDE multiple times, corresponding to different realizations of the Wiener process.
}
\label{fig:neural_sde}
\end{figure}

\emph{Neural stochastic differential equations}~(NSDEs)~\cite{Liu2019} can be viewed as a generalization of an ODE that includes one or more stochastic terms in addition to the deterministic dynamics.
Like the DTMC, a SDE often includes a deterministic drift term and a stochastic diffusion term, such as \emph{Wiener process}:
\begin{equation} \label{eq:sde_diff}
dX = f(x(t)) dt + g(x(t)) dW_t.
\end{equation}
Conventionally, SDEs are expressed in \emph{differential form} unlike the derivative form of an ODE.
The reason for this is that many stochastic processes are continuous but cannot be differentiated.
The meaning of \cref{eq:sde_diff} is per definition the integral equation:
\begin{align} \label{eq:sde_int}
x(t) &= x_0 + \int_0^t f(x(s)) ds + \int_0^t g(x(s)) dW_s.
\end{align}

As is the case for ODEs, most SDEs must be solved numerically, since only very few SDEs have analytical solutions.
Solving SDEs requires the use of algorithms which are different from those used to solve deterministic ODEs.
Covering the solvers is outside the scope of this paper, instead we refer to~\cite[Chapter~9]{Kloeden1992} for an in depth coverage.
However, in the context of NSDEs we can simply think of the solver as a means to simulate systems with stochastic dynamics.

There are several choices for how to incorporate the use of NNs for modelling SDEs.
For instance, if the stochastic diffusion term is known, a NN can be trained to approximate the deterministic drift term in~\cref{eq:sde_diff} as in the case of~\cite{Oganesyan2020, Liu2019}.
Another approach is to use NNs to parameterize both the drift and diffusion terms~\cite{Hegde2018}.
Additionally, there are approaches such as~\cite{Xu2021}, which incorporate the idea from both NSDEs and BNNs, by modeling both evolution of the state variables and network parameters as SDEs.

While NSDEs provide a strong theoretical framework for modeling uncertainty, they are complex compared to their deterministic counterparts.
One way to address this is to examine if simpler and computationally efficient mechanisms like injecting noise or using dropout can achieve some of the same effects as adopting a fully SDE based framework.

\section{Discussion}
\label{sec:discussion}


An important question is how to pick the right type of model for a given application.
The two fundamentally different approaches for simulating a system are having i) a NN approximate the solution of the problem, as described in \cref{sec:continuous_time_models}, or ii) having a NN approximate the dynamics of the system, as described in~\cref{sec:time_stepper_models}.
Each approach has inherent advantages and limitations, that can be derived by looking at what the NN is used for within the respective type of model.
A comparison between the two types of models can be seen in \cref{tab:direct_vs_ts}.

\begin{table}[h]
\caption{Comparison of direct-solution and time-stepper models.}
\label{tab:direct_vs_ts}
\centering
\resizebox{\columnwidth}{!}{%
\begin{tabular}{@{}lll@{}}
\toprule
Name &
  Advantages &
  Limitations \\ \midrule
Direct-solution &
  \begin{tabular}[c]{@{}l@{}}+ Easy to apply to PDEs\\ + No discretization of time and spatial coordinates\\ + No accumulation of error during simulation\\ + Parallel evaluation of simulation\end{tabular} &
  \begin{tabular}[c]{@{}l@{}}- Fixed Initial condition\\ - Fixed temporal and spatial domain\\ - Difficult to incorporate inputs\end{tabular} \\ \midrule
Time-stepper &
  \begin{tabular}[c]{@{}l@{}}+ Initial condition not fixed\\ + Easy to incorporate inputs\\ + Leverage knowledge from numerical simulation\end{tabular} &
  \begin{tabular}[c]{@{}l@{}}- Not trivial to apply to PDEs\\ - Accumulation of error during simulation\\ - No parallel evaluation of simulation\end{tabular} \\ \bottomrule
\end{tabular}%
}
\end{table}

In this survey, we described several variants of direct-solution and time-stepper models.
The way that these are presented in the literature, often gives the impression that they are fundamentally different.
However, applying them to the ideal pendulum system, makes it clear that many models are closely related; set apart only by a small extension of the original idea.
In the case of the direct-solution models, we observed that the differences between the vanilla direct-solution and the PINN is the application of physics based regularization and use of AD for obtaining the velocity.
In the case of time-stepper models, the main differences boil down to the architecture of the NN and the numerical integration scheme being applied.
The ability to pick a NN architecture for a specific application makes it possible to model a wide range of physical phenomena.
Additionally, the ideas of one model can easily be transferred to another, allowing for the creation of novel architectures.
This inherent variability makes it difficult to define concrete guidelines for picking a type of model for a certain application.
Instead, we urge the reader to consider what capabilities are needed for the application and how knowledge of the physics incorporated.
The topics described by \cref{fig:survey_structure} may serve as a starting point for this.

Evaluating the performance of different models on a benchmark dataset consisting of data from various dynamical systems would be very useful.
This dataset should be representative of the systems which are encountered in disciplines such as physics, chemistry and engineering.
This would allow us to identify general trends and heuristics, which would serve as a starting point for new practitioners and future applications.
Drawing inspiration from other applications of DL, such as image classification, we see that large image databases have contributed greatly towards developing better NN architectures.
A standardized benchmark dataset is an essential step towards gaining more insight into which types of models work well.
Not only would it allow for a fair comparison between the NN-based models, but it would also allow us to answer the question of how well these models work compared to traditional models originating from various fields.

Another valuable contribution, would be to define a procedure for evaluating a model's ability to approximate a dynamical system.
We are interested in verifying that the model can produce accurate simulations for the initial conditions that we would encounter when using the model. 
Given the diverse nature of these dynamical systems, some may be more difficult for a NN to approximate than others.
For instance, a small approximation error in a chaotic system may result in the accumulation of a large error over time.
An interesting research topic is determining metrics that allow a fair comparison across multiple dynamical systems.

Another valuable contribution would be to develop concrete guidelines on how to train models of dynamical systems.
Finding a rule of thumb for how much training data is necessary to reach a certain degree of accuracy, would make it easier to determine if a data-driven approach is feasible for a given application.
In addition to determining how much data we need, it would be useful to develop best practices on how to split the data into training and validation sets.
For instance, in the context of training time-stepper models, we may examine which length of trajectories result in a good ratio between accuracy and training time.
Likewise, it would be useful to determine how to formulate the loss function such that the process of optimizing the model's parameters is fast and robust.


\section{Summary}
\label{sec:conclusion}

In recent years, there has been an increased interest in applying NNs to solve a diverse set of problems encountered in various branches of engineering and natural sciences.
This has resulted in a wealth of papers; each proposing how a particular physical phenomenon can be simulated using NNs.
As a consequence, the terminology and notation used in each paper vary greatly, making it difficult to digest for all but experts in the respective field.
These papers, often constrained in space, put great emphasis on describing the application and the physics involved, often at a cost of omitting details like how the NN was trained and limitations of proposed methods.

This survey provides an easy-to-follow overview of the techniques for simulating dynamical systems based on NNs.
Specifically, we categorized the models encounter in the literature into two distinct types: direct-solution- and time-stepper models.
For each type of model, we provided a concrete guide on how to construct, train and use the model for simulation.
Starting from the simplest possible model, we incrementally introduced more advanced variants and established the differences and similarities between the models.
Additionally, we supply source code for many of the models described in the paper, which can be used as a reference for detailed implementation of each model. 

An open research question is determining how well these methods work across a broad set of problems that are representative of real-world applications.
We hope that this survey will support this goal by presenting the most important concepts in a way that is accessible to practitioners coming from DL as well as various branches of physics and engineering.


\begin{acks}
  We acknowledge the Poul Due Jensen Foundation for funding the project Digital Twins for Cyber-Physical Systems (DiT4CPS) and Legaard would also like to acknowledge partial support from the MADE Digital project.
  
  This research was supported by 
the Data Model Convergence (DMC) initiative
via the Laboratory Directed Research and Development (LDRD) investments at Pacific Northwest National Laboratory (PNNL). PNNL is a multi-program national laboratory
operated for the U.S. Department of Energy (DoE) by Battelle Memorial Institute under Contract
No. DE-AC05-76RL0-1830.
\end{acks}

\bibliographystyle{ACM-Reference-Format}
\bibliography{deep_learning_for_dynamical_systems}


\begin{thebibliography}{143}


\ifx \showCODEN    \undefined \def \showCODEN     #1{\unskip}     \fi
\ifx \showDOI      \undefined \def \showDOI       #1{#1}\fi
\ifx \showISBNx    \undefined \def \showISBNx     #1{\unskip}     \fi
\ifx \showISBNxiii \undefined \def \showISBNxiii  #1{\unskip}     \fi
\ifx \showISSN     \undefined \def \showISSN      #1{\unskip}     \fi
\ifx \showLCCN     \undefined \def \showLCCN      #1{\unskip}     \fi
\ifx \shownote     \undefined \def \shownote      #1{#1}          \fi
\ifx \showarticletitle \undefined \def \showarticletitle #1{#1}   \fi
\ifx \showURL      \undefined \def \showURL       {\relax}        \fi
\providecommand\bibfield[2]{#2}
\providecommand\bibinfo[2]{#2}
\providecommand\natexlab[1]{#1}
\providecommand\showeprint[2][]{arXiv:#2}

\bibitem[\protect\citeauthoryear{Abadi, Agarwal, Barham, Brevdo, Chen, Citro,
  Corrado, Davis, Dean, Devin, Ghemawat, Goodfellow, Harp, Irving, Isard, Jia,
  Jozefowicz, Kaiser, Kudlur, Levenberg, Mane, Monga, Moore, Murray, Olah,
  Schuster, Shlens, Steiner, Sutskever, Talwar, Tucker, Vanhoucke, Vasudevan,
  Viegas, Vinyals, Warden, Wattenberg, Wicke, Yu, and Zheng}{Abadi
  et~al\mbox{.}}{2016}]%
        {Abadi2016}
\bibfield{author}{\bibinfo{person}{Mart{\'i}n Abadi}, \bibinfo{person}{Ashish
  Agarwal}, \bibinfo{person}{Paul Barham}, \bibinfo{person}{Eugene Brevdo},
  \bibinfo{person}{Zhifeng Chen}, \bibinfo{person}{Craig Citro},
  \bibinfo{person}{Greg~S. Corrado}, \bibinfo{person}{Andy Davis},
  \bibinfo{person}{Jeffrey Dean}, \bibinfo{person}{Matthieu Devin},
  \bibinfo{person}{Sanjay Ghemawat}, \bibinfo{person}{Ian Goodfellow},
  \bibinfo{person}{Andrew Harp}, \bibinfo{person}{Geoffrey Irving},
  \bibinfo{person}{Michael Isard}, \bibinfo{person}{Yangqing Jia},
  \bibinfo{person}{Rafal Jozefowicz}, \bibinfo{person}{Lukasz Kaiser},
  \bibinfo{person}{Manjunath Kudlur}, \bibinfo{person}{Josh Levenberg},
  \bibinfo{person}{Dan Mane}, \bibinfo{person}{Rajat Monga},
  \bibinfo{person}{Sherry Moore}, \bibinfo{person}{Derek Murray},
  \bibinfo{person}{Chris Olah}, \bibinfo{person}{Mike Schuster},
  \bibinfo{person}{Jonathon Shlens}, \bibinfo{person}{Benoit Steiner},
  \bibinfo{person}{Ilya Sutskever}, \bibinfo{person}{Kunal Talwar},
  \bibinfo{person}{Paul Tucker}, \bibinfo{person}{Vincent Vanhoucke},
  \bibinfo{person}{Vijay Vasudevan}, \bibinfo{person}{Fernanda Viegas},
  \bibinfo{person}{Oriol Vinyals}, \bibinfo{person}{Pete Warden},
  \bibinfo{person}{Martin Wattenberg}, \bibinfo{person}{Martin Wicke},
  \bibinfo{person}{Yuan Yu}, {and} \bibinfo{person}{Xiaoqiang Zheng}.}
  \bibinfo{year}{2016}\natexlab{}.
\newblock \bibinfo{booktitle}{\emph{{{TensorFlow}}: Large-{{Scale Machine
  Learning}} on {{Heterogeneous Distributed Systems}}}}.
\newblock \bibinfo{type}{{T}echnical {R}eport} 1603.04467.
\newblock
\showeprint[arxiv]{1603.04467}


\bibitem[\protect\citeauthoryear{Awiszus and Rosenhahn}{Awiszus and
  Rosenhahn}{2018}]%
        {awiszus2018markov}
\bibfield{author}{\bibinfo{person}{Maren Awiszus} {and} \bibinfo{person}{Bodo
  Rosenhahn}.} \bibinfo{year}{2018}\natexlab{}.
\newblock \showarticletitle{Markov Chain Neural Networks}. In
  \bibinfo{booktitle}{\emph{Proceedings of the {{IEEE}} Conference on Computer
  Vision and Pattern Recognition Workshops}}. \bibinfo{pages}{2180--2187}.
\newblock


\bibitem[\protect\citeauthoryear{Battaglia, Hamrick, Bapst, {Sanchez-Gonzalez},
  Zambaldi, Malinowski, Tacchetti, Raposo, Santoro, Faulkner, G{\"u}l{\c
  c}ehre, Song, Ballard, Gilmer, Dahl, Vaswani, Allen, Nash, Langston, Dyer,
  Heess, Wierstra, Kohli, Botvinick, Vinyals, Li, and Pascanu}{Battaglia
  et~al\mbox{.}}{2018}]%
        {Battaglia2021}
\bibfield{author}{\bibinfo{person}{Peter~W. Battaglia},
  \bibinfo{person}{Jessica~B. Hamrick}, \bibinfo{person}{Victor Bapst},
  \bibinfo{person}{Alvaro {Sanchez-Gonzalez}},
  \bibinfo{person}{Vin{\'i}cius~Flores Zambaldi}, \bibinfo{person}{Mateusz
  Malinowski}, \bibinfo{person}{Andrea Tacchetti}, \bibinfo{person}{David
  Raposo}, \bibinfo{person}{Adam Santoro}, \bibinfo{person}{Ryan Faulkner},
  \bibinfo{person}{{\c C}aglar G{\"u}l{\c c}ehre}, \bibinfo{person}{H.~Francis
  Song}, \bibinfo{person}{Andrew~J. Ballard}, \bibinfo{person}{Justin Gilmer},
  \bibinfo{person}{George~E. Dahl}, \bibinfo{person}{Ashish Vaswani},
  \bibinfo{person}{Kelsey~R. Allen}, \bibinfo{person}{Charles Nash},
  \bibinfo{person}{Victoria Langston}, \bibinfo{person}{Chris Dyer},
  \bibinfo{person}{Nicolas Heess}, \bibinfo{person}{Daan Wierstra},
  \bibinfo{person}{Pushmeet Kohli}, \bibinfo{person}{Matthew Botvinick},
  \bibinfo{person}{Oriol Vinyals}, \bibinfo{person}{Yujia Li}, {and}
  \bibinfo{person}{Razvan Pascanu}.} \bibinfo{year}{2018}\natexlab{}.
\newblock \showarticletitle{Relational Inductive Biases, Deep Learning, and
  Graph Networks}.
\newblock \bibinfo{journal}{\emph{CoRR}}  \bibinfo{volume}{abs/1806.01261}
  (\bibinfo{year}{2018}).
\newblock
\showeprint[arxiv]{1806.01261}


\bibitem[\protect\citeauthoryear{Battaglia, Pascanu, Lai, Rezende, and
  Kavukcuoglu}{Battaglia et~al\mbox{.}}{2016}]%
        {BattagliaPLRK16}
\bibfield{author}{\bibinfo{person}{Peter~W. Battaglia}, \bibinfo{person}{Razvan
  Pascanu}, \bibinfo{person}{Matthew Lai}, \bibinfo{person}{Danilo~Jimenez
  Rezende}, {and} \bibinfo{person}{Koray Kavukcuoglu}.}
  \bibinfo{year}{2016}\natexlab{}.
\newblock \showarticletitle{Interaction Networks for Learning about Objects,
  Relations and Physics}.
\newblock \bibinfo{journal}{\emph{CoRR}}  \bibinfo{volume}{abs/1612.00222}
  (\bibinfo{year}{2016}).
\newblock
\showeprint[arxiv]{1612.00222}


\bibitem[\protect\citeauthoryear{Baydin, Pearlmutter, Radul, and
  Siskind}{Baydin et~al\mbox{.}}{2018}]%
        {Baydin2018}
\bibfield{author}{\bibinfo{person}{Atilim~Gunes Baydin},
  \bibinfo{person}{Barak~A. Pearlmutter}, \bibinfo{person}{Alexey~Andreyevich
  Radul}, {and} \bibinfo{person}{Jeffrey~Mark Siskind}.}
  \bibinfo{year}{2018}\natexlab{}.
\newblock \showarticletitle{Automatic Differentiation in Machine Learning: A
  Survey}.
\newblock \bibinfo{journal}{\emph{arXiv:1502.05767 [cs, stat]}}
  (\bibinfo{date}{Feb.} \bibinfo{year}{2018}).
\newblock
\showeprint[arxiv]{1502.05767}~[cs, stat]


\bibitem[\protect\citeauthoryear{Behler}{Behler}{2015}]%
        {Behler2015}
\bibfield{author}{\bibinfo{person}{J{\"o}rg Behler}.}
  \bibinfo{year}{2015}\natexlab{}.
\newblock \showarticletitle{Constructing High-Dimensional Neural Network
  Potentials: A Tutorial Review}.
\newblock \bibinfo{journal}{\emph{International Journal of Quantum Chemistry}}
  \bibinfo{volume}{115}, \bibinfo{number}{16} (\bibinfo{year}{2015}),
  \bibinfo{pages}{1032--1050}.
\newblock
\urldef\tempurl%
\url{https://doi.org/10.1002/qua.24890}
\showDOI{\tempurl}
\showeprint{https://onlinelibrary.wiley.com/doi/pdf/10.1002/qua.24890}


\bibitem[\protect\citeauthoryear{Behrmann, Grathwohl, Chen, Duvenaud, and
  Jacobsen}{Behrmann et~al\mbox{.}}{2019}]%
        {Behrmann2019}
\bibfield{author}{\bibinfo{person}{Jens Behrmann}, \bibinfo{person}{Will
  Grathwohl}, \bibinfo{person}{Ricky T.~Q. Chen}, \bibinfo{person}{David
  Duvenaud}, {and} \bibinfo{person}{Joern-Henrik Jacobsen}.}
  \bibinfo{year}{2019}\natexlab{}.
\newblock \showarticletitle{Invertible Residual Networks}. In
  \bibinfo{booktitle}{\emph{Proceedings of the 36th International Conference on
  Machine Learning}} \emph{(\bibinfo{series}{Proceedings of Machine Learning
  Research}, Vol.~\bibinfo{volume}{97})},
  \bibfield{editor}{\bibinfo{person}{Kamalika Chaudhuri} {and}
  \bibinfo{person}{Ruslan Salakhutdinov}} (Eds.). \bibinfo{publisher}{{PMLR}},
  \bibinfo{pages}{573--582}.
\newblock


\bibitem[\protect\citeauthoryear{Bingham, Chen, Jankowiak, Obermeyer, Pradhan,
  Karaletsos, Singh, Szerlip, Horsfall, and Goodman}{Bingham
  et~al\mbox{.}}{2019}]%
        {bingham2019pyro}
\bibfield{author}{\bibinfo{person}{Eli Bingham}, \bibinfo{person}{Jonathan~P
  Chen}, \bibinfo{person}{Martin Jankowiak}, \bibinfo{person}{Fritz Obermeyer},
  \bibinfo{person}{Neeraj Pradhan}, \bibinfo{person}{Theofanis Karaletsos},
  \bibinfo{person}{Rohit Singh}, \bibinfo{person}{Paul Szerlip},
  \bibinfo{person}{Paul Horsfall}, {and} \bibinfo{person}{Noah~D Goodman}.}
  \bibinfo{year}{2019}\natexlab{}.
\newblock \showarticletitle{Pyro: Deep Universal Probabilistic Programming}.
\newblock \bibinfo{journal}{\emph{The Journal of Machine Learning Research}}
  \bibinfo{volume}{20}, \bibinfo{number}{1} (\bibinfo{year}{2019}),
  \bibinfo{pages}{973--978}.
\newblock


\bibitem[\protect\citeauthoryear{Bishop}{Bishop}{2006}]%
        {Bishop2006}
\bibfield{author}{\bibinfo{person}{Christopher Bishop}.}
  \bibinfo{year}{2006}\natexlab{}.
\newblock \bibinfo{booktitle}{\emph{Pattern {{Recognition}} and {{Machine
  Learning}}}}.
\newblock \bibinfo{publisher}{{Springer-Verlag}}, \bibinfo{address}{{New
  York}}.
\newblock
\showISBNx{978-0-387-31073-2}


\bibitem[\protect\citeauthoryear{Bronstein, Bruna, Cohen, and
  Velickovic}{Bronstein et~al\mbox{.}}{2021}]%
        {Bronstein2021}
\bibfield{author}{\bibinfo{person}{Michael~M. Bronstein}, \bibinfo{person}{Joan
  Bruna}, \bibinfo{person}{Taco Cohen}, {and} \bibinfo{person}{Petar
  Velickovic}.} \bibinfo{year}{2021}\natexlab{}.
\newblock \showarticletitle{Geometric Deep Learning: Grids, Groups, Graphs,
  Geodesics, and Gauges}.
\newblock \bibinfo{journal}{\emph{CoRR}}  \bibinfo{volume}{abs/2104.13478}
  (\bibinfo{year}{2021}).
\newblock
\showeprint[arxiv]{2104.13478}


\bibitem[\protect\citeauthoryear{Brunton, Noack, and Koumoutsakos}{Brunton
  et~al\mbox{.}}{2020}]%
        {Brunton2020}
\bibfield{author}{\bibinfo{person}{Steven~L. Brunton},
  \bibinfo{person}{Bernd~R. Noack}, {and} \bibinfo{person}{Petros
  Koumoutsakos}.} \bibinfo{year}{2020}\natexlab{}.
\newblock \showarticletitle{Machine {{Learning}} for {{Fluid Mechanics}}}.
\newblock \bibinfo{journal}{\emph{Annual Review of Fluid Mechanics}}
  \bibinfo{volume}{52}, \bibinfo{number}{1} (\bibinfo{year}{2020}),
  \bibinfo{pages}{477--508}.
\newblock
\urldef\tempurl%
\url{https://doi.org/10.1146/annurev-fluid-010719-060214}
\showDOI{\tempurl}


\bibitem[\protect\citeauthoryear{Butler, Davies, Cartwright, Isayev, and
  Walsh}{Butler et~al\mbox{.}}{2018}]%
        {Butler2018}
\bibfield{author}{\bibinfo{person}{Keith~T. Butler}, \bibinfo{person}{Daniel~W.
  Davies}, \bibinfo{person}{Hugh Cartwright}, \bibinfo{person}{Olexandr
  Isayev}, {and} \bibinfo{person}{Aron Walsh}.}
  \bibinfo{year}{2018}\natexlab{}.
\newblock \showarticletitle{Machine Learning for Molecular and Materials
  Science}.
\newblock \bibinfo{journal}{\emph{Nature}} \bibinfo{volume}{559},
  \bibinfo{number}{7715} (\bibinfo{date}{July} \bibinfo{year}{2018}),
  \bibinfo{pages}{547--555}.
\newblock
\showISSN{0028-0836, 1476-4687}
\urldef\tempurl%
\url{https://doi.org/10.1038/s41586-018-0337-2}
\showDOI{\tempurl}


\bibitem[\protect\citeauthoryear{Cellier}{Cellier}{1991}]%
        {Cellier1991}
\bibfield{author}{\bibinfo{person}{Fran{\c c}ois~Edouard Cellier}.}
  \bibinfo{year}{1991}\natexlab{}.
\newblock \bibinfo{booktitle}{\emph{Continuous System Modeling}}.
\newblock \bibinfo{publisher}{{Springer Science \& Business Media}}.
\newblock


\bibitem[\protect\citeauthoryear{Cellier and Kofman}{Cellier and
  Kofman}{2006}]%
        {Cellier2006}
\bibfield{author}{\bibinfo{person}{Fran{\c c}ois~Edouard Cellier} {and}
  \bibinfo{person}{Ernesto Kofman}.} \bibinfo{year}{2006}\natexlab{}.
\newblock \bibinfo{booktitle}{\emph{Continuous {{System Simulation}}}}.
\newblock \bibinfo{publisher}{{Springer Science \& Business Media}}.
\newblock
\showISBNx{978-0-387-26102-7}


\bibitem[\protect\citeauthoryear{Chang, Meng, Haber, Tung, and Begert}{Chang
  et~al\mbox{.}}{2018}]%
        {Chang2018}
\bibfield{author}{\bibinfo{person}{Bo Chang}, \bibinfo{person}{Lili Meng},
  \bibinfo{person}{Eldad Haber}, \bibinfo{person}{Frederick Tung}, {and}
  \bibinfo{person}{David Begert}.} \bibinfo{year}{2018}\natexlab{}.
\newblock \showarticletitle{Multi-Level {{Residual Networks}} from {{Dynamical
  Systems View}}}.
\newblock \bibinfo{journal}{\emph{arXiv:1710.10348 [cs, stat]}}
  (\bibinfo{date}{Feb.} \bibinfo{year}{2018}).
\newblock
\showeprint[arxiv]{1710.10348}~[cs, stat]


\bibitem[\protect\citeauthoryear{Chang, Ullman, Torralba, and Tenenbaum}{Chang
  et~al\mbox{.}}{2016}]%
        {ChangUTT16}
\bibfield{author}{\bibinfo{person}{Michael~B. Chang}, \bibinfo{person}{Tomer
  Ullman}, \bibinfo{person}{Antonio Torralba}, {and} \bibinfo{person}{Joshua~B.
  Tenenbaum}.} \bibinfo{year}{2016}\natexlab{}.
\newblock \showarticletitle{A Compositional Object-Based Approach to Learning
  Physical Dynamics}.
\newblock \bibinfo{journal}{\emph{CoRR}}  \bibinfo{volume}{abs/1612.00341}
  (\bibinfo{year}{2016}).
\newblock
\showeprint[arxiv]{1612.00341}


\bibitem[\protect\citeauthoryear{Che, Purushotham, Li, Jiang, and Liu}{Che
  et~al\mbox{.}}{2018}]%
        {pmlr-v80-che18a}
\bibfield{author}{\bibinfo{person}{Zhengping Che}, \bibinfo{person}{Sanjay
  Purushotham}, \bibinfo{person}{Guangyu Li}, \bibinfo{person}{Bo Jiang}, {and}
  \bibinfo{person}{Yan Liu}.} \bibinfo{year}{2018}\natexlab{}.
\newblock \showarticletitle{Hierarchical Deep Generative Models for Multi-Rate
  Multivariate Time Series}. In \bibinfo{booktitle}{\emph{Proceedings of the
  35th International Conference on Machine Learning}}
  \emph{(\bibinfo{series}{Proceedings of Machine Learning Research},
  Vol.~\bibinfo{volume}{80})}, \bibfield{editor}{\bibinfo{person}{Jennifer Dy}
  {and} \bibinfo{person}{Andreas Krause}} (Eds.). \bibinfo{publisher}{{PMLR}},
  \bibinfo{address}{{Stockholmsm\"assan, Stockholm Sweden}},
  \bibinfo{pages}{784--793}.
\newblock


\bibitem[\protect\citeauthoryear{Chen, Rubanova, Bettencourt, and
  Duvenaud}{Chen et~al\mbox{.}}{2019}]%
        {Chen2019}
\bibfield{author}{\bibinfo{person}{Ricky T.~Q. Chen}, \bibinfo{person}{Yulia
  Rubanova}, \bibinfo{person}{Jesse Bettencourt}, {and} \bibinfo{person}{David
  Duvenaud}.} \bibinfo{year}{2019}\natexlab{}.
\newblock \showarticletitle{Neural {{Ordinary Differential Equations}}}.
\newblock \bibinfo{journal}{\emph{arXiv:1806.07366 [cs, stat]}}
  (\bibinfo{date}{Dec.} \bibinfo{year}{2019}).
\newblock
\showeprint[arxiv]{1806.07366}~[cs, stat]


\bibitem[\protect\citeauthoryear{Ching, Himmelstein, {Beaulieu-Jones}, Kalinin,
  Do, Way, Ferrero, Agapow, Zietz, Hoffman, Xie, Rosen, Lengerich, Israeli,
  Lanchantin, Woloszynek, Carpenter, Shrikumar, Xu, Cofer, Lavender, Turaga,
  Alexandari, Lu, Harris, DeCaprio, Qi, Kundaje, Peng, Wiley, Segler, Boca,
  Swamidass, Huang, Gitter, and Greene}{Ching et~al\mbox{.}}{2018}]%
        {Ching2018}
\bibfield{author}{\bibinfo{person}{Travers Ching}, \bibinfo{person}{Daniel~S.
  Himmelstein}, \bibinfo{person}{Brett~K. {Beaulieu-Jones}},
  \bibinfo{person}{Alexandr~A. Kalinin}, \bibinfo{person}{Brian~T. Do},
  \bibinfo{person}{Gregory~P. Way}, \bibinfo{person}{Enrico Ferrero},
  \bibinfo{person}{Paul-Michael Agapow}, \bibinfo{person}{Michael Zietz},
  \bibinfo{person}{Michael~M. Hoffman}, \bibinfo{person}{Wei Xie},
  \bibinfo{person}{Gail~L. Rosen}, \bibinfo{person}{Benjamin~J. Lengerich},
  \bibinfo{person}{Johnny Israeli}, \bibinfo{person}{Jack Lanchantin},
  \bibinfo{person}{Stephen Woloszynek}, \bibinfo{person}{Anne~E. Carpenter},
  \bibinfo{person}{Avanti Shrikumar}, \bibinfo{person}{Jinbo Xu},
  \bibinfo{person}{Evan~M. Cofer}, \bibinfo{person}{Christopher~A. Lavender},
  \bibinfo{person}{Srinivas~C. Turaga}, \bibinfo{person}{Amr~M. Alexandari},
  \bibinfo{person}{Zhiyong Lu}, \bibinfo{person}{David~J. Harris},
  \bibinfo{person}{Dave DeCaprio}, \bibinfo{person}{Yanjun Qi},
  \bibinfo{person}{Anshul Kundaje}, \bibinfo{person}{Yifan Peng},
  \bibinfo{person}{Laura~K. Wiley}, \bibinfo{person}{Marwin H.~S. Segler},
  \bibinfo{person}{Simina~M. Boca}, \bibinfo{person}{S.~Joshua Swamidass},
  \bibinfo{person}{Austin Huang}, \bibinfo{person}{Anthony Gitter}, {and}
  \bibinfo{person}{Casey~S. Greene}.} \bibinfo{year}{2018}\natexlab{}.
\newblock \showarticletitle{Opportunities and Obstacles for Deep Learning in
  Biology and Medicine}.
\newblock \bibinfo{journal}{\emph{Journal of The Royal Society Interface}}
  \bibinfo{volume}{15}, \bibinfo{number}{141} (\bibinfo{date}{April}
  \bibinfo{year}{2018}), \bibinfo{pages}{20170387}.
\newblock
\urldef\tempurl%
\url{https://doi.org/10.1098/rsif.2017.0387}
\showDOI{\tempurl}


\bibitem[\protect\citeauthoryear{Chung, Kastner, Dinh, Goel, Courville, and
  Bengio}{Chung et~al\mbox{.}}{2015}]%
        {NIPS2015_b618c321}
\bibfield{author}{\bibinfo{person}{Junyoung Chung}, \bibinfo{person}{Kyle
  Kastner}, \bibinfo{person}{Laurent Dinh}, \bibinfo{person}{Kratarth Goel},
  \bibinfo{person}{Aaron~C Courville}, {and} \bibinfo{person}{Yoshua Bengio}.}
  \bibinfo{year}{2015}\natexlab{}.
\newblock \showarticletitle{A Recurrent Latent Variable Model for Sequential
  Data}. In \bibinfo{booktitle}{\emph{Advances in Neural Information Processing
  Systems}}, \bibfield{editor}{\bibinfo{person}{C.~Cortes},
  \bibinfo{person}{N.~Lawrence}, \bibinfo{person}{D.~Lee},
  \bibinfo{person}{M.~Sugiyama}, {and} \bibinfo{person}{R.~Garnett}} (Eds.),
  Vol.~\bibinfo{volume}{28}. \bibinfo{publisher}{{Curran Associates, Inc.}}
\newblock


\bibitem[\protect\citeauthoryear{Cranmer, Greydanus, Hoyer, Battaglia, Spergel,
  and Ho}{Cranmer et~al\mbox{.}}{2020}]%
        {Cranmer2020}
\bibfield{author}{\bibinfo{person}{Miles Cranmer}, \bibinfo{person}{Sam
  Greydanus}, \bibinfo{person}{Stephan Hoyer}, \bibinfo{person}{Peter
  Battaglia}, \bibinfo{person}{David Spergel}, {and} \bibinfo{person}{Shirley
  Ho}.} \bibinfo{year}{2020}\natexlab{}.
\newblock \showarticletitle{Lagrangian {{Neural Networks}}}.
\newblock \bibinfo{journal}{\emph{arXiv:2003.04630 [physics, stat]}}
  (\bibinfo{date}{July} \bibinfo{year}{2020}).
\newblock
\showeprint[arxiv]{2003.04630}~[physics, stat]


\bibitem[\protect\citeauthoryear{Dandekar, Chung, Dixit, Tarek,
  {Garcia-Valadez}, Vemula, and Rackauckas}{Dandekar et~al\mbox{.}}{2021}]%
        {Dandekar2021}
\bibfield{author}{\bibinfo{person}{Raj Dandekar}, \bibinfo{person}{Karen
  Chung}, \bibinfo{person}{Vaibhav Dixit}, \bibinfo{person}{Mohamed Tarek},
  \bibinfo{person}{Aslan {Garcia-Valadez}}, \bibinfo{person}{Krishna~Vishal
  Vemula}, {and} \bibinfo{person}{Chris Rackauckas}.}
  \bibinfo{year}{2021}\natexlab{}.
\newblock \showarticletitle{Bayesian {{Neural Ordinary Differential
  Equations}}}.
\newblock \bibinfo{journal}{\emph{arXiv:2012.07244 [cs]}}
  (\bibinfo{date}{March} \bibinfo{year}{2021}).
\newblock
\showeprint[arxiv]{2012.07244}~[cs]


\bibitem[\protect\citeauthoryear{Diehl, Bock, Schl{\"o}der, Findeisen, Nagy,
  and Allg{\"o}wer}{Diehl et~al\mbox{.}}{2002}]%
        {DIEHL2002577}
\bibfield{author}{\bibinfo{person}{Moritz Diehl}, \bibinfo{person}{H.Georg
  Bock}, \bibinfo{person}{Johannes~P. Schl{\"o}der}, \bibinfo{person}{Rolf
  Findeisen}, \bibinfo{person}{Zoltan Nagy}, {and} \bibinfo{person}{Frank
  Allg{\"o}wer}.} \bibinfo{year}{2002}\natexlab{}.
\newblock \showarticletitle{Real-Time Optimization and Nonlinear Model
  Predictive Control of Processes Governed by Differential-Algebraic
  Equations}.
\newblock \bibinfo{journal}{\emph{Journal of Process Control}}
  \bibinfo{volume}{12}, \bibinfo{number}{4} (\bibinfo{year}{2002}),
  \bibinfo{pages}{577--585}.
\newblock
\showISSN{0959-1524}
\urldef\tempurl%
\url{https://doi.org/10.1016/S0959-1524(01)00023-3}
\showDOI{\tempurl}


\bibitem[\protect\citeauthoryear{Dillon, Langmore, Tran, Brevdo, Vasudevan,
  Moore, Patton, Alemi, Hoffman, and Saurous}{Dillon et~al\mbox{.}}{2017}]%
        {Dillon2017}
\bibfield{author}{\bibinfo{person}{Joshua~V. Dillon}, \bibinfo{person}{Ian
  Langmore}, \bibinfo{person}{Dustin Tran}, \bibinfo{person}{Eugene Brevdo},
  \bibinfo{person}{Srinivas Vasudevan}, \bibinfo{person}{Dave Moore},
  \bibinfo{person}{Brian Patton}, \bibinfo{person}{Alex Alemi},
  \bibinfo{person}{Matthew~D. Hoffman}, {and} \bibinfo{person}{Rif~A.
  Saurous}.} \bibinfo{year}{2017}\natexlab{}.
\newblock \showarticletitle{{{TensorFlow}} Distributions}.
\newblock \bibinfo{journal}{\emph{CoRR}}  \bibinfo{volume}{abs/1711.10604}
  (\bibinfo{year}{2017}).
\newblock
\showeprint[arxiv]{1711.10604}


\bibitem[\protect\citeauthoryear{Drgo{\v n}a, Arroyo, Cupeiro~Figueroa, Blum,
  Arendt, Kim, Oll{\'e}, Oravec, Wetter, Vrabie, and Helsen}{Drgo{\v n}a
  et~al\mbox{.}}{2020}]%
        {DRGONA2020190}
\bibfield{author}{\bibinfo{person}{J{\'a}n Drgo{\v n}a},
  \bibinfo{person}{Javier Arroyo}, \bibinfo{person}{Iago Cupeiro~Figueroa},
  \bibinfo{person}{David Blum}, \bibinfo{person}{Krzysztof Arendt},
  \bibinfo{person}{Donghun Kim}, \bibinfo{person}{Enric~Perarnau Oll{\'e}},
  \bibinfo{person}{Juraj Oravec}, \bibinfo{person}{Michael Wetter},
  \bibinfo{person}{Draguna~L. Vrabie}, {and} \bibinfo{person}{Lieve Helsen}.}
  \bibinfo{year}{2020}\natexlab{}.
\newblock \showarticletitle{All You Need to Know about Model Predictive Control
  for Buildings}.
\newblock \bibinfo{journal}{\emph{Annual Reviews in Control}}
  \bibinfo{volume}{50} (\bibinfo{year}{2020}), \bibinfo{pages}{190--232}.
\newblock
\showISSN{1367-5788}
\urldef\tempurl%
\url{https://doi.org/10.1016/j.arcontrol.2020.09.001}
\showDOI{\tempurl}


\bibitem[\protect\citeauthoryear{Drgona, Tuor, Chandan, and Vrabie}{Drgona
  et~al\mbox{.}}{2020}]%
        {drgona2020physicsconstrained}
\bibfield{author}{\bibinfo{person}{Jan Drgona}, \bibinfo{person}{Aaron~R.
  Tuor}, \bibinfo{person}{Vikas Chandan}, {and} \bibinfo{person}{Draguna~L.
  Vrabie}.} \bibinfo{year}{2020}\natexlab{}.
\newblock \bibinfo{title}{Physics-Constrained Deep Learning of Multi-Zone
  Building Thermal Dynamics}.
\newblock
\newblock
\showeprint[arxiv]{2011.05987}~[cs.LG]


\bibitem[\protect\citeauthoryear{Dupont, Doucet, and Teh}{Dupont
  et~al\mbox{.}}{2019}]%
        {dupont2019augmented}
\bibfield{author}{\bibinfo{person}{Emilien Dupont}, \bibinfo{person}{Arnaud
  Doucet}, {and} \bibinfo{person}{Yee~Whye Teh}.}
  \bibinfo{year}{2019}\natexlab{}.
\newblock \bibinfo{title}{Augmented Neural {{ODEs}}}.
\newblock
\newblock
\showeprint[arxiv]{1904.01681}~[stat.ML]


\bibitem[\protect\citeauthoryear{Finlay, Jacobsen, Nurbekyan, and
  Oberman}{Finlay et~al\mbox{.}}{2020}]%
        {finlay2020train}
\bibfield{author}{\bibinfo{person}{Chris Finlay},
  \bibinfo{person}{J{\"o}rn-Henrik Jacobsen}, \bibinfo{person}{Levon
  Nurbekyan}, {and} \bibinfo{person}{Adam~M Oberman}.}
  \bibinfo{year}{2020}\natexlab{}.
\newblock \bibinfo{title}{How to Train Your Neural {{ODE}}: The World of
  {{Jacobian}} and Kinetic Regularization}.
\newblock
\newblock
\showeprint[arxiv]{2002.02798}~[stat.ML]


\bibitem[\protect\citeauthoryear{Finzi, Wang, and Wilson}{Finzi
  et~al\mbox{.}}{2020}]%
        {Finzi2020}
\bibfield{author}{\bibinfo{person}{Marc Finzi}, \bibinfo{person}{Ke~Alexander
  Wang}, {and} \bibinfo{person}{Andrew~Gordon Wilson}.}
  \bibinfo{year}{2020}\natexlab{}.
\newblock \showarticletitle{Simplifying Hamiltonian and Lagrangian Neural
  Networks via Explicit Constraints}.
\newblock \bibinfo{journal}{\emph{CoRR}}  \bibinfo{volume}{abs/2010.13581}
  (\bibinfo{year}{2020}).
\newblock
\showeprint[arxiv]{2010.13581}


\bibitem[\protect\citeauthoryear{Forgione and Piga}{Forgione and Piga}{2020}]%
        {forgione2020dynonet}
\bibfield{author}{\bibinfo{person}{Marco Forgione} {and} \bibinfo{person}{Dario
  Piga}.} \bibinfo{year}{2020}\natexlab{}.
\newblock \bibinfo{title}{{{dynoNet}}: A Neural Network Architecture for
  Learning Dynamical Systems}.
\newblock
\newblock
\showeprint[arxiv]{2006.02250}~[cs.LG]


\bibitem[\protect\citeauthoryear{Forrester and Keane}{Forrester and
  Keane}{2009}]%
        {Forrester2009}
\bibfield{author}{\bibinfo{person}{Alexander~I.J. Forrester} {and}
  \bibinfo{person}{Andy~J. Keane}.} \bibinfo{year}{2009}\natexlab{}.
\newblock \showarticletitle{Recent Advances in Surrogate-Based Optimization}.
\newblock \bibinfo{journal}{\emph{Progress in Aerospace Sciences}}
  \bibinfo{volume}{45}, \bibinfo{number}{1-3} (\bibinfo{date}{Jan.}
  \bibinfo{year}{2009}), \bibinfo{pages}{50--79}.
\newblock
\showISSN{03760421}
\urldef\tempurl%
\url{https://doi.org/10.1016/j.paerosci.2008.11.001}
\showDOI{\tempurl}


\bibitem[\protect\citeauthoryear{Fraccaro, S{\o}nderby, Paquet, and
  Winther}{Fraccaro et~al\mbox{.}}{2016}]%
        {fraccaro2016sequential}
\bibfield{author}{\bibinfo{person}{Marco Fraccaro},
  \bibinfo{person}{S{\o}ren~Kaae S{\o}nderby}, \bibinfo{person}{Ulrich Paquet},
  {and} \bibinfo{person}{Ole Winther}.} \bibinfo{year}{2016}\natexlab{}.
\newblock \showarticletitle{Sequential Neural Models with Stochastic Layers}.
\newblock \bibinfo{journal}{\emph{arXiv preprint arXiv:1605.07571}}
  (\bibinfo{year}{2016}).
\newblock
\showeprint[arxiv]{1605.07571}


\bibitem[\protect\citeauthoryear{Friedman and Ghidella}{Friedman and
  Ghidella}{2006}]%
        {Friedman2006}
\bibfield{author}{\bibinfo{person}{Jonathan Friedman} {and}
  \bibinfo{person}{Jason Ghidella}.} \bibinfo{year}{2006}\natexlab{}.
\newblock \showarticletitle{Using {{Model}}-{{Based Design}} for {{Automotive
  Systems Engineering}} - {{Requirements Analysis}} of the {{Power Window
  Example}}}. In \bibinfo{booktitle}{\emph{Transactions {{Journal}} of
  {{Passenger Cars}}: Electronic and {{Electrical Systems}}}}
  \emph{(\bibinfo{series}{Automotive {{Systems Engineering}}},
  Vol.~\bibinfo{volume}{115})}. \bibinfo{publisher}{{SAE Technical Paper}},
  \bibinfo{address}{{Detroit, USA}}, \bibinfo{pages}{8}.
\newblock
\showISBNx{0148-7191}
\urldef\tempurl%
\url{https://doi.org/10.4271/2006-01-1217}
\showDOI{\tempurl}


\bibitem[\protect\citeauthoryear{Gao, Ramezanghorbani, Isayev, Smith, and
  Roitberg}{Gao et~al\mbox{.}}{2020}]%
        {TorchANI2020}
\bibfield{author}{\bibinfo{person}{Xiang Gao}, \bibinfo{person}{Farhad
  Ramezanghorbani}, \bibinfo{person}{Olexandr Isayev},
  \bibinfo{person}{Justin~S. Smith}, {and} \bibinfo{person}{Adrian~E.
  Roitberg}.} \bibinfo{year}{2020}\natexlab{}.
\newblock \showarticletitle{{{TorchANI}}: A Free and Open Source
  {{PyTorch}}-{{Based}} Deep Learning Implementation of the {{ANI}} Neural
  Network Potentials}.
\newblock \bibinfo{journal}{\emph{Journal of Chemical Information and
  Modeling}} \bibinfo{volume}{60}, \bibinfo{number}{7} (\bibinfo{year}{2020}),
  \bibinfo{pages}{3408--3415}.
\newblock
\urldef\tempurl%
\url{https://doi.org/10.1021/acs.jcim.0c00451}
\showDOI{\tempurl}
\showeprint{https://doi.org/10.1021/acs.jcim.0c00451}


\bibitem[\protect\citeauthoryear{Garc{\'i}a, Prett, and Morari}{Garc{\'i}a
  et~al\mbox{.}}{1989}]%
        {GARCIA1989335}
\bibfield{author}{\bibinfo{person}{Carlos~E. Garc{\'i}a},
  \bibinfo{person}{David~M. Prett}, {and} \bibinfo{person}{Manfred Morari}.}
  \bibinfo{year}{1989}\natexlab{}.
\newblock \showarticletitle{Model Predictive Control: Theory and
  Practice\textemdash{{A}} Survey}.
\newblock \bibinfo{journal}{\emph{Automatica}} \bibinfo{volume}{25},
  \bibinfo{number}{3} (\bibinfo{year}{1989}), \bibinfo{pages}{335--348}.
\newblock
\showISSN{0005-1098}
\urldef\tempurl%
\url{https://doi.org/10.1016/0005-1098(89)90002-2}
\showDOI{\tempurl}


\bibitem[\protect\citeauthoryear{Gear and Osterby}{Gear and Osterby}{1984}]%
        {Gear1984b}
\bibfield{author}{\bibinfo{person}{C~W Gear} {and} \bibinfo{person}{O
  Osterby}.} \bibinfo{year}{1984}\natexlab{}.
\newblock \showarticletitle{Solving {{Ordinary Differential Equations}} with
  {{Discontinuities}}}.
\newblock \bibinfo{journal}{\emph{ACM Trans. Math. Softw.}}
  \bibinfo{volume}{10}, \bibinfo{number}{1} (\bibinfo{date}{Jan.}
  \bibinfo{year}{1984}), \bibinfo{pages}{23--44}.
\newblock
\showISSN{0098-3500}
\urldef\tempurl%
\url{https://doi.org/10.1145/356068.356071}
\showDOI{\tempurl}


\bibitem[\protect\citeauthoryear{Gedon, Wahlstr{\"o}m, Sch{\"o}n, and
  Ljung}{Gedon et~al\mbox{.}}{2020}]%
        {gedon2020deep}
\bibfield{author}{\bibinfo{person}{Daniel Gedon}, \bibinfo{person}{Niklas
  Wahlstr{\"o}m}, \bibinfo{person}{Thomas~B. Sch{\"o}n}, {and}
  \bibinfo{person}{Lennart Ljung}.} \bibinfo{year}{2020}\natexlab{}.
\newblock \bibinfo{title}{Deep State Space Models for Nonlinear System
  Identification}.
\newblock
\newblock
\showeprint[arxiv]{2003.14162}~[eess.SY]


\bibitem[\protect\citeauthoryear{Ghosh, Honor{\'e}, Liu, Henter, and
  Chatterjee}{Ghosh et~al\mbox{.}}{2021}]%
        {DMM_NF_2021}
\bibfield{author}{\bibinfo{person}{Anubhab Ghosh}, \bibinfo{person}{Antoine
  Honor{\'e}}, \bibinfo{person}{Dong Liu}, \bibinfo{person}{Gustav~Eje Henter},
  {and} \bibinfo{person}{Saikat Chatterjee}.} \bibinfo{year}{2021}\natexlab{}.
\newblock \showarticletitle{Robust Classification Using Hidden Markov Models
  and Mixtures of Normalizing Flows}.
\newblock \bibinfo{journal}{\emph{CoRR}}  \bibinfo{volume}{abs/2102.07284}
  (\bibinfo{year}{2021}).
\newblock
\showeprint[arxiv]{2102.07284}


\bibitem[\protect\citeauthoryear{Gilmer, Schoenholz, Riley, Vinyals, and
  Dahl}{Gilmer et~al\mbox{.}}{2017}]%
        {GilmerSRVD17}
\bibfield{author}{\bibinfo{person}{Justin Gilmer}, \bibinfo{person}{Samuel~S.
  Schoenholz}, \bibinfo{person}{Patrick~F. Riley}, \bibinfo{person}{Oriol
  Vinyals}, {and} \bibinfo{person}{George~E. Dahl}.}
  \bibinfo{year}{2017}\natexlab{}.
\newblock \showarticletitle{Neural Message Passing for Quantum Chemistry}.
\newblock \bibinfo{journal}{\emph{CoRR}}  \bibinfo{volume}{abs/1704.01212}
  (\bibinfo{year}{2017}).
\newblock
\showeprint[arxiv]{1704.01212}


\bibitem[\protect\citeauthoryear{Goodfellow, Bengio, Courville, and
  Bengio}{Goodfellow et~al\mbox{.}}{2016}]%
        {Goodfellow2016}
\bibfield{author}{\bibinfo{person}{Ian Goodfellow}, \bibinfo{person}{Yoshua
  Bengio}, \bibinfo{person}{Aaron Courville}, {and} \bibinfo{person}{Yoshua
  Bengio}.} \bibinfo{year}{2016}\natexlab{}.
\newblock \bibinfo{booktitle}{\emph{Deep Learning}}. Vol.~\bibinfo{volume}{1}.
\newblock \bibinfo{publisher}{{MIT press Cambridge}}.
\newblock


\bibitem[\protect\citeauthoryear{Greydanus, Dzamba, and Yosinski}{Greydanus
  et~al\mbox{.}}{2019}]%
        {Greydanus2019}
\bibfield{author}{\bibinfo{person}{Samuel Greydanus}, \bibinfo{person}{Misko
  Dzamba}, {and} \bibinfo{person}{Jason Yosinski}.}
  \bibinfo{year}{2019}\natexlab{}.
\newblock \showarticletitle{Hamiltonian {{Neural Networks}}}.
\newblock In \bibinfo{booktitle}{\emph{Advances in {{Neural Information
  Processing Systems}} 32}}, \bibfield{editor}{\bibinfo{person}{H.~Wallach},
  \bibinfo{person}{H.~Larochelle}, \bibinfo{person}{A.~Beygelzimer},
  \bibinfo{person}{F.~d{\textbackslash}textquotesingle {Alch{\'e}-Buc}},
  \bibinfo{person}{E.~Fox}, {and} \bibinfo{person}{R.~Garnett}} (Eds.).
  \bibinfo{publisher}{{Curran Associates, Inc.}},
  \bibinfo{pages}{15379--15389}.
\newblock


\bibitem[\protect\citeauthoryear{G{\"u}ler, Laignelet, and Parpas}{G{\"u}ler
  et~al\mbox{.}}{2019}]%
        {guler2019robust}
\bibfield{author}{\bibinfo{person}{Batuhan G{\"u}ler}, \bibinfo{person}{Alexis
  Laignelet}, {and} \bibinfo{person}{Panos Parpas}.}
  \bibinfo{year}{2019}\natexlab{}.
\newblock \bibinfo{title}{Towards Robust and Stable Deep Learning Algorithms
  for Forward Backward Stochastic Differential Equations}.
\newblock
\newblock
\showeprint[arxiv]{1910.11623}~[stat.ML]


\bibitem[\protect\citeauthoryear{Hafner, Lillicrap, Fischer, Villegas, Ha, Lee,
  and Davidson}{Hafner et~al\mbox{.}}{2018}]%
        {LatentDynamics2018}
\bibfield{author}{\bibinfo{person}{Danijar Hafner}, \bibinfo{person}{Timothy~P.
  Lillicrap}, \bibinfo{person}{Ian Fischer}, \bibinfo{person}{Ruben Villegas},
  \bibinfo{person}{David Ha}, \bibinfo{person}{Honglak Lee}, {and}
  \bibinfo{person}{James Davidson}.} \bibinfo{year}{2018}\natexlab{}.
\newblock \showarticletitle{Learning Latent Dynamics for Planning from Pixels}.
\newblock \bibinfo{journal}{\emph{CoRR}}  \bibinfo{volume}{abs/1811.04551}
  (\bibinfo{year}{2018}).
\newblock
\showeprint[arxiv]{1811.04551}


\bibitem[\protect\citeauthoryear{Hairer and Wanner}{Hairer and Wanner}{1996}]%
        {Hairer1996}
\bibfield{author}{\bibinfo{person}{Ernst Hairer} {and} \bibinfo{person}{Gerhard
  Wanner}.} \bibinfo{year}{1996}\natexlab{}.
\newblock \bibinfo{booktitle}{\emph{Solving Ordinary Differential Equations
  {{II}}: Stiff and Differential-Algebraic Problems}}.
\newblock Number~14. \bibinfo{publisher}{{Springer-Verlag Berlin Heidelberg}}.
\newblock
\showISBNx{3-540-60452-9}


\bibitem[\protect\citeauthoryear{He, Zhang, Ren, and Sun}{He
  et~al\mbox{.}}{2015}]%
        {He2015}
\bibfield{author}{\bibinfo{person}{Kaiming He}, \bibinfo{person}{Xiangyu
  Zhang}, \bibinfo{person}{Shaoqing Ren}, {and} \bibinfo{person}{Jian Sun}.}
  \bibinfo{year}{2015}\natexlab{}.
\newblock \showarticletitle{Deep {{Residual Learning}} for {{Image
  Recognition}}}.
\newblock \bibinfo{journal}{\emph{arXiv:1512.03385 [cs]}} (\bibinfo{date}{Dec.}
  \bibinfo{year}{2015}).
\newblock
\showeprint[arxiv]{1512.03385}~[cs]


\bibitem[\protect\citeauthoryear{Hegde, Heinonen, L{\"a}hdesm{\"a}ki, and
  Kaski}{Hegde et~al\mbox{.}}{2018}]%
        {Hegde2018}
\bibfield{author}{\bibinfo{person}{Pashupati Hegde}, \bibinfo{person}{Markus
  Heinonen}, \bibinfo{person}{Harri L{\"a}hdesm{\"a}ki}, {and}
  \bibinfo{person}{Samuel Kaski}.} \bibinfo{year}{2018}\natexlab{}.
\newblock \showarticletitle{Deep Learning with Differential {{Gaussian}}
  Process Flows}.
\newblock \bibinfo{journal}{\emph{arXiv:1810.04066 [cs, stat]}}
  (\bibinfo{date}{Oct.} \bibinfo{year}{2018}).
\newblock
\showeprint[arxiv]{1810.04066}~[cs, stat]


\bibitem[\protect\citeauthoryear{{Jeen-Shing Wang} and {Yi-Chung
  Chen}}{{Jeen-Shing Wang} and {Yi-Chung Chen}}{2008}]%
        {HW_RNN2008}
\bibfield{author}{\bibinfo{person}{{Jeen-Shing Wang}} {and}
  \bibinfo{person}{{Yi-Chung Chen}}.} \bibinfo{year}{2008}\natexlab{}.
\newblock \showarticletitle{A {{Hammerstein}}-{{Wiener}} Recurrent Neural
  Network with Universal Approximation Capability}. In
  \bibinfo{booktitle}{\emph{2008 {{IEEE}} International Conference on Systems,
  Man and Cybernetics}}. \bibinfo{pages}{1832--1837}.
\newblock
\showISSN{1062-922X}
\urldef\tempurl%
\url{https://doi.org/10.1109/ICSMC.2008.4811555}
\showDOI{\tempurl}


\bibitem[\protect\citeauthoryear{Jia and Benson}{Jia and Benson}{2019}]%
        {JumpStochNODEs2019}
\bibfield{author}{\bibinfo{person}{Junteng Jia} {and}
  \bibinfo{person}{Austin~R. Benson}.} \bibinfo{year}{2019}\natexlab{}.
\newblock \showarticletitle{Neural Jump Stochastic Differential Equations}.
\newblock \bibinfo{journal}{\emph{CoRR}}  \bibinfo{volume}{abs/1905.10403}
  (\bibinfo{year}{2019}).
\newblock
\showeprint[arxiv]{1905.10403}


\bibitem[\protect\citeauthoryear{Jia, Wang, Chen, Lu, Lin, Car, E, and
  Zhang}{Jia et~al\mbox{.}}{2020}]%
        {jia2020pushing}
\bibfield{author}{\bibinfo{person}{Weile Jia}, \bibinfo{person}{Han Wang},
  \bibinfo{person}{Mohan Chen}, \bibinfo{person}{Denghui Lu},
  \bibinfo{person}{Lin Lin}, \bibinfo{person}{Roberto Car},
  \bibinfo{person}{Weinan E}, {and} \bibinfo{person}{Linfeng Zhang}.}
  \bibinfo{year}{2020}\natexlab{}.
\newblock \bibinfo{title}{Pushing the Limit of Molecular Dynamics with Ab
  Initio Accuracy to 100 Million Atoms with Machine Learning}.
\newblock
\newblock
\showeprint[arxiv]{2005.00223}~[physics.comp-ph]


\bibitem[\protect\citeauthoryear{Jiang, Li, and Guo}{Jiang
  et~al\mbox{.}}{2016}]%
        {Jiang2016}
\bibfield{author}{\bibinfo{person}{Bin Jiang}, \bibinfo{person}{Jun Li}, {and}
  \bibinfo{person}{Hua Guo}.} \bibinfo{year}{2016}\natexlab{}.
\newblock \showarticletitle{Potential Energy Surfaces from High Fidelity
  Fitting of Ab Initio Points: The Permutation Invariant Polynomial - Neural
  Network Approach}.
\newblock \bibinfo{journal}{\emph{International Reviews in Physical Chemistry}}
  \bibinfo{volume}{35}, \bibinfo{number}{3} (\bibinfo{year}{2016}),
  \bibinfo{pages}{479--506}.
\newblock
\urldef\tempurl%
\url{https://doi.org/10.1080/0144235X.2016.1200347}
\showDOI{\tempurl}
\showeprint{https://doi.org/10.1080/0144235X.2016.1200347}


\bibitem[\protect\citeauthoryear{Jiang, Pajic, Alur, and Mangharam}{Jiang
  et~al\mbox{.}}{2014}]%
        {Jiang2014}
\bibfield{author}{\bibinfo{person}{Zhihao Jiang}, \bibinfo{person}{Miroslav
  Pajic}, \bibinfo{person}{Rajeev Alur}, {and} \bibinfo{person}{Rahul
  Mangharam}.} \bibinfo{year}{2014}\natexlab{}.
\newblock \showarticletitle{Closed-Loop Verification of Medical Devices with
  Model Abstraction and Refinement}.
\newblock \bibinfo{journal}{\emph{International Journal on Software Tools for
  Technology Transfer}} \bibinfo{volume}{16}, \bibinfo{number}{2}
  (\bibinfo{date}{April} \bibinfo{year}{2014}), \bibinfo{pages}{191--213}.
\newblock
\showISSN{1433-2787}
\urldef\tempurl%
\url{https://doi.org/10.1007/s10009-013-0289-7}
\showDOI{\tempurl}


\bibitem[\protect\citeauthoryear{Jin, Zhu, Karniadakis, and Tang}{Jin
  et~al\mbox{.}}{2020}]%
        {SympNets2020}
\bibfield{author}{\bibinfo{person}{Pengzhan Jin}, \bibinfo{person}{Aiqing Zhu},
  \bibinfo{person}{George~Em Karniadakis}, {and} \bibinfo{person}{Yifa Tang}.}
  \bibinfo{year}{2020}\natexlab{}.
\newblock \showarticletitle{Symplectic Networks: Intrinsic Structure-Preserving
  Networks for Identifying {{Hamiltonian}} Systems}.
\newblock \bibinfo{journal}{\emph{CoRR}}  \bibinfo{volume}{abs/2001.03750}
  (\bibinfo{year}{2020}).
\newblock
\showeprint[arxiv]{2001.03750}


\bibitem[\protect\citeauthoryear{Jospin, Buntine, Boussaid, Laga, and
  Bennamoun}{Jospin et~al\mbox{.}}{2020}]%
        {Jospin2020}
\bibfield{author}{\bibinfo{person}{Laurent~Valentin Jospin},
  \bibinfo{person}{Wray Buntine}, \bibinfo{person}{Farid Boussaid},
  \bibinfo{person}{Hamid Laga}, {and} \bibinfo{person}{Mohammed Bennamoun}.}
  \bibinfo{year}{2020}\natexlab{}.
\newblock \showarticletitle{Hands-on {{Bayesian Neural Networks}} -- a
  {{Tutorial}} for {{Deep Learning Users}}}.
\newblock \bibinfo{journal}{\emph{arXiv:2007.06823 [cs, stat]}}
  (\bibinfo{date}{July} \bibinfo{year}{2020}).
\newblock
\showeprint[arxiv]{2007.06823}~[cs, stat]


\bibitem[\protect\citeauthoryear{Karpatne, Atluri, Faghmous, Steinbach,
  Banerjee, Ganguly, Shekhar, Samatova, and Kumar}{Karpatne
  et~al\mbox{.}}{2017}]%
        {Karpatne2017}
\bibfield{author}{\bibinfo{person}{Anuj Karpatne}, \bibinfo{person}{Gowtham
  Atluri}, \bibinfo{person}{James~H. Faghmous}, \bibinfo{person}{Michael
  Steinbach}, \bibinfo{person}{Arindam Banerjee}, \bibinfo{person}{Auroop
  Ganguly}, \bibinfo{person}{Shashi Shekhar}, \bibinfo{person}{Nagiza
  Samatova}, {and} \bibinfo{person}{Vipin Kumar}.}
  \bibinfo{year}{2017}\natexlab{}.
\newblock \showarticletitle{Theory-{{Guided Data Science}}: A {{New Paradigm}}
  for {{Scientific Discovery}} from {{Data}}}.
\newblock \bibinfo{journal}{\emph{IEEE Transactions on Knowledge and Data
  Engineering}} \bibinfo{volume}{29}, \bibinfo{number}{10}
  (\bibinfo{date}{Oct.} \bibinfo{year}{2017}), \bibinfo{pages}{2318--2331}.
\newblock
\showISSN{1041-4347}
\urldef\tempurl%
\url{https://doi.org/10.1109/TKDE.2017.2720168}
\showDOI{\tempurl}


\bibitem[\protect\citeauthoryear{Kelly, Bettencourt, Johnson, and
  Duvenaud}{Kelly et~al\mbox{.}}{2020}]%
        {kelly2020learning}
\bibfield{author}{\bibinfo{person}{Jacob Kelly}, \bibinfo{person}{Jesse
  Bettencourt}, \bibinfo{person}{Matthew~James Johnson}, {and}
  \bibinfo{person}{David Duvenaud}.} \bibinfo{year}{2020}\natexlab{}.
\newblock \bibinfo{title}{Learning Differential Equations That Are Easy to
  Solve}.
\newblock
\newblock
\showeprint[arxiv]{2007.04504}~[cs.LG]


\bibitem[\protect\citeauthoryear{Kerschen, Worden, Vakakis, and
  Golinval}{Kerschen et~al\mbox{.}}{2006}]%
        {KERSCHEN2006505}
\bibfield{author}{\bibinfo{person}{Ga{\"e}tan Kerschen}, \bibinfo{person}{Keith
  Worden}, \bibinfo{person}{Alexander~F. Vakakis}, {and}
  \bibinfo{person}{Jean-Claude Golinval}.} \bibinfo{year}{2006}\natexlab{}.
\newblock \showarticletitle{Past, Present and Future of Nonlinear System
  Identification in Structural Dynamics}.
\newblock \bibinfo{journal}{\emph{Mechanical Systems and Signal Processing}}
  \bibinfo{volume}{20}, \bibinfo{number}{3} (\bibinfo{year}{2006}),
  \bibinfo{pages}{505--592}.
\newblock
\showISSN{0888-3270}
\urldef\tempurl%
\url{https://doi.org/10.1016/j.ymssp.2005.04.008}
\showDOI{\tempurl}


\bibitem[\protect\citeauthoryear{Kidger, Chen, and Lyons}{Kidger
  et~al\mbox{.}}{2020}]%
        {Kidger2020}
\bibfield{author}{\bibinfo{person}{Patrick Kidger}, \bibinfo{person}{Ricky
  T.~Q. Chen}, {and} \bibinfo{person}{Terry Lyons}.}
  \bibinfo{year}{2020}\natexlab{}.
\newblock \showarticletitle{"{{Hey}}, That's Not an {{ODE}}": Faster {{ODE
  Adjoints}} with 12 {{Lines}} of {{Code}}}.
\newblock \bibinfo{journal}{\emph{arXiv:2009.09457 [cs, math]}}
  (\bibinfo{date}{Sept.} \bibinfo{year}{2020}).
\newblock
\showeprint[arxiv]{2009.09457}~[cs, math]


\bibitem[\protect\citeauthoryear{Kipf, Fetaya, Wang, Welling, and Zemel}{Kipf
  et~al\mbox{.}}{2018}]%
        {kipf2018neural}
\bibfield{author}{\bibinfo{person}{Thomas Kipf}, \bibinfo{person}{Ethan
  Fetaya}, \bibinfo{person}{Kuan-Chieh Wang}, \bibinfo{person}{Max Welling},
  {and} \bibinfo{person}{Richard Zemel}.} \bibinfo{year}{2018}\natexlab{}.
\newblock \bibinfo{title}{Neural Relational Inference for Interacting Systems}.
\newblock
\newblock
\showeprint[arxiv]{1802.04687}~[stat.ML]


\bibitem[\protect\citeauthoryear{Kloeden and Platen}{Kloeden and
  Platen}{1992}]%
        {Kloeden1992}
\bibfield{author}{\bibinfo{person}{Peter~E Kloeden} {and}
  \bibinfo{person}{Eckhard Platen}.} \bibinfo{year}{1992}\natexlab{}.
\newblock \bibinfo{booktitle}{\emph{Numerical {{Solution}} of {{Stochastic
  Differential Equations}}}}.
\newblock
\showISBNx{978-3-662-12616-5}


\bibitem[\protect\citeauthoryear{Kofman and Junco}{Kofman and Junco}{2001}]%
        {Kofman2001}
\bibfield{author}{\bibinfo{person}{Ernesto Kofman} {and}
  \bibinfo{person}{Sergio Junco}.} \bibinfo{year}{2001}\natexlab{}.
\newblock \showarticletitle{Quantized-State Systems: A {{DEVS Approach}} for
  Continuous System Simulation}.
\newblock \bibinfo{journal}{\emph{Transactions of The Society for Modeling and
  Simulation International}} \bibinfo{volume}{18}, \bibinfo{number}{3}
  (\bibinfo{year}{2001}), \bibinfo{pages}{123--132}.
\newblock
\showISSN{0740-6797}


\bibitem[\protect\citeauthoryear{Koziel and {Pietrenko-Dabrowska}}{Koziel and
  {Pietrenko-Dabrowska}}{2020}]%
        {Koziel2020}
\bibfield{author}{\bibinfo{person}{Slawomir Koziel} {and} \bibinfo{person}{Anna
  {Pietrenko-Dabrowska}}.} \bibinfo{year}{2020}\natexlab{}.
\newblock \bibinfo{booktitle}{\emph{Basics of {{Data}}-{{Driven Surrogate
  Modeling}}}}.
\newblock \bibinfo{publisher}{{Springer International Publishing}},
  \bibinfo{address}{{Cham}}, \bibinfo{pages}{23--58}.
\newblock
\showISBNx{978-3-030-38925-3 978-3-030-38926-0}
\urldef\tempurl%
\url{https://doi.org/10.1007/978-3-030-38926-0_2}
\showDOI{\tempurl}


\bibitem[\protect\citeauthoryear{Krishnan, Shalit, and Sontag}{Krishnan
  et~al\mbox{.}}{2017}]%
        {Krishnan2017StructuredIN}
\bibfield{author}{\bibinfo{person}{R. Krishnan}, \bibinfo{person}{U. Shalit},
  {and} \bibinfo{person}{D. Sontag}.} \bibinfo{year}{2017}\natexlab{}.
\newblock \showarticletitle{Structured Inference Networks for Nonlinear State
  Space Models}. In \bibinfo{booktitle}{\emph{{{AAAI}}}}.
\newblock


\bibitem[\protect\citeauthoryear{Krishnan, Shalit, and Sontag}{Krishnan
  et~al\mbox{.}}{2015}]%
        {krishnan2015deep}
\bibfield{author}{\bibinfo{person}{Rahul~G. Krishnan}, \bibinfo{person}{Uri
  Shalit}, {and} \bibinfo{person}{David Sontag}.}
  \bibinfo{year}{2015}\natexlab{}.
\newblock \bibinfo{title}{Deep Kalman Filters}.
\newblock
\newblock
\showeprint[arxiv]{1511.05121}~[stat.ML]


\bibitem[\protect\citeauthoryear{Krishnan, Shalit, and Sontag}{Krishnan
  et~al\mbox{.}}{2016a}]%
        {krishnan2016structured}
\bibfield{author}{\bibinfo{person}{Rahul~G. Krishnan}, \bibinfo{person}{Uri
  Shalit}, {and} \bibinfo{person}{David Sontag}.}
  \bibinfo{year}{2016}\natexlab{a}.
\newblock \bibinfo{title}{Structured Inference Networks for Nonlinear State
  Space Models}.
\newblock
\newblock
\showeprint[arxiv]{1609.09869}~[stat.ML]


\bibitem[\protect\citeauthoryear{Krishnan, Shalit, and Sontag}{Krishnan
  et~al\mbox{.}}{2016b}]%
        {Krishnan2016}
\bibfield{author}{\bibinfo{person}{Rahul~G. Krishnan}, \bibinfo{person}{Uri
  Shalit}, {and} \bibinfo{person}{David Sontag}.}
  \bibinfo{year}{2016}\natexlab{b}.
\newblock \showarticletitle{Structured {{Inference Networks}} for {{Nonlinear
  State Space Models}}}.
\newblock \bibinfo{journal}{\emph{arXiv:1609.09869 [cs, stat]}}
  (\bibinfo{date}{Dec.} \bibinfo{year}{2016}).
\newblock
\showeprint[arxiv]{1609.09869}~[cs, stat]


\bibitem[\protect\citeauthoryear{Kroll and Schulte}{Kroll and Schulte}{2014}]%
        {KROLL2014496}
\bibfield{author}{\bibinfo{person}{Andreas Kroll} {and} \bibinfo{person}{Horst
  Schulte}.} \bibinfo{year}{2014}\natexlab{}.
\newblock \showarticletitle{Benchmark Problems for Nonlinear System
  Identification and Control Using {{Soft Computing}} Methods: Need and
  Overview}.
\newblock \bibinfo{journal}{\emph{Applied Soft Computing}}
  \bibinfo{volume}{25} (\bibinfo{year}{2014}), \bibinfo{pages}{496--513}.
\newblock
\showISSN{1568-4946}
\urldef\tempurl%
\url{https://doi.org/10.1016/j.asoc.2014.08.034}
\showDOI{\tempurl}


\bibitem[\protect\citeauthoryear{Lee and Parish}{Lee and Parish}{2021}]%
        {Lee2021}
\bibfield{author}{\bibinfo{person}{Kookjin Lee} {and} \bibinfo{person}{Eric~J.
  Parish}.} \bibinfo{year}{2021}\natexlab{}.
\newblock \showarticletitle{Parameterized Neural Ordinary Differential
  Equations: Applications to Computational Physics Problems}.
\newblock \bibinfo{journal}{\emph{Proceedings of the Royal Society A:
  Mathematical, Physical and Engineering Sciences}} \bibinfo{volume}{477},
  \bibinfo{number}{2253} (\bibinfo{date}{Sept.} \bibinfo{year}{2021}),
  \bibinfo{pages}{20210162}.
\newblock
\urldef\tempurl%
\url{https://doi.org/10.1098/rspa.2021.0162}
\showDOI{\tempurl}


\bibitem[\protect\citeauthoryear{Lenz, Knepper, and Saxena}{Lenz
  et~al\mbox{.}}{2015}]%
        {Lenz2015DeepMPCLD}
\bibfield{author}{\bibinfo{person}{I. Lenz}, \bibinfo{person}{Ross~A. Knepper},
  {and} \bibinfo{person}{A. Saxena}.} \bibinfo{year}{2015}\natexlab{}.
\newblock \showarticletitle{{{DeepMPC}}: Learning Deep Latent Features for
  Model Predictive Control}. In \bibinfo{booktitle}{\emph{Robotics: Science and
  Systems}}.
\newblock


\bibitem[\protect\citeauthoryear{LeVeque}{LeVeque}{2007}]%
        {LeVeque2007}
\bibfield{author}{\bibinfo{person}{Randall~J LeVeque}.}
  \bibinfo{year}{2007}\natexlab{}.
\newblock \bibinfo{booktitle}{\emph{Finite Difference Methods for Ordinary and
  Partial Differential Equations: Steady-State and Time-Dependent Problems}}.
  Vol.~\bibinfo{volume}{98}.
\newblock \bibinfo{publisher}{{Siam}}.
\newblock
\showISBNx{0-89871-629-2}


\bibitem[\protect\citeauthoryear{Li, Wong, Chen, and Duvenaud}{Li
  et~al\mbox{.}}{2020}]%
        {li2020scalable}
\bibfield{author}{\bibinfo{person}{Xuechen Li},
  \bibinfo{person}{Ting-Kam~Leonard Wong}, \bibinfo{person}{Ricky T.~Q. Chen},
  {and} \bibinfo{person}{David Duvenaud}.} \bibinfo{year}{2020}\natexlab{}.
\newblock \bibinfo{title}{Scalable Gradients for Stochastic Differential
  Equations}.
\newblock
\newblock
\showeprint[arxiv]{2001.01328}~[cs.LG]


\bibitem[\protect\citeauthoryear{Li, Wu, Tedrake, Tenenbaum, and Torralba}{Li
  et~al\mbox{.}}{2018a}]%
        {Yunzhu2018}
\bibfield{author}{\bibinfo{person}{Yunzhu Li}, \bibinfo{person}{Jiajun Wu},
  \bibinfo{person}{Russ Tedrake}, \bibinfo{person}{Joshua~B. Tenenbaum}, {and}
  \bibinfo{person}{Antonio Torralba}.} \bibinfo{year}{2018}\natexlab{a}.
\newblock \showarticletitle{Learning Particle Dynamics for Manipulating Rigid
  Bodies, Deformable Objects, and Fluids}.
\newblock \bibinfo{journal}{\emph{CoRR}}  \bibinfo{volume}{abs/1810.01566}
  (\bibinfo{year}{2018}).
\newblock
\showeprint[arxiv]{1810.01566}


\bibitem[\protect\citeauthoryear{Li, Wu, Zhu, Tenenbaum, Torralba, and
  Tedrake}{Li et~al\mbox{.}}{2018b}]%
        {propagation2018}
\bibfield{author}{\bibinfo{person}{Yunzhu Li}, \bibinfo{person}{Jiajun Wu},
  \bibinfo{person}{Jun-Yan Zhu}, \bibinfo{person}{Joshua~B. Tenenbaum},
  \bibinfo{person}{Antonio Torralba}, {and} \bibinfo{person}{Russ Tedrake}.}
  \bibinfo{year}{2018}\natexlab{b}.
\newblock \showarticletitle{Propagation Networks for Model-Based Control under
  Partial Observation}.
\newblock \bibinfo{journal}{\emph{CoRR}}  \bibinfo{volume}{abs/1809.11169}
  (\bibinfo{year}{2018}).
\newblock
\showeprint[arxiv]{1809.11169}


\bibitem[\protect\citeauthoryear{Liu, Honor{\'e}, Chatterjee, and
  Rasmussen}{Liu et~al\mbox{.}}{2019a}]%
        {liu2019powering}
\bibfield{author}{\bibinfo{person}{Dong Liu}, \bibinfo{person}{Antoine
  Honor{\'e}}, \bibinfo{person}{Saikat Chatterjee}, {and}
  \bibinfo{person}{Lars~K Rasmussen}.} \bibinfo{year}{2019}\natexlab{a}.
\newblock \showarticletitle{Powering Hidden Markov Model by Neural Network
  Based Generative Models}.
\newblock \bibinfo{journal}{\emph{arXiv preprint arXiv:1910.05744}}
  (\bibinfo{year}{2019}).
\newblock
\showeprint[arxiv]{1910.05744}


\bibitem[\protect\citeauthoryear{Liu, Xiao, Si, Cao, Kumar, and Hsieh}{Liu
  et~al\mbox{.}}{2019b}]%
        {liu2019neural}
\bibfield{author}{\bibinfo{person}{Xuanqing Liu}, \bibinfo{person}{Tesi Xiao},
  \bibinfo{person}{Si Si}, \bibinfo{person}{Qin Cao}, \bibinfo{person}{Sanjiv
  Kumar}, {and} \bibinfo{person}{Cho-Jui Hsieh}.}
  \bibinfo{year}{2019}\natexlab{b}.
\newblock \bibinfo{title}{Neural {{SDE}}: Stabilizing Neural {{ODE}} Networks
  with Stochastic Noise}.
\newblock
\newblock
\showeprint[arxiv]{1906.02355}~[cs.LG]


\bibitem[\protect\citeauthoryear{Liu, Xiao, Si, Cao, Kumar, and Hsieh}{Liu
  et~al\mbox{.}}{2019c}]%
        {Liu2019}
\bibfield{author}{\bibinfo{person}{Xuanqing Liu}, \bibinfo{person}{Tesi Xiao},
  \bibinfo{person}{Si Si}, \bibinfo{person}{Qin Cao}, \bibinfo{person}{Sanjiv
  Kumar}, {and} \bibinfo{person}{Cho-Jui Hsieh}.}
  \bibinfo{year}{2019}\natexlab{c}.
\newblock \showarticletitle{Neural {{SDE}}: Stabilizing {{Neural ODE Networks}}
  with {{Stochastic Noise}}}.
\newblock \bibinfo{journal}{\emph{arXiv:1906.02355 [cs, stat]}}
  (\bibinfo{date}{June} \bibinfo{year}{2019}).
\newblock
\showeprint[arxiv]{1906.02355}~[cs, stat]


\bibitem[\protect\citeauthoryear{Ljung}{Ljung}{2006}]%
        {Ljung2006}
\bibfield{author}{\bibinfo{person}{Lennart Ljung}.}
  \bibinfo{year}{2006}\natexlab{}.
\newblock \showarticletitle{Some Aspects of Non Linear System Identification}.
\newblock \bibinfo{journal}{\emph{IFAC Proceedings Volumes}}
  \bibinfo{volume}{39}, \bibinfo{number}{1} (\bibinfo{year}{2006}),
  \bibinfo{pages}{110--121}.
\newblock
\showISSN{14746670}
\urldef\tempurl%
\url{https://doi.org/10.3182/20060329-3-AU-2901.00009}
\showDOI{\tempurl}


\bibitem[\protect\citeauthoryear{Lutter, Ritter, and Peters}{Lutter
  et~al\mbox{.}}{2019}]%
        {Lutter2019}
\bibfield{author}{\bibinfo{person}{Michael Lutter}, \bibinfo{person}{Christian
  Ritter}, {and} \bibinfo{person}{Jan Peters}.}
  \bibinfo{year}{2019}\natexlab{}.
\newblock \showarticletitle{Deep {{Lagrangian Networks}}: Using {{Physics}} as
  {{Model Prior}} for {{Deep Learning}}}.
\newblock \bibinfo{journal}{\emph{arXiv:1907.04490 [cs, eess, stat]}}
  (\bibinfo{date}{July} \bibinfo{year}{2019}).
\newblock
\showeprint[arxiv]{1907.04490}~[cs, eess, stat]


\bibitem[\protect\citeauthoryear{Marsden and West}{Marsden and West}{2001}]%
        {Marsden2001}
\bibfield{author}{\bibinfo{person}{J.~E. Marsden} {and} \bibinfo{person}{M.
  West}.} \bibinfo{year}{2001}\natexlab{}.
\newblock \showarticletitle{Discrete Mechanics and Variational Integrators}.
\newblock \bibinfo{journal}{\emph{Acta Numerica}}  \bibinfo{volume}{10}
  (\bibinfo{date}{May} \bibinfo{year}{2001}), \bibinfo{pages}{357--514}.
\newblock
\showISSN{0962-4929, 1474-0508}
\urldef\tempurl%
\url{https://doi.org/10.1017/S096249290100006X}
\showDOI{\tempurl}


\bibitem[\protect\citeauthoryear{Massaroli, Poli, Bin, Park, Yamashita, and
  Asama}{Massaroli et~al\mbox{.}}{2020}]%
        {massaroli2020stable}
\bibfield{author}{\bibinfo{person}{Stefano Massaroli}, \bibinfo{person}{Michael
  Poli}, \bibinfo{person}{Michelangelo Bin}, \bibinfo{person}{Jinkyoo Park},
  \bibinfo{person}{Atsushi Yamashita}, {and} \bibinfo{person}{Hajime Asama}.}
  \bibinfo{year}{2020}\natexlab{}.
\newblock \bibinfo{title}{Stable Neural Flows}.
\newblock
\newblock
\showeprint[arxiv]{2003.08063}~[cs.LG]


\bibitem[\protect\citeauthoryear{Massaroli, Poli, Park, Yamashita, and
  Asama}{Massaroli et~al\mbox{.}}{2021}]%
        {massaroli2021dissecting}
\bibfield{author}{\bibinfo{person}{Stefano Massaroli}, \bibinfo{person}{Michael
  Poli}, \bibinfo{person}{Jinkyoo Park}, \bibinfo{person}{Atsushi Yamashita},
  {and} \bibinfo{person}{Hajime Asama}.} \bibinfo{year}{2021}\natexlab{}.
\newblock \bibinfo{title}{Dissecting Neural {{ODEs}}}.
\newblock
\newblock
\showeprint[arxiv]{2002.08071}~[cs.LG]


\bibitem[\protect\citeauthoryear{Masti and Bemporad}{Masti and
  Bemporad}{2018}]%
        {MastiCDC2018}
\bibfield{author}{\bibinfo{person}{D. Masti} {and} \bibinfo{person}{A.
  Bemporad}.} \bibinfo{year}{2018}\natexlab{}.
\newblock \showarticletitle{Learning Nonlinear State-Space Models Using Deep
  Autoencoders}. In \bibinfo{booktitle}{\emph{2018 {{IEEE}} Conference on
  Decision and Control ({{CDC}})}}. \bibinfo{pages}{3862--3867}.
\newblock


\bibitem[\protect\citeauthoryear{Mittal and Vaishay}{Mittal and
  Vaishay}{2019}]%
        {Mittal2019}
\bibfield{author}{\bibinfo{person}{Sparsh Mittal} {and}
  \bibinfo{person}{Shraiysh Vaishay}.} \bibinfo{year}{2019}\natexlab{}.
\newblock \showarticletitle{A Survey of Techniques for Optimizing Deep Learning
  on {{GPUs}}}.
\newblock \bibinfo{journal}{\emph{Journal of Systems Architecture}}
  \bibinfo{volume}{99} (\bibinfo{date}{Oct.} \bibinfo{year}{2019}),
  \bibinfo{pages}{101635}.
\newblock
\showISSN{13837621}
\urldef\tempurl%
\url{https://doi.org/10.1016/j.sysarc.2019.101635}
\showDOI{\tempurl}


\bibitem[\protect\citeauthoryear{Montanez, Amizadeh, and Laptev}{Montanez
  et~al\mbox{.}}{2015}]%
        {montanez2015inertial}
\bibfield{author}{\bibinfo{person}{George Montanez}, \bibinfo{person}{Saeed
  Amizadeh}, {and} \bibinfo{person}{Nikolay Laptev}.}
  \bibinfo{year}{2015}\natexlab{}.
\newblock \showarticletitle{Inertial Hidden Markov Models: Modeling Change in
  Multivariate Time Series}. In \bibinfo{booktitle}{\emph{Proceedings of the
  {{AAAI}} Conference on Artificial Intelligence}}, Vol.~\bibinfo{volume}{29}.
\newblock


\bibitem[\protect\citeauthoryear{Moradi, Gomes, Oakes, and Denil}{Moradi
  et~al\mbox{.}}{2019}]%
        {Moradi2019}
\bibfield{author}{\bibinfo{person}{Mehrdad Moradi},
  \bibinfo{person}{Cl{\'a}udio Gomes}, \bibinfo{person}{Bentley~James Oakes},
  {and} \bibinfo{person}{Joachim Denil}.} \bibinfo{year}{2019}\natexlab{}.
\newblock \showarticletitle{Optimizing {{Fault Injection}} in {{FMI
  Co}}-Simulation}. In \bibinfo{booktitle}{\emph{Proceedings of the 2019
  {{Summer Simulation Conference}}}}. \bibinfo{publisher}{{Society for Computer
  Simulation International}}, \bibinfo{address}{{Berlin, Germany}},
  \bibinfo{pages}{12}.
\newblock
\urldef\tempurl%
\url{https://doi.org/10.5555/3374138.3374170}
\showDOI{\tempurl}


\bibitem[\protect\citeauthoryear{Murphy}{Murphy}{2012}]%
        {Murphy2012}
\bibfield{author}{\bibinfo{person}{Kevin~P Murphy}.}
  \bibinfo{year}{2012}\natexlab{}.
\newblock \bibinfo{booktitle}{\emph{Machine Learning: A Probabilistic
  Perspective}}.
\newblock \bibinfo{publisher}{{MIT press}}.
\newblock


\bibitem[\protect\citeauthoryear{Mustafa, Allen, and Appiah}{Mustafa
  et~al\mbox{.}}{2019}]%
        {mustafa2019comparative}
\bibfield{author}{\bibinfo{person}{Mohammed~Kyari Mustafa},
  \bibinfo{person}{Tony Allen}, {and} \bibinfo{person}{Kofi Appiah}.}
  \bibinfo{year}{2019}\natexlab{}.
\newblock \showarticletitle{A Comparative Review of Dynamic Neural Networks and
  Hidden {{Markov}} Model Methods for Mobile On-Device Speech Recognition}.
\newblock \bibinfo{journal}{\emph{Neural Computing and Applications}}
  \bibinfo{volume}{31}, \bibinfo{number}{2} (\bibinfo{year}{2019}),
  \bibinfo{pages}{891--899}.
\newblock


\bibitem[\protect\citeauthoryear{Nelles}{Nelles}{2001}]%
        {Nelles2001}
\bibfield{author}{\bibinfo{person}{Oliver Nelles}.}
  \bibinfo{year}{2001}\natexlab{}.
\newblock \bibinfo{booktitle}{\emph{Nonlinear {{System Identification}}: From
  {{Classical Approaches}} to {{Neural Networks}} and {{Fuzzy Models}}}}.
\newblock \bibinfo{publisher}{{Springer-Verlag}}, \bibinfo{address}{{Berlin
  Heidelberg}}.
\newblock
\showISBNx{978-3-540-67369-9}
\urldef\tempurl%
\url{https://doi.org/10.1007/978-3-662-04323-3}
\showDOI{\tempurl}


\bibitem[\protect\citeauthoryear{Norcliffe, Bodnar, Day, Simidjievski, and
  Li{\`o}}{Norcliffe et~al\mbox{.}}{2020}]%
        {norcliffe2020second}
\bibfield{author}{\bibinfo{person}{Alexander Norcliffe},
  \bibinfo{person}{Cristian Bodnar}, \bibinfo{person}{Ben Day},
  \bibinfo{person}{Nikola Simidjievski}, {and} \bibinfo{person}{Pietro
  Li{\`o}}.} \bibinfo{year}{2020}\natexlab{}.
\newblock \bibinfo{title}{On Second Order Behaviour in Augmented Neural
  {{ODEs}}}.
\newblock
\newblock
\showeprint[arxiv]{2006.07220}~[cs.LG]


\bibitem[\protect\citeauthoryear{Oganesyan, Volokhova, and Vetrov}{Oganesyan
  et~al\mbox{.}}{2020}]%
        {Oganesyan2020}
\bibfield{author}{\bibinfo{person}{Viktor Oganesyan},
  \bibinfo{person}{Alexandra Volokhova}, {and} \bibinfo{person}{Dmitry
  Vetrov}.} \bibinfo{year}{2020}\natexlab{}.
\newblock \showarticletitle{Stochasticity in {{Neural ODEs}}: An {{Empirical
  Study}}}.
\newblock \bibinfo{journal}{\emph{arXiv:2002.09779 [cs, stat]}}
  (\bibinfo{date}{June} \bibinfo{year}{2020}).
\newblock
\showeprint[arxiv]{2002.09779}~[cs, stat]


\bibitem[\protect\citeauthoryear{Ogunmolu, Gu, Jiang, and Gans}{Ogunmolu
  et~al\mbox{.}}{2016a}]%
        {ogunmolu_nonlinear_2016}
\bibfield{author}{\bibinfo{person}{Olalekan Ogunmolu}, \bibinfo{person}{Xuejun
  Gu}, \bibinfo{person}{Steve Jiang}, {and} \bibinfo{person}{Nicholas Gans}.}
  \bibinfo{year}{2016}\natexlab{a}.
\newblock \showarticletitle{Nonlinear {{Systems Identification Using Deep
  Dynamic Neural Networks}}}.
\newblock \bibinfo{journal}{\emph{arXiv:1610.01439 [cs]}} (\bibinfo{date}{Oct.}
  \bibinfo{year}{2016}).
\newblock
\showeprint[arxiv]{1610.01439}~[cs]


\bibitem[\protect\citeauthoryear{Ogunmolu, Gu, Jiang, and Gans}{Ogunmolu
  et~al\mbox{.}}{2016b}]%
        {OgunmoluGJG16}
\bibfield{author}{\bibinfo{person}{Olalekan~P. Ogunmolu},
  \bibinfo{person}{Xuejun Gu}, \bibinfo{person}{Steve~B. Jiang}, {and}
  \bibinfo{person}{Nicholas~R. Gans}.} \bibinfo{year}{2016}\natexlab{b}.
\newblock \showarticletitle{Nonlinear Systems Identification Using Deep Dynamic
  Neural Networks}.
\newblock \bibinfo{journal}{\emph{CoRR}}  \bibinfo{volume}{abs/1610.01439}
  (\bibinfo{year}{2016}).
\newblock
\showeprint[arxiv]{1610.01439}


\bibitem[\protect\citeauthoryear{Ott, Katiyar, Hennig, and Tiemann}{Ott
  et~al\mbox{.}}{2020}]%
        {Ott2020}
\bibfield{author}{\bibinfo{person}{Katharina Ott}, \bibinfo{person}{Prateek
  Katiyar}, \bibinfo{person}{Philipp Hennig}, {and} \bibinfo{person}{Michael
  Tiemann}.} \bibinfo{year}{2020}\natexlab{}.
\newblock \showarticletitle{When Are {{Neural ODE Solutions Proper ODEs}}?}
\newblock \bibinfo{journal}{\emph{arXiv:2007.15386 [cs, stat]}}
  (\bibinfo{date}{July} \bibinfo{year}{2020}).
\newblock
\showeprint[arxiv]{2007.15386}~[cs, stat]


\bibitem[\protect\citeauthoryear{Paszke, Gross, Massa, Lerer, Bradbury, Chanan,
  Killeen, Lin, Gimelshein, Antiga, Desmaison, Kopf, Yang, DeVito, Raison,
  Tejani, Chilamkurthy, Steiner, Fang, Bai, and Chintala}{Paszke
  et~al\mbox{.}}{2019}]%
        {Paszke2019}
\bibfield{author}{\bibinfo{person}{Adam Paszke}, \bibinfo{person}{Sam Gross},
  \bibinfo{person}{Francisco Massa}, \bibinfo{person}{Adam Lerer},
  \bibinfo{person}{James Bradbury}, \bibinfo{person}{Gregory Chanan},
  \bibinfo{person}{Trevor Killeen}, \bibinfo{person}{Zeming Lin},
  \bibinfo{person}{Natalia Gimelshein}, \bibinfo{person}{Luca Antiga},
  \bibinfo{person}{Alban Desmaison}, \bibinfo{person}{Andreas Kopf},
  \bibinfo{person}{Edward Yang}, \bibinfo{person}{Zachary DeVito},
  \bibinfo{person}{Martin Raison}, \bibinfo{person}{Alykhan Tejani},
  \bibinfo{person}{Sasank Chilamkurthy}, \bibinfo{person}{Benoit Steiner},
  \bibinfo{person}{Lu Fang}, \bibinfo{person}{Junjie Bai}, {and}
  \bibinfo{person}{Soumith Chintala}.} \bibinfo{year}{2019}\natexlab{}.
\newblock \showarticletitle{{{PyTorch}}: An Imperative Style, High-Performance
  Deep Learning Library}.
\newblock In \bibinfo{booktitle}{\emph{Advances in Neural Information
  Processing Systems 32}}, \bibfield{editor}{\bibinfo{person}{H.~Wallach},
  \bibinfo{person}{H.~Larochelle}, \bibinfo{person}{A.~Beygelzimer},
  \bibinfo{person}{F.~{dAlch{\'e}-Buc}}, \bibinfo{person}{E.~Fox}, {and}
  \bibinfo{person}{R.~Garnett}} (Eds.). \bibinfo{publisher}{{Curran Associates,
  Inc.}}, \bibinfo{pages}{8026--8037}.
\newblock


\bibitem[\protect\citeauthoryear{Pintard, Fabre, Kanoun, Leeman, and
  Roy}{Pintard et~al\mbox{.}}{2013}]%
        {Pintard2013}
\bibfield{author}{\bibinfo{person}{Ludovic Pintard},
  \bibinfo{person}{Jean-Charles Fabre}, \bibinfo{person}{Karama Kanoun},
  \bibinfo{person}{Michel Leeman}, {and} \bibinfo{person}{Matthieu Roy}.}
  \bibinfo{year}{2013}\natexlab{}.
\newblock \showarticletitle{Fault {{Injection}} in the {{Automotive Standard
  ISO}} 26262: An {{Initial Approach}}}.
\newblock In \bibinfo{booktitle}{\emph{Dependable {{Computing}}}},
  \bibfield{editor}{\bibinfo{person}{David Hutchison}, \bibinfo{person}{Takeo
  Kanade}, \bibinfo{person}{Josef Kittler}, \bibinfo{person}{Jon~M. Kleinberg},
  \bibinfo{person}{Friedemann Mattern}, \bibinfo{person}{John~C. Mitchell},
  \bibinfo{person}{Moni Naor}, \bibinfo{person}{Oscar Nierstrasz},
  \bibinfo{person}{C.~Pandu~Rangan}, \bibinfo{person}{Bernhard Steffen},
  \bibinfo{person}{Madhu Sudan}, \bibinfo{person}{Demetri Terzopoulos},
  \bibinfo{person}{Doug Tygar}, \bibinfo{person}{Moshe~Y. Vardi},
  \bibinfo{person}{Gerhard Weikum}, \bibinfo{person}{Marco Vieira}, {and}
  \bibinfo{person}{Jo{\~a}o~Carlos Cunha}} (Eds.). Vol.~\bibinfo{volume}{7869}.
  \bibinfo{publisher}{{Springer Berlin Heidelberg}}, \bibinfo{address}{{Berlin,
  Heidelberg}}, \bibinfo{pages}{126--133}.
\newblock
\showISBNx{978-3-642-38788-3 978-3-642-38789-0}
\urldef\tempurl%
\url{https://doi.org/10.1007/978-3-642-38789-0_11}
\showDOI{\tempurl}


\bibitem[\protect\citeauthoryear{Plebe and Grasso}{Plebe and Grasso}{2019}]%
        {Plebe2019}
\bibfield{author}{\bibinfo{person}{Alessio Plebe} {and}
  \bibinfo{person}{Giorgio Grasso}.} \bibinfo{year}{2019}\natexlab{}.
\newblock \showarticletitle{The {{Unbearable Shallow Understanding}} of {{Deep
  Learning}}}.
\newblock \bibinfo{journal}{\emph{Minds and Machines}} \bibinfo{volume}{29},
  \bibinfo{number}{4} (\bibinfo{date}{Dec.} \bibinfo{year}{2019}),
  \bibinfo{pages}{515--553}.
\newblock
\showISSN{1572-8641}
\urldef\tempurl%
\url{https://doi.org/10.1007/s11023-019-09512-8}
\showDOI{\tempurl}


\bibitem[\protect\citeauthoryear{Poli, Massaroli, Park, Yamashita, Asama, and
  Park}{Poli et~al\mbox{.}}{2020}]%
        {poli2020graph}
\bibfield{author}{\bibinfo{person}{Michael Poli}, \bibinfo{person}{Stefano
  Massaroli}, \bibinfo{person}{Junyoung Park}, \bibinfo{person}{Atsushi
  Yamashita}, \bibinfo{person}{Hajime Asama}, {and} \bibinfo{person}{Jinkyoo
  Park}.} \bibinfo{year}{2020}\natexlab{}.
\newblock \bibinfo{title}{Graph Neural Ordinary Differential Equations}.
\newblock
\newblock
\showeprint[arxiv]{1911.07532}~[cs.LG]


\bibitem[\protect\citeauthoryear{Qin, Wu, and Xiu}{Qin et~al\mbox{.}}{2019}]%
        {Qin2019}
\bibfield{author}{\bibinfo{person}{Tong Qin}, \bibinfo{person}{Kailiang Wu},
  {and} \bibinfo{person}{Dongbin Xiu}.} \bibinfo{year}{2019}\natexlab{}.
\newblock \showarticletitle{Data Driven Governing Equations Approximation Using
  Deep Neural Networks}.
\newblock \bibinfo{journal}{\emph{J. Comput. Phys.}}  \bibinfo{volume}{395}
  (\bibinfo{date}{Oct.} \bibinfo{year}{2019}), \bibinfo{pages}{620--635}.
\newblock
\showISSN{0021-9991}
\urldef\tempurl%
\url{https://doi.org/10.1016/j.jcp.2019.06.042}
\showDOI{\tempurl}


\bibitem[\protect\citeauthoryear{Qu, Bengio, and Tang}{Qu
  et~al\mbox{.}}{2019}]%
        {qu2019gmnn}
\bibfield{author}{\bibinfo{person}{Meng Qu}, \bibinfo{person}{Yoshua Bengio},
  {and} \bibinfo{person}{Jian Tang}.} \bibinfo{year}{2019}\natexlab{}.
\newblock \showarticletitle{Gmnn: Graph Markov Neural Networks}. In
  \bibinfo{booktitle}{\emph{International Conference on Machine Learning}}.
  {PMLR}, \bibinfo{pages}{5241--5250}.
\newblock


\bibitem[\protect\citeauthoryear{Quaglino, Gallieri, Masci, and
  Koutn{\'i}k}{Quaglino et~al\mbox{.}}{2020}]%
        {quaglino2020snode}
\bibfield{author}{\bibinfo{person}{Alessio Quaglino}, \bibinfo{person}{Marco
  Gallieri}, \bibinfo{person}{Jonathan Masci}, {and} \bibinfo{person}{Jan
  Koutn{\'i}k}.} \bibinfo{year}{2020}\natexlab{}.
\newblock \bibinfo{title}{{{SNODE}}: Spectral Discretization of Neural {{ODEs}}
  for System Identification}.
\newblock
\newblock
\showeprint[arxiv]{1906.07038}~[cs.NE]


\bibitem[\protect\citeauthoryear{Rai and Sahu}{Rai and Sahu}{2020}]%
        {Rai2020}
\bibfield{author}{\bibinfo{person}{R. Rai} {and} \bibinfo{person}{C.~K. Sahu}.}
  \bibinfo{year}{2020}\natexlab{}.
\newblock \showarticletitle{Driven by Data or Derived through Physics? A Review
  of Hybrid Physics Guided Machine Learning Techniques with Cyber-Physical
  System ({{CPS}}) Focus}.
\newblock \bibinfo{journal}{\emph{IEEE Access}}  \bibinfo{volume}{8}
  (\bibinfo{year}{2020}), \bibinfo{pages}{71050--71073}.
\newblock


\bibitem[\protect\citeauthoryear{Raissi, Perdikaris, and Karniadakis}{Raissi
  et~al\mbox{.}}{2019}]%
        {Raissi2019}
\bibfield{author}{\bibinfo{person}{M. Raissi}, \bibinfo{person}{P. Perdikaris},
  {and} \bibinfo{person}{G.E. Karniadakis}.} \bibinfo{year}{2019}\natexlab{}.
\newblock \showarticletitle{Physics-Informed Neural Networks: A Deep Learning
  Framework for Solving Forward and Inverse Problems Involving Nonlinear
  Partial Differential Equations}.
\newblock \bibinfo{journal}{\emph{J. Comput. Phys.}}  \bibinfo{volume}{378}
  (\bibinfo{date}{Feb.} \bibinfo{year}{2019}), \bibinfo{pages}{686--707}.
\newblock
\showISSN{00219991}
\urldef\tempurl%
\url{https://doi.org/10.1016/j.jcp.2018.10.045}
\showDOI{\tempurl}


\bibitem[\protect\citeauthoryear{Raissi, Perdikaris, and Karniadakis}{Raissi
  et~al\mbox{.}}{2018}]%
        {Raissi2018a}
\bibfield{author}{\bibinfo{person}{Maziar Raissi}, \bibinfo{person}{Paris
  Perdikaris}, {and} \bibinfo{person}{George~Em Karniadakis}.}
  \bibinfo{year}{2018}\natexlab{}.
\newblock \showarticletitle{Multistep {{Neural Networks}} for {{Data}}-Driven
  {{Discovery}} of {{Nonlinear Dynamical Systems}}}.
\newblock \bibinfo{journal}{\emph{arXiv:1801.01236 [nlin, physics:physics,
  stat]}} (\bibinfo{date}{Jan.} \bibinfo{year}{2018}).
\newblock
\showeprint[arxiv]{1801.01236}~[nlin, physics:physics, stat]


\bibitem[\protect\citeauthoryear{Raissi, Yazdani, and Karniadakis}{Raissi
  et~al\mbox{.}}{2020}]%
        {Raissi2020}
\bibfield{author}{\bibinfo{person}{Maziar Raissi}, \bibinfo{person}{Alireza
  Yazdani}, {and} \bibinfo{person}{George~Em Karniadakis}.}
  \bibinfo{year}{2020}\natexlab{}.
\newblock \showarticletitle{Hidden Fluid Mechanics: Learning Velocity and
  Pressure Fields from Flow Visualizations}.
\newblock \bibinfo{journal}{\emph{Science}} \bibinfo{volume}{367},
  \bibinfo{number}{6481} (\bibinfo{date}{Feb.} \bibinfo{year}{2020}),
  \bibinfo{pages}{1026--1030}.
\newblock
\showISSN{0036-8075, 1095-9203}
\urldef\tempurl%
\url{https://doi.org/10.1126/science.aaw4741}
\showDOI{\tempurl}


\bibitem[\protect\citeauthoryear{Rangapuram, Seeger, Gasthaus, Stella, Wang,
  and Januschowski}{Rangapuram et~al\mbox{.}}{2018}]%
        {NIPS2018_8004}
\bibfield{author}{\bibinfo{person}{Syama~S. Rangapuram},
  \bibinfo{person}{Matthias~W. Seeger}, \bibinfo{person}{Jan Gasthaus},
  \bibinfo{person}{Lorenzo Stella}, \bibinfo{person}{Yuyang Wang}, {and}
  \bibinfo{person}{Tim Januschowski}.} \bibinfo{year}{2018}\natexlab{}.
\newblock \showarticletitle{Deep State Space Models for Time Series
  Forecasting}. In \bibinfo{booktitle}{\emph{Advances in Neural Information
  Processing Systems 31}}, \bibfield{editor}{\bibinfo{person}{S.~Bengio},
  \bibinfo{person}{H.~Wallach}, \bibinfo{person}{H.~Larochelle},
  \bibinfo{person}{K.~Grauman}, \bibinfo{person}{N.~{Cesa-Bianchi}}, {and}
  \bibinfo{person}{R.~Garnett}} (Eds.). \bibinfo{publisher}{{Curran Associates,
  Inc.}}, \bibinfo{pages}{7785--7794}.
\newblock


\bibitem[\protect\citeauthoryear{Razavi, Tolson, and Burn}{Razavi
  et~al\mbox{.}}{2012}]%
        {Razavi2012}
\bibfield{author}{\bibinfo{person}{Saman Razavi}, \bibinfo{person}{Bryan~A.
  Tolson}, {and} \bibinfo{person}{Donald~H. Burn}.}
  \bibinfo{year}{2012}\natexlab{}.
\newblock \showarticletitle{Review of Surrogate Modeling in Water Resources:
  {{REVIEW}}}.
\newblock \bibinfo{journal}{\emph{Water Resources Research}}
  \bibinfo{volume}{48}, \bibinfo{number}{7} (\bibinfo{date}{July}
  \bibinfo{year}{2012}).
\newblock
\showISSN{00431397}
\urldef\tempurl%
\url{https://doi.org/10.1029/2011WR011527}
\showDOI{\tempurl}


\bibitem[\protect\citeauthoryear{Rezende and Mohamed}{Rezende and
  Mohamed}{2016}]%
        {rezende2016variational}
\bibfield{author}{\bibinfo{person}{Danilo~Jimenez Rezende} {and}
  \bibinfo{person}{Shakir Mohamed}.} \bibinfo{year}{2016}\natexlab{}.
\newblock \bibinfo{title}{Variational Inference with Normalizing Flows}.
\newblock
\newblock
\showeprint[arxiv]{1505.05770}~[stat.ML]


\bibitem[\protect\citeauthoryear{Rezende, Mohamed, and Wierstra}{Rezende
  et~al\mbox{.}}{2014}]%
        {rezende2014stochastic}
\bibfield{author}{\bibinfo{person}{Danilo~Jimenez Rezende},
  \bibinfo{person}{Shakir Mohamed}, {and} \bibinfo{person}{Daan Wierstra}.}
  \bibinfo{year}{2014}\natexlab{}.
\newblock \showarticletitle{Stochastic Backpropagation and Approximate
  Inference in Deep Generative Models}. In
  \bibinfo{booktitle}{\emph{International Conference on Machine Learning}}.
  {PMLR}, \bibinfo{pages}{1278--1286}.
\newblock


\bibitem[\protect\citeauthoryear{Rolnick, Donti, Kaack, Kochanski, Lacoste,
  Sankaran, Ross, {Milojevic-Dupont}, Jaques, {Waldman-Brown}, Luccioni,
  Maharaj, Sherwin, Mukkavilli, Kording, Gomes, Ng, Hassabis, Platt, Creutzig,
  Chayes, and Bengio}{Rolnick et~al\mbox{.}}{2019}]%
        {Rolnick2019}
\bibfield{author}{\bibinfo{person}{David Rolnick}, \bibinfo{person}{Priya~L.
  Donti}, \bibinfo{person}{Lynn~H. Kaack}, \bibinfo{person}{Kelly Kochanski},
  \bibinfo{person}{Alexandre Lacoste}, \bibinfo{person}{Kris Sankaran},
  \bibinfo{person}{Andrew~Slavin Ross}, \bibinfo{person}{Nikola
  {Milojevic-Dupont}}, \bibinfo{person}{Natasha Jaques}, \bibinfo{person}{Anna
  {Waldman-Brown}}, \bibinfo{person}{Alexandra Luccioni},
  \bibinfo{person}{Tegan Maharaj}, \bibinfo{person}{Evan~D. Sherwin},
  \bibinfo{person}{S.~Karthik Mukkavilli}, \bibinfo{person}{Konrad~P. Kording},
  \bibinfo{person}{Carla Gomes}, \bibinfo{person}{Andrew~Y. Ng},
  \bibinfo{person}{Demis Hassabis}, \bibinfo{person}{John~C. Platt},
  \bibinfo{person}{Felix Creutzig}, \bibinfo{person}{Jennifer Chayes}, {and}
  \bibinfo{person}{Yoshua Bengio}.} \bibinfo{year}{2019}\natexlab{}.
\newblock \showarticletitle{Tackling {{Climate Change}} with {{Machine
  Learning}}}.
\newblock \bibinfo{journal}{\emph{arXiv:1906.05433 [cs, stat]}}
  (\bibinfo{date}{Nov.} \bibinfo{year}{2019}).
\newblock
\showeprint[arxiv]{1906.05433}~[cs, stat]


\bibitem[\protect\citeauthoryear{Rossi, Chamberlain, Frasca, Eynard, Monti, and
  Bronstein}{Rossi et~al\mbox{.}}{2020}]%
        {DynamicGraphNN2020}
\bibfield{author}{\bibinfo{person}{Emanuele Rossi}, \bibinfo{person}{Ben
  Chamberlain}, \bibinfo{person}{Fabrizio Frasca}, \bibinfo{person}{Davide
  Eynard}, \bibinfo{person}{Federico Monti}, {and} \bibinfo{person}{Michael~M.
  Bronstein}.} \bibinfo{year}{2020}\natexlab{}.
\newblock \showarticletitle{Temporal Graph Networks for Deep Learning on
  Dynamic Graphs}.
\newblock \bibinfo{journal}{\emph{CoRR}}  \bibinfo{volume}{abs/2006.10637}
  (\bibinfo{year}{2020}).
\newblock
\showeprint[arxiv]{2006.10637}


\bibitem[\protect\citeauthoryear{Ruthotto and Haber}{Ruthotto and
  Haber}{2018}]%
        {Ruthotto2018}
\bibfield{author}{\bibinfo{person}{Lars Ruthotto} {and} \bibinfo{person}{Eldad
  Haber}.} \bibinfo{year}{2018}\natexlab{}.
\newblock \showarticletitle{Deep {{Neural Networks Motivated}} by {{Partial
  Differential Equations}}}.
\newblock \bibinfo{journal}{\emph{arXiv:1804.04272 [cs, math, stat]}}
  (\bibinfo{date}{Dec.} \bibinfo{year}{2018}).
\newblock
\showeprint[arxiv]{1804.04272}~[cs, math, stat]


\bibitem[\protect\citeauthoryear{Ruthotto and Haber}{Ruthotto and
  Haber}{2020}]%
        {Ruthotto2020}
\bibfield{author}{\bibinfo{person}{Lars Ruthotto} {and} \bibinfo{person}{Eldad
  Haber}.} \bibinfo{year}{2020}\natexlab{}.
\newblock \showarticletitle{Deep {{Neural Networks Motivated}} by {{Partial
  Differential Equations}}}.
\newblock \bibinfo{journal}{\emph{Journal of Mathematical Imaging and Vision}}
  \bibinfo{volume}{62}, \bibinfo{number}{3} (\bibinfo{date}{April}
  \bibinfo{year}{2020}), \bibinfo{pages}{352--364}.
\newblock
\showISSN{0924-9907, 1573-7683}
\urldef\tempurl%
\url{https://doi.org/10.1007/s10851-019-00903-1}
\showDOI{\tempurl}


\bibitem[\protect\citeauthoryear{{Sanchez-Gonzalez}, Bapst, Cranmer, and
  Battaglia}{{Sanchez-Gonzalez} et~al\mbox{.}}{2019}]%
        {Gonzalez2019}
\bibfield{author}{\bibinfo{person}{Alvaro {Sanchez-Gonzalez}},
  \bibinfo{person}{Victor Bapst}, \bibinfo{person}{Kyle Cranmer}, {and}
  \bibinfo{person}{Peter~W. Battaglia}.} \bibinfo{year}{2019}\natexlab{}.
\newblock \showarticletitle{Hamiltonian Graph Networks with {{ODE}}
  Integrators}.
\newblock \bibinfo{journal}{\emph{CoRR}}  \bibinfo{volume}{abs/1909.12790}
  (\bibinfo{year}{2019}).
\newblock
\showeprint[arxiv]{1909.12790}


\bibitem[\protect\citeauthoryear{{Sanchez-Gonzalez}, Godwin, Pfaff, Ying,
  Leskovec, and Battaglia}{{Sanchez-Gonzalez} et~al\mbox{.}}{2020}]%
        {Gonzalez2020}
\bibfield{author}{\bibinfo{person}{Alvaro {Sanchez-Gonzalez}},
  \bibinfo{person}{Jonathan Godwin}, \bibinfo{person}{Tobias Pfaff},
  \bibinfo{person}{Rex Ying}, \bibinfo{person}{Jure Leskovec}, {and}
  \bibinfo{person}{Peter~W. Battaglia}.} \bibinfo{year}{2020}\natexlab{}.
\newblock \showarticletitle{Learning to Simulate Complex Physics with Graph
  Networks}.
\newblock \bibinfo{journal}{\emph{CoRR}}  \bibinfo{volume}{abs/2002.09405}
  (\bibinfo{year}{2020}).
\newblock
\showeprint[arxiv]{2002.09405}


\bibitem[\protect\citeauthoryear{{Sanchez-Gonzalez}, Heess, Springenberg,
  Merel, Riedmiller, Hadsell, and Battaglia}{{Sanchez-Gonzalez}
  et~al\mbox{.}}{2018}]%
        {GNNs_control2018}
\bibfield{author}{\bibinfo{person}{Alvaro {Sanchez-Gonzalez}},
  \bibinfo{person}{Nicolas Heess}, \bibinfo{person}{Jost~Tobias Springenberg},
  \bibinfo{person}{Josh Merel}, \bibinfo{person}{Martin~A. Riedmiller},
  \bibinfo{person}{Raia Hadsell}, {and} \bibinfo{person}{Peter~W. Battaglia}.}
  \bibinfo{year}{2018}\natexlab{}.
\newblock \showarticletitle{Graph Networks as Learnable Physics Engines for
  Inference and Control}.
\newblock \bibinfo{journal}{\emph{CoRR}}  \bibinfo{volume}{abs/1806.01242}
  (\bibinfo{year}{2018}).
\newblock
\showeprint[arxiv]{1806.01242}


\bibitem[\protect\citeauthoryear{Scarselli, Gori, Tsoi, Hagenbuchner, and
  Monfardini}{Scarselli et~al\mbox{.}}{2009}]%
        {Scarselli2009}
\bibfield{author}{\bibinfo{person}{Franco Scarselli}, \bibinfo{person}{Marco
  Gori}, \bibinfo{person}{Ah~Chung Tsoi}, \bibinfo{person}{Markus
  Hagenbuchner}, {and} \bibinfo{person}{Gabriele Monfardini}.}
  \bibinfo{year}{2009}\natexlab{}.
\newblock \showarticletitle{The Graph Neural Network Model}.
\newblock \bibinfo{journal}{\emph{IEEE Transactions on Neural Networks}}
  \bibinfo{volume}{20}, \bibinfo{number}{1} (\bibinfo{year}{2009}),
  \bibinfo{pages}{61--80}.
\newblock
\urldef\tempurl%
\url{https://doi.org/10.1109/TNN.2008.2005605}
\showDOI{\tempurl}


\bibitem[\protect\citeauthoryear{Schoukens and Ljung}{Schoukens and
  Ljung}{2019}]%
        {Schoukens2019}
\bibfield{author}{\bibinfo{person}{Johan Schoukens} {and}
  \bibinfo{person}{Lennart Ljung}.} \bibinfo{year}{2019}\natexlab{}.
\newblock \showarticletitle{Nonlinear System Identification: A User-Oriented
  Roadmap}.
\newblock \bibinfo{journal}{\emph{CoRR}}  \bibinfo{volume}{abs/1902.00683}
  (\bibinfo{year}{2019}).
\newblock
\showeprint[arxiv]{1902.00683}


\bibitem[\protect\citeauthoryear{Schoukens and No{\"e}l}{Schoukens and
  No{\"e}l}{2017}]%
        {SCHOUKENS2017446}
\bibfield{author}{\bibinfo{person}{M. Schoukens} {and} \bibinfo{person}{J.P.
  No{\"e}l}.} \bibinfo{year}{2017}\natexlab{}.
\newblock \showarticletitle{Three {{Benchmarks Addressing Open Challenges}} in
  {{Nonlinear System Identification}}**{{We}} Thank {{Torbjorn Wigren}} and
  {{Per Mattsson}} ({{Uppsala University}}, {{Sweden}}) for Their Help in
  Realizing the Cascaded Tanks Benchmark. {{This}} Work Was Funded by the
  {{Fund}} for {{Scientific Research}} ({{FWO}}), the {{Methusalem}} Grant of
  the {{Flemish Government}} ({{METH}}-1), the {{IAP VII}}/19 {{DYSCO}}
  Program, and the {{ERC}} Advanced Grant {{SNLSID}} under Contract 320378.
  {{The}} Author {{J}}.{{P}}. {{Noel}} Is a {{Postdoctoral Researcher}} of the
  {{Fonds}} de La {{Recherche Scientifique}} - {{FNRS}} Which Is Gratefully
  Acknowledged.}
\newblock \bibinfo{journal}{\emph{IFAC-PapersOnLine}} \bibinfo{volume}{50},
  \bibinfo{number}{1} (\bibinfo{year}{2017}), \bibinfo{pages}{446--451}.
\newblock
\showISSN{2405-8963}
\urldef\tempurl%
\url{https://doi.org/10.1016/j.ifacol.2017.08.071}
\showDOI{\tempurl}


\bibitem[\protect\citeauthoryear{Schramm, Lalo, and Unterreiner}{Schramm
  et~al\mbox{.}}{2010}]%
        {Schramm2010}
\bibfield{author}{\bibinfo{person}{Dieter Schramm}, \bibinfo{person}{Wildan
  Lalo}, {and} \bibinfo{person}{Michael Unterreiner}.}
  \bibinfo{year}{2010}\natexlab{}.
\newblock \showarticletitle{Application of {{Simulators}} and {{Simulation
  Tools}} for the {{Functional Design}} of {{Mechatronic Systems}}}.
\newblock \bibinfo{journal}{\emph{Solid State Phenomena}}
  \bibinfo{volume}{166--167} (\bibinfo{date}{Sept.} \bibinfo{year}{2010}),
  \bibinfo{pages}{1--14}.
\newblock
\showISSN{1662-9779}
\urldef\tempurl%
\url{https://doi.org/10.4028/www.scientific.net/SSP.166-167.1}
\showDOI{\tempurl}


\bibitem[\protect\citeauthoryear{Sch{\"u}tt, Kindermans, Sauceda, Chmiela,
  Tkatchenko, and M{\"u}ller}{Sch{\"u}tt et~al\mbox{.}}{2017}]%
        {schutt2017schnet}
\bibfield{author}{\bibinfo{person}{Kristof~T. Sch{\"u}tt},
  \bibinfo{person}{Pieter-Jan Kindermans}, \bibinfo{person}{Huziel~E. Sauceda},
  \bibinfo{person}{Stefan Chmiela}, \bibinfo{person}{Alexandre Tkatchenko},
  {and} \bibinfo{person}{Klaus-Robert M{\"u}ller}.}
  \bibinfo{year}{2017}\natexlab{}.
\newblock \bibinfo{title}{{{SchNet}}: A Continuous-Filter Convolutional Neural
  Network for Modeling Quantum Interactions}.
\newblock
\newblock
\showeprint[arxiv]{1706.08566}~[stat.ML]


\bibitem[\protect\citeauthoryear{Skomski, Drgona, and Tuor}{Skomski
  et~al\mbox{.}}{2020}]%
        {skomski2020physicsinformed}
\bibfield{author}{\bibinfo{person}{Elliott Skomski}, \bibinfo{person}{Jan
  Drgona}, {and} \bibinfo{person}{Aaron Tuor}.}
  \bibinfo{year}{2020}\natexlab{}.
\newblock \bibinfo{title}{Physics-Informed Neural State Space Models via
  Learning and Evolution}.
\newblock
\newblock
\showeprint[arxiv]{2011.13497}~[cs.NE]


\bibitem[\protect\citeauthoryear{Skomski, Vasisht, Wight, Tuor, Drgona, and
  Vrabie}{Skomski et~al\mbox{.}}{2021}]%
        {skomski2021constrained}
\bibfield{author}{\bibinfo{person}{Elliott Skomski}, \bibinfo{person}{Soumya
  Vasisht}, \bibinfo{person}{Colby Wight}, \bibinfo{person}{Aaron Tuor},
  \bibinfo{person}{Jan Drgona}, {and} \bibinfo{person}{Draguna Vrabie}.}
  \bibinfo{year}{2021}\natexlab{}.
\newblock \bibinfo{title}{Constrained Block Nonlinear Neural Dynamical Models}.
\newblock
\newblock
\showeprint[arxiv]{2101.01864}~[math.DS]


\bibitem[\protect\citeauthoryear{Sohlberg and Jacobsen}{Sohlberg and
  Jacobsen}{2008}]%
        {Sohlberg2008}
\bibfield{author}{\bibinfo{person}{B. Sohlberg} {and} \bibinfo{person}{E.W.
  Jacobsen}.} \bibinfo{year}{2008}\natexlab{}.
\newblock \showarticletitle{{{GREY BOX MODELLING}} \textendash{} {{BRANCHES AND
  EXPERIENCES}}}.
\newblock \bibinfo{journal}{\emph{IFAC Proceedings Volumes}}
  \bibinfo{volume}{41}, \bibinfo{number}{2} (\bibinfo{year}{2008}),
  \bibinfo{pages}{11415--11420}.
\newblock
\showISSN{14746670}
\urldef\tempurl%
\url{https://doi.org/10.3182/20080706-5-KR-1001.01934}
\showDOI{\tempurl}


\bibitem[\protect\citeauthoryear{Suk, Wee, Lee, and Shen}{Suk
  et~al\mbox{.}}{2016}]%
        {SUK2016292}
\bibfield{author}{\bibinfo{person}{Heung-Il Suk}, \bibinfo{person}{Chong-Yaw
  Wee}, \bibinfo{person}{Seong-Whan Lee}, {and} \bibinfo{person}{Dinggang
  Shen}.} \bibinfo{year}{2016}\natexlab{}.
\newblock \showarticletitle{State-Space Model with Deep Learning for Functional
  Dynamics Estimation in Resting-State {{fMRI}}}.
\newblock \bibinfo{journal}{\emph{NeuroImage}}  \bibinfo{volume}{129}
  (\bibinfo{year}{2016}), \bibinfo{pages}{292--307}.
\newblock
\showISSN{1053-8119}
\urldef\tempurl%
\url{https://doi.org/10.1016/j.neuroimage.2016.01.005}
\showDOI{\tempurl}


\bibitem[\protect\citeauthoryear{Toth, Rezende, Jaegle, Racani{\`e}re, Botev,
  and Higgins}{Toth et~al\mbox{.}}{2020}]%
        {Toth2020}
\bibfield{author}{\bibinfo{person}{Peter Toth}, \bibinfo{person}{Danilo~Jimenez
  Rezende}, \bibinfo{person}{Andrew Jaegle}, \bibinfo{person}{S{\'e}bastien
  Racani{\`e}re}, \bibinfo{person}{Aleksandar Botev}, {and}
  \bibinfo{person}{Irina Higgins}.} \bibinfo{year}{2020}\natexlab{}.
\newblock \showarticletitle{Hamiltonian {{Generative Networks}}}.
\newblock \bibinfo{journal}{\emph{arXiv:1909.13789 [cs, stat]}}
  (\bibinfo{date}{Feb.} \bibinfo{year}{2020}).
\newblock
\showeprint[arxiv]{1909.13789}~[cs, stat]


\bibitem[\protect\citeauthoryear{Unke and Meuwly}{Unke and Meuwly}{2018}]%
        {Unke2018}
\bibfield{author}{\bibinfo{person}{Oliver~T. Unke} {and}
  \bibinfo{person}{Markus Meuwly}.} \bibinfo{year}{2018}\natexlab{}.
\newblock \showarticletitle{A Reactive, Scalable, and Transferable Model for
  Molecular Energies from a Neural Network Approach Based on Local
  Information}.
\newblock \bibinfo{journal}{\emph{The Journal of Chemical Physics}}
  \bibinfo{volume}{148}, \bibinfo{number}{24} (\bibinfo{year}{2018}),
  \bibinfo{pages}{241708}.
\newblock
\urldef\tempurl%
\url{https://doi.org/10.1063/1.5017898}
\showDOI{\tempurl}
\showeprint{https://doi.org/10.1063/1.5017898}


\bibitem[\protect\citeauthoryear{Unke and Meuwly}{Unke and Meuwly}{2019}]%
        {PhysNet2019}
\bibfield{author}{\bibinfo{person}{Oliver~T. Unke} {and}
  \bibinfo{person}{Markus Meuwly}.} \bibinfo{year}{2019}\natexlab{}.
\newblock \showarticletitle{{{PhysNet}}: A Neural Network for Predicting
  Energies, Forces, Dipole Moments, and Partial Charges}.
\newblock \bibinfo{journal}{\emph{Journal of Chemical Theory and Computation}}
  \bibinfo{volume}{15}, \bibinfo{number}{6} (\bibinfo{year}{2019}),
  \bibinfo{pages}{3678--3693}.
\newblock
\urldef\tempurl%
\url{https://doi.org/10.1021/acs.jctc.9b00181}
\showDOI{\tempurl}
\showeprint{https://doi.org/10.1021/acs.jctc.9b00181}


\bibitem[\protect\citeauthoryear{Viana, Gogu, and Haftka}{Viana
  et~al\mbox{.}}{2010}]%
        {Viana2010}
\bibfield{author}{\bibinfo{person}{Felipe A.~C. Viana},
  \bibinfo{person}{Christian Gogu}, {and} \bibinfo{person}{Raphael~T. Haftka}.}
  \bibinfo{year}{2010}\natexlab{}.
\newblock \showarticletitle{Making the {{Most Out}} of {{Surrogate Models}}:
  Tricks of the {{Trade}}}. In \bibinfo{booktitle}{\emph{Volume 1: 36th
  {{Design Automation Conference}}, {{Parts A}} and {{B}}}}.
  \bibinfo{publisher}{{ASMEDC}}, \bibinfo{address}{{Montreal, Quebec, Canada}},
  \bibinfo{pages}{587--598}.
\newblock
\showISBNx{978-0-7918-4409-0}
\urldef\tempurl%
\url{https://doi.org/10.1115/DETC2010-28813}
\showDOI{\tempurl}


\bibitem[\protect\citeauthoryear{{von Rueden}, Mayer, Beckh, Georgiev,
  Giesselbach, Heese, Kirsch, Pfrommer, Pick, Ramamurthy, Walczak, Garcke,
  Bauckhage, and Schuecker}{{von Rueden} et~al\mbox{.}}{2020a}]%
        {Rueden2020a}
\bibfield{author}{\bibinfo{person}{Laura {von Rueden}},
  \bibinfo{person}{Sebastian Mayer}, \bibinfo{person}{Katharina Beckh},
  \bibinfo{person}{Bogdan Georgiev}, \bibinfo{person}{Sven Giesselbach},
  \bibinfo{person}{Raoul Heese}, \bibinfo{person}{Birgit Kirsch},
  \bibinfo{person}{Julius Pfrommer}, \bibinfo{person}{Annika Pick},
  \bibinfo{person}{Rajkumar Ramamurthy}, \bibinfo{person}{Michal Walczak},
  \bibinfo{person}{Jochen Garcke}, \bibinfo{person}{Christian Bauckhage}, {and}
  \bibinfo{person}{Jannis Schuecker}.} \bibinfo{year}{2020}\natexlab{a}.
\newblock \showarticletitle{Informed {{Machine Learning}} -- {{A Taxonomy}} and
  {{Survey}} of {{Integrating Knowledge}} into {{Learning Systems}}}.
\newblock \bibinfo{journal}{\emph{arXiv:1903.12394 [cs, stat]}}
  (\bibinfo{date}{Feb.} \bibinfo{year}{2020}).
\newblock
\showeprint[arxiv]{1903.12394}~[cs, stat]


\bibitem[\protect\citeauthoryear{{von Rueden}, Mayer, Sifa, Bauckhage, and
  Garcke}{{von Rueden} et~al\mbox{.}}{2020b}]%
        {Rueden2020}
\bibfield{author}{\bibinfo{person}{Laura {von Rueden}},
  \bibinfo{person}{Sebastian Mayer}, \bibinfo{person}{Rafet Sifa},
  \bibinfo{person}{Christian Bauckhage}, {and} \bibinfo{person}{Jochen
  Garcke}.} \bibinfo{year}{2020}\natexlab{b}.
\newblock \showarticletitle{Combining {{Machine Learning}} and {{Simulation}}
  to a {{Hybrid Modelling Approach}}: Current and {{Future Directions}}}.
\newblock \bibinfo{journal}{\emph{Advances in Intelligent Data Analysis XVIII}}
   \bibinfo{volume}{12080} (\bibinfo{year}{2020}), \bibinfo{pages}{548--560}.
\newblock
\urldef\tempurl%
\url{https://doi.org/10.1007/978-3-030-44584-3_43}
\showDOI{\tempurl}


\bibitem[\protect\citeauthoryear{Wang, Olsson, Wehmeyer, P{\'e}rez, Charron,
  {de Fabritiis}, No{\'e}, and Clementi}{Wang et~al\mbox{.}}{2019}]%
        {Wang2019}
\bibfield{author}{\bibinfo{person}{Jiang Wang}, \bibinfo{person}{Simon Olsson},
  \bibinfo{person}{Christoph Wehmeyer}, \bibinfo{person}{Adri{\`a} P{\'e}rez},
  \bibinfo{person}{Nicholas~E. Charron}, \bibinfo{person}{Gianni {de
  Fabritiis}}, \bibinfo{person}{Frank No{\'e}}, {and} \bibinfo{person}{Cecilia
  Clementi}.} \bibinfo{year}{2019}\natexlab{}.
\newblock \showarticletitle{Machine Learning of Coarse-Grained Molecular
  Dynamics Force Fields}.
\newblock \bibinfo{journal}{\emph{ACS Central Science}} \bibinfo{volume}{5},
  \bibinfo{number}{5} (\bibinfo{year}{2019}), \bibinfo{pages}{755--767}.
\newblock
\urldef\tempurl%
\url{https://doi.org/10.1021/acscentsci.8b00913}
\showDOI{\tempurl}
\showeprint{https://doi.org/10.1021/acscentsci.8b00913}


\bibitem[\protect\citeauthoryear{Wang, Teng, and Perdikaris}{Wang
  et~al\mbox{.}}{2020a}]%
        {Wang2020a}
\bibfield{author}{\bibinfo{person}{Sifan Wang}, \bibinfo{person}{Yujun Teng},
  {and} \bibinfo{person}{Paris Perdikaris}.} \bibinfo{year}{2020}\natexlab{a}.
\newblock \showarticletitle{Understanding and Mitigating Gradient Pathologies
  in Physics-Informed Neural Networks}.
\newblock \bibinfo{journal}{\emph{arXiv:2001.04536 [cs, math, stat]}}
  (\bibinfo{date}{Jan.} \bibinfo{year}{2020}).
\newblock
\showeprint[arxiv]{2001.04536}~[cs, math, stat]


\bibitem[\protect\citeauthoryear{Wang, Yu, and Perdikaris}{Wang
  et~al\mbox{.}}{2020b}]%
        {Wang2020}
\bibfield{author}{\bibinfo{person}{Sifan Wang}, \bibinfo{person}{Xinling Yu},
  {and} \bibinfo{person}{Paris Perdikaris}.} \bibinfo{year}{2020}\natexlab{b}.
\newblock \showarticletitle{When and Why {{PINNs}} Fail to Train: A Neural
  Tangent Kernel Perspective}.
\newblock \bibinfo{journal}{\emph{arXiv:2007.14527 [cs, math, stat]}}
  (\bibinfo{date}{July} \bibinfo{year}{2020}).
\newblock
\showeprint[arxiv]{2007.14527}~[cs, math, stat]


\bibitem[\protect\citeauthoryear{Wanner and Hairer}{Wanner and Hairer}{1991}]%
        {Wanner1991}
\bibfield{author}{\bibinfo{person}{G. Wanner} {and} \bibinfo{person}{E.
  Hairer}.} \bibinfo{year}{1991}\natexlab{}.
\newblock \bibinfo{booktitle}{\emph{Solving Ordinary Differential Equations
  {{I}}: Nonstiff {{Problems}}} (\bibinfo{edition}{springer s} ed.)}.
  Vol.~\bibinfo{volume}{1}.
\newblock \bibinfo{publisher}{{Springer-Verlag}}.
\newblock


\bibitem[\protect\citeauthoryear{Watters, Zoran, Weber, Battaglia, Pascanu, and
  Tacchetti}{Watters et~al\mbox{.}}{2017}]%
        {NIPS2017_8cbd005a}
\bibfield{author}{\bibinfo{person}{Nicholas Watters}, \bibinfo{person}{Daniel
  Zoran}, \bibinfo{person}{Theophane Weber}, \bibinfo{person}{Peter Battaglia},
  \bibinfo{person}{Razvan Pascanu}, {and} \bibinfo{person}{Andrea Tacchetti}.}
  \bibinfo{year}{2017}\natexlab{}.
\newblock \showarticletitle{Visual Interaction Networks: Learning a Physics
  Simulator from Video}. In \bibinfo{booktitle}{\emph{Advances in Neural
  Information Processing Systems}},
  \bibfield{editor}{\bibinfo{person}{I.~Guyon}, \bibinfo{person}{U.~V.
  Luxburg}, \bibinfo{person}{S.~Bengio}, \bibinfo{person}{H.~Wallach},
  \bibinfo{person}{R.~Fergus}, \bibinfo{person}{S.~Vishwanathan}, {and}
  \bibinfo{person}{R.~Garnett}} (Eds.), Vol.~\bibinfo{volume}{30}.
  \bibinfo{publisher}{{Curran Associates, Inc.}}
\newblock


\bibitem[\protect\citeauthoryear{Westermann and Evins}{Westermann and
  Evins}{2019}]%
        {Westermann2019}
\bibfield{author}{\bibinfo{person}{Paul Westermann} {and}
  \bibinfo{person}{Ralph Evins}.} \bibinfo{year}{2019}\natexlab{}.
\newblock \showarticletitle{Surrogate Modelling for Sustainable Building Design
  \textendash{} {{A}} Review}.
\newblock \bibinfo{journal}{\emph{Energy and Buildings}}  \bibinfo{volume}{198}
  (\bibinfo{date}{Sept.} \bibinfo{year}{2019}), \bibinfo{pages}{170--186}.
\newblock
\showISSN{03787788}
\urldef\tempurl%
\url{https://doi.org/10.1016/j.enbuild.2019.05.057}
\showDOI{\tempurl}


\bibitem[\protect\citeauthoryear{Wu, Mardt, Pasquali, and Noe}{Wu
  et~al\mbox{.}}{2018}]%
        {wu2019deep}
\bibfield{author}{\bibinfo{person}{Hao Wu}, \bibinfo{person}{Andreas Mardt},
  \bibinfo{person}{Luca Pasquali}, {and} \bibinfo{person}{Frank Noe}.}
  \bibinfo{year}{2018}\natexlab{}.
\newblock \showarticletitle{Deep Generative Markov State Models}.
\newblock \bibinfo{journal}{\emph{arXiv preprint arXiv:1805.07601}}
  (\bibinfo{year}{2018}).
\newblock
\showeprint[arxiv]{1805.07601}


\bibitem[\protect\citeauthoryear{Wu, Pan, Chen, Long, Zhang, and Yu}{Wu
  et~al\mbox{.}}{2019}]%
        {GNN_survey2019}
\bibfield{author}{\bibinfo{person}{Zonghan Wu}, \bibinfo{person}{Shirui Pan},
  \bibinfo{person}{Fengwen Chen}, \bibinfo{person}{Guodong Long},
  \bibinfo{person}{Chengqi Zhang}, {and} \bibinfo{person}{Philip~S. Yu}.}
  \bibinfo{year}{2019}\natexlab{}.
\newblock \showarticletitle{A Comprehensive Survey on Graph Neural Networks}.
\newblock \bibinfo{journal}{\emph{CoRR}}  \bibinfo{volume}{abs/1901.00596}
  (\bibinfo{year}{2019}).
\newblock
\showeprint[arxiv]{1901.00596}


\bibitem[\protect\citeauthoryear{Xu, Chen, Li, and Duvenaud}{Xu
  et~al\mbox{.}}{2021}]%
        {Xu2021}
\bibfield{author}{\bibinfo{person}{Winnie Xu}, \bibinfo{person}{Ricky T.~Q.
  Chen}, \bibinfo{person}{Xuechen Li}, {and} \bibinfo{person}{David Duvenaud}.}
  \bibinfo{year}{2021}\natexlab{}.
\newblock \showarticletitle{Infinitely {{Deep Bayesian Neural Networks}} with
  {{Stochastic Differential Equations}}}.
\newblock \bibinfo{journal}{\emph{arXiv:2102.06559 [cs, stat]}}
  (\bibinfo{date}{Aug.} \bibinfo{year}{2021}).
\newblock
\showeprint[arxiv]{2102.06559}~[cs, stat]


\bibitem[\protect\citeauthoryear{Zhang, Han, Wang, Car, and E}{Zhang
  et~al\mbox{.}}{2018b}]%
        {PhysRevLett143001}
\bibfield{author}{\bibinfo{person}{Linfeng Zhang}, \bibinfo{person}{Jiequn
  Han}, \bibinfo{person}{Han Wang}, \bibinfo{person}{Roberto Car}, {and}
  \bibinfo{person}{Weinan E}.} \bibinfo{year}{2018}\natexlab{b}.
\newblock \showarticletitle{Deep Potential Molecular Dynamics: A Scalable Model
  with the Accuracy of Quantum Mechanics}.
\newblock \bibinfo{journal}{\emph{Physical Review Letters}}
  \bibinfo{volume}{120}, \bibinfo{number}{14} (\bibinfo{date}{April}
  \bibinfo{year}{2018}), \bibinfo{pages}{143001}.
\newblock
\urldef\tempurl%
\url{https://doi.org/10.1103/PhysRevLett.120.143001}
\showDOI{\tempurl}


\bibitem[\protect\citeauthoryear{Zhang, Han, Wang, Saidi, Car, and E}{Zhang
  et~al\mbox{.}}{2018c}]%
        {zhang2018endtoend}
\bibfield{author}{\bibinfo{person}{Linfeng Zhang}, \bibinfo{person}{Jiequn
  Han}, \bibinfo{person}{Han Wang}, \bibinfo{person}{Wissam~A. Saidi},
  \bibinfo{person}{Roberto Car}, {and} \bibinfo{person}{Weinan E}.}
  \bibinfo{year}{2018}\natexlab{c}.
\newblock \bibinfo{title}{End-to-End Symmetry Preserving Inter-Atomic Potential
  Energy Model for Finite and Extended Systems}.
\newblock
\newblock
\showeprint[arxiv]{1805.09003}~[physics.comp-ph]


\bibitem[\protect\citeauthoryear{Zhang, Cui, and Zhu}{Zhang
  et~al\mbox{.}}{2018a}]%
        {Zhang2015}
\bibfield{author}{\bibinfo{person}{Ziwei Zhang}, \bibinfo{person}{Peng Cui},
  {and} \bibinfo{person}{Wenwu Zhu}.} \bibinfo{year}{2018}\natexlab{a}.
\newblock \showarticletitle{Deep Learning on Graphs: A Survey}.
\newblock \bibinfo{journal}{\emph{CoRR}}  \bibinfo{volume}{abs/1812.04202}
  (\bibinfo{year}{2018}).
\newblock
\showeprint[arxiv]{1812.04202}


\bibitem[\protect\citeauthoryear{Zhong, Dey, and Chakraborty}{Zhong
  et~al\mbox{.}}{2019}]%
        {SymODEnet2019}
\bibfield{author}{\bibinfo{person}{Yaofeng~Desmond Zhong},
  \bibinfo{person}{Biswadip Dey}, {and} \bibinfo{person}{Amit Chakraborty}.}
  \bibinfo{year}{2019}\natexlab{}.
\newblock \showarticletitle{Symplectic {{ODE}}-{{Net}}: Learning Hamiltonian
  Dynamics with Control}.
\newblock \bibinfo{journal}{\emph{CoRR}}  \bibinfo{volume}{abs/1909.12077}
  (\bibinfo{year}{2019}).
\newblock
\showeprint[arxiv]{1909.12077}


\bibitem[\protect\citeauthoryear{Zhou, Cui, Hu, Zhang, Yang, Liu, Wang, Li, and
  Sun}{Zhou et~al\mbox{.}}{2020}]%
        {ZHOU202057}
\bibfield{author}{\bibinfo{person}{Jie Zhou}, \bibinfo{person}{Ganqu Cui},
  \bibinfo{person}{Shengding Hu}, \bibinfo{person}{Zhengyan Zhang},
  \bibinfo{person}{Cheng Yang}, \bibinfo{person}{Zhiyuan Liu},
  \bibinfo{person}{Lifeng Wang}, \bibinfo{person}{Changcheng Li}, {and}
  \bibinfo{person}{Maosong Sun}.} \bibinfo{year}{2020}\natexlab{}.
\newblock \showarticletitle{Graph Neural Networks: A Review of Methods and
  Applications}.
\newblock \bibinfo{journal}{\emph{AI Open}}  \bibinfo{volume}{1}
  (\bibinfo{year}{2020}), \bibinfo{pages}{57--81}.
\newblock
\showISSN{2666-6510}
\urldef\tempurl%
\url{https://doi.org/10.1016/j.aiopen.2021.01.001}
\showDOI{\tempurl}


\end{thebibliography}

\end{document}